\pdfoutput=1
\documentclass[10pt,twocolumn,letterpaper]{article}

\usepackage{iccv}
\usepackage{times}
\usepackage{epsfig}
\usepackage{graphicx}
\usepackage{amsmath}
\usepackage{amssymb}

\usepackage{textcomp}
\usepackage{multirow}

\usepackage[symbol]{footmisc}

\usepackage{tikz}
\usetikzlibrary{positioning,calc}
\usetikzlibrary{decorations,decorations.markings}
\usetikzlibrary{fit}
\usetikzlibrary{shapes,arrows,shadows}

\newcommand{\refsec}[1]{Section~\ref{#1}}
\newcommand{\reffig}[1]{Figure~\ref{#1}}

\newcommand{\normd}{\mathcal{N}}

\newcommand{\kld}{\mathcal{D}_{\text{KL}}}
\newcommand{\dkl}{\kld}

\newcommand{\real}{\mathbb{R}}

\renewcommand{\hat}[1]{\widehat{#1}}

\usepackage[pagebackref=true,breaklinks=true,letterpaper=true,colorlinks,bookmarks=false]{hyperref}

\iccvfinalcopy 

\def\iccvPaperID{4094} 

\ificcvfinal\fi
\begin{document}

\title{Geometric Disentanglement for Generative Latent Shape Models}

\author{
Tristan Aumentado-Armstrong\thanks{\scriptsize{The work in this article was done while Tristan A.A.\ was a student at the University of Toronto. Sven Dickinson, Allan Jepson, and Stavros Tsogkas contributed in their capacity as Professors and Postdoc at the University of Toronto, respectively. 
The views expressed (or the conclusions reached) are their own and do not necessarily represent the views of Samsung Research America, Inc.}},~ 
Stavros Tsogkas, Allan Jepson, Sven Dickinson \\
University of Toronto\quad Vector Institute for AI\quad Samsung AI Center, Toronto\\
{\tt\small taumen@cs.utoronto.ca,  \{stavros.t,allan.jepson,s.dickinson\}@samsung.com}
}

\maketitle
\ificcvfinal\thispagestyle{empty}\fi

\begin{abstract}
   Representing 3D shape is a fundamental problem in artificial intelligence, which has numerous applications within computer vision and graphics. 
   One avenue that has recently begun to be explored is the use of latent representations of generative models.
   However, it remains an open problem to learn a generative model of shape that is 
   interpretable
   and 
   easily manipulated,  
   particularly in the absence of supervised labels.
   In this paper, we propose an unsupervised approach to partitioning the latent space of a variational autoencoder for 3D point clouds in a natural way, using only geometric information.
   Our method makes use of tools from spectral differential geometry to separate intrinsic and extrinsic shape information, 
   and then considers several hierarchical disentanglement penalties for dividing the latent space in this manner, including a novel one that penalizes the Jacobian of the latent representation of the decoded output with respect to the latent encoding. 
   We show that the resulting representation exhibits intuitive and interpretable behavior, 
   enabling tasks such as pose transfer
   and pose-aware shape retrieval
    that cannot easily be performed by models with an entangled representation. 
\end{abstract}
\section{Introduction} \label{sec:introduction}

\begin{figure}[ht]
    \centering
    \includegraphics[height=1.4in]{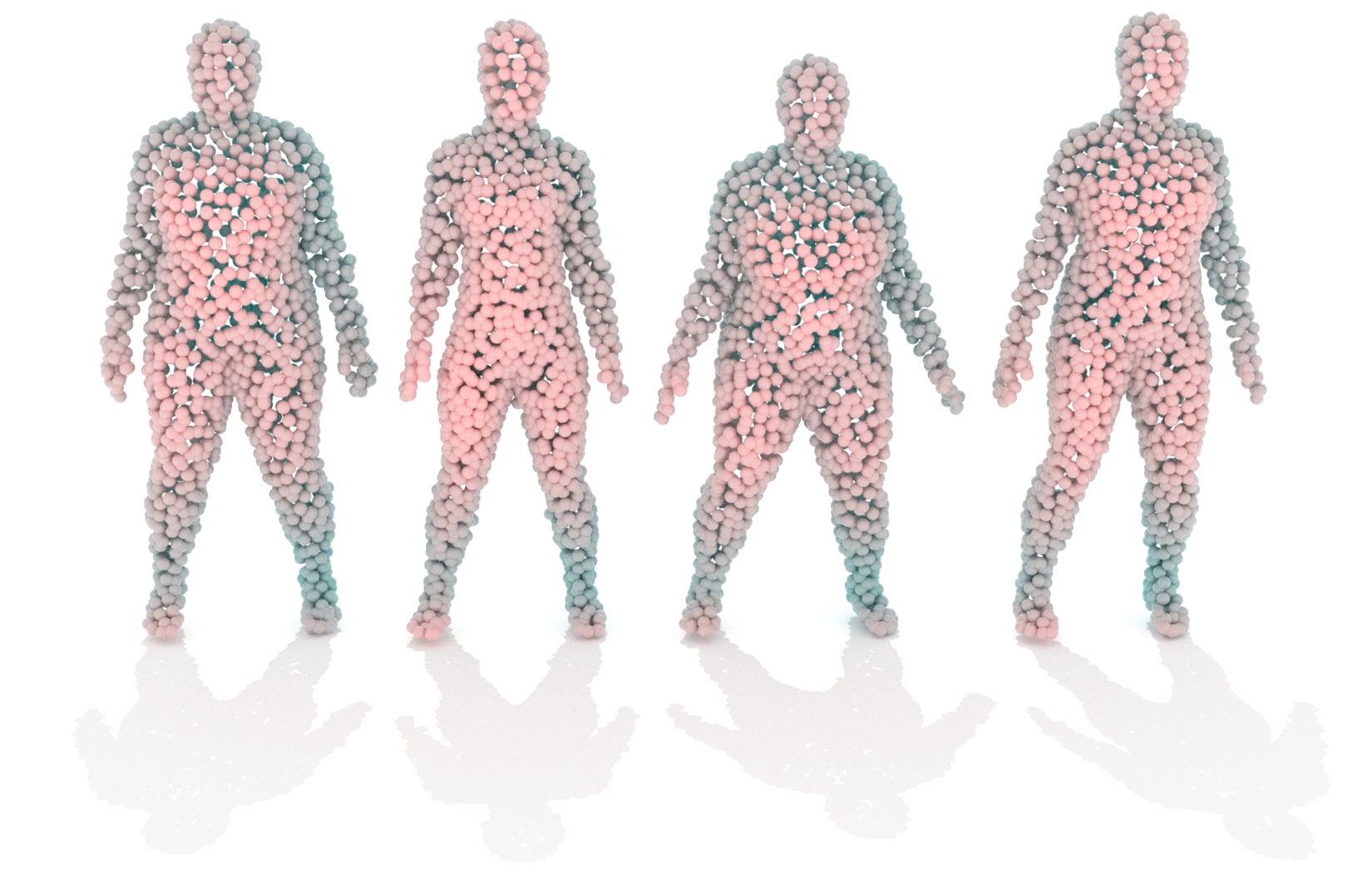}\hfill 
    \includegraphics[width=0.95\linewidth]{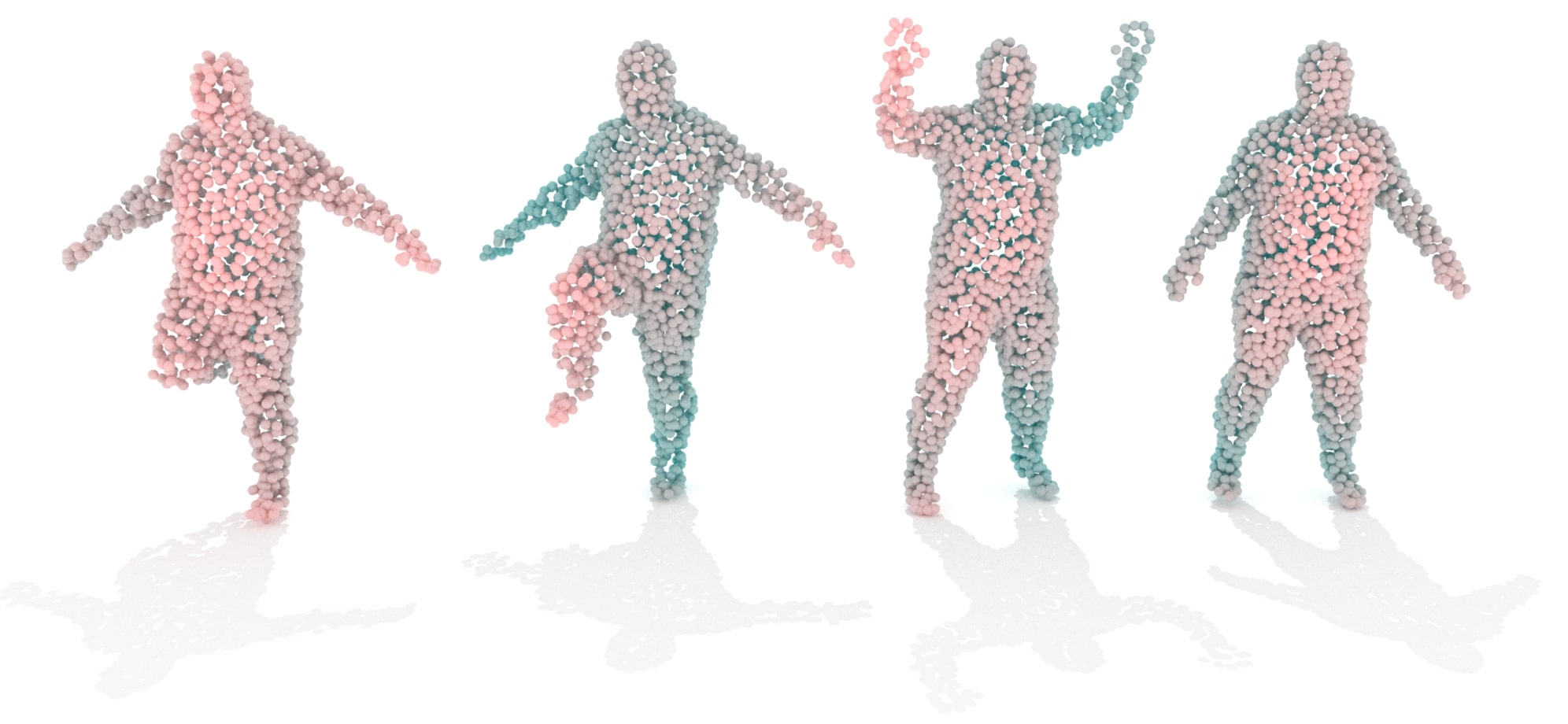}
    \caption{Factoring pose and intrinsic shape within a disentangled latent space offers fine-grained control when generating
        shapes using a generative model. 
        \textbf{Top:} decoded shapes with constant latent extrinsic group and randomly sampled latent intrinsics.
       	\textbf{Bottom:} decoded shapes with fixed latent intrinsic group and random extrinsics.
       	Colors denote depth (i.e., distance from the camera).
    }
    \label{fig:teaser}
\end{figure}
Fitting and manipulating 3D shape (e.g., for inferring 3D structure from images or efficiently computing animations) are core problems in computer vision and graphics. 
Unfortunately, designing an appropriate representation of 3D object shape is a non-trivial, and, often, task-dependent issue. 

One way to approach this problem is to use deep generative models, such as generative adversarial networks (GANs)~\cite{goodfellow2014generative} or variational autoencoders (VAEs)~\cite{rezende2014stochastic,kingma2013auto}.  
These methods are not only capable of generating novel examples of data points, but also produce a latent space that provides a compressed, continuous vector representation of the data, allowing efficient manipulation.
Rather than performing explicit physical calculations, for example, one can imagine performing approximate ``intuitive'' physics by predicting movements in the latent space instead.

However, a natural representation for 3D objects is likely to be highly structured, 
with different variables controlling separate aspects of an object.
In general, this notion of \textit{disentanglement}~\cite{bengio2013representation}
is a major tenet of representation learning, that 
closely aligns with human reasoning, and is supported by neuroscientific
findings~\cite{barlow1961possible,higgins2011role,higgins2016early}.
Given the utility of disentangled representations, a natural question is 
whether we can structure the latent space in a purely unsupervised manner. 
In the context of 3D shapes, this is equivalent to asking how one can factor the 
representation into interpretable components using geometric information alone. 

We take two main steps in this direction. 
First, we leverage methods from spectral differential 
geometry, defining a notion of intrinsic shape based on the Laplace-Beltrami 
operator (LBO) spectrum.
This provides a fully unsupervised descriptor of shape that can be computed from 
the geometry alone and is invariant to isometric pose changes.
Furthermore, unlike semantic labels, the spectrum is continuous, 
catering to the intuition that ``shape'' should 
be a smoothly deformable object property.
It also automatically divorces the intrinsic or ``core'' shape representation 
from rigid or isometric (e.g., articulated) transforms, which we call extrinsic shape. 
Second, we build on a two-level architecture for 
generative point cloud models~\cite{achlioptas2018learning} and examine several 
approaches to hierarchical latent disentanglement. 
In addition to a previously used information-theoretic penalty based on  
total correlation, we describe a hierarchical flavor of a covariance-based technique, 
and propose a novel penalty term, based on the Jacobian between latent variables. 
Together, these methods allow us to learn a factored representation of 3D shape 
using only geometric information in an unsupervised manner. 
This representation can then be applied to several tasks, including non-rigid pose manipulation (as in~\reffig{fig:teaser}) and pose-aware shape retrieval, in addition to generative sampling of new shapes.

\section{Related Work} \label{sec:related}

\subsection{Latent Disentanglement in Generative Models}
\label{sec:related:latent}

A number of techniques for disentangling VAEs have recently arisen, often based on the distributional properties of the latent prior.
One such method is the $\beta$-VAE \cite{higgins2017beta,burgess2018understanding},
in which one can enforce greater disentanglement at the cost of poorer reconstruction quality.
As a result, researchers have proposed several information-theoretic approaches that utilize a penalty on the total correlation (TC), a multivariate generalization of the mutual information \cite{watanabe1960information}.
Minimizing TC corresponds to minimizing the information shared among variables, making it a powerful disentanglement technique \cite{gao2018auto,chen2018isolating,kim2018disentangling}.
Yet, such methods do not consider \emph{groups} of latent variables,
and do not control the strength of disentanglement \emph{between} versus \emph{within} groups.
Since geometric shape properties in our model cannot be described with a single variable, 
our intrinsic-extrinsic factorization requires \emph{hierarchical} disentanglement.
Fortunately, a multi-level decomposition of the ELBO can be used to obtain a hierarchical TC penalty \cite{esmaeili2018structured}.

Other examples of disentanglement algorithms include 
information-theoretic methods in GANs \cite{chen2016infogan},
latent whitening \cite{hahn2018disentangling},
covariance penalization \cite{kumar2017variational},
and
Bayesian hyper-priors \cite{ansari2018hyperprior}.
A number of techniques also utilize known groupings or discrete labels of the data \cite{hosoya2018simple,bouchacourt2017multi,ruiz:hal-01896007,hadad2018two}.
In contrast, our work does not have access to discrete groupings (given the continuity of the spectrum), requires a hierarchical structuring, and utilizes no domain knowledge outside of the geometry itself.
We therefore consider
three approaches to hierarchical disentanglement: 
(i) a TC penalty; (ii) a decomposed covariance loss; and (iii) shrinking the Jacobian between latent groups.

\subsection{Deep Generative Models of 3D Point Clouds} \label{sec:related:point}

Point clouds represent a practical alternative to voxel and mesh representations for 3D shape.
Although they do not model the complex connectivity information of meshes, 
point clouds can still capture high resolution details at lower computational cost than voxel-based methods.
One other benefit is that much real-world data in computer vision is captured as point sets, 
which has resulted in
considerable effort on learning from point cloud data. 
However, complications arise from the set-valued nature of each datum~\cite{ravanbakhsh2016deep}. 
PointNet~\cite{qi2017pointnet} handles that by using a series of 1D convolutions and affine transforms, 
followed by pooling and fully-connected layers.
Many approaches have tried to integrate neighborhood information into this encoder
(e.g., \cite{qi2017pointnet++,hermosilla2018monte,xu2018spidercnn,atzmon2018point}), 
but this remains an open problem.

Several generative models of point clouds exist: 
Nash and Williams~\cite{nash2017shape} utilize a VAE on data of 3D part segmentations and
associated normals, whereas Achlioptas et al.~\cite{achlioptas2018learning} 
use a GAN.
Li et al.~\cite{li2018point} adopt a hierarchical sampling approach with a more general GAN loss, while Valsesia et al.~\cite{valsesia2018learning} utilize a graph convolutional method with a GAN loss.
In comparison to these methods, we focus on unsupervised geometric disentanglement of the latent representation, allowing us to factor pose and intrinsic shape, and use it for downstream tasks. 
We also do not require additional information, such as part segmentations.
Compared to standard GANs, the use of a VAE permits natural probabilistic approaches to hierarchical disentanglement, as well as the presence of an encoder, which is necessary for latent representation manipulations and tasks such as retrieval.
In this sense, our work is orthogonal to GAN-based representation learning, and both techniques may be mutually applicable as joint VAE-GAN models advance 
(e.g., \cite{mescheder2017adversarial,zhao2018information}).

Two recent related works utilize meshes for deformation-aware 3D generative modelling.
Tan et al.~\cite{tan2018variational} utilize latent manipulation to perform a variety of tasks, but does not explicitly separate pose and shape.
Gao et al.~\cite{gao2018automatic} fix two domains per model, making intrinsic shape variation and comparing latent vectors difficult.
Both works are limited by the need for identical connectivity.
In contrast, we can smoothly explore latent shape and pose independently, without labels or correspondence.
We further note that our disentanglement framework is modality-agnostic to the extent that only the AE details need change.


In this work, we utilize point cloud data to learn a latent representation of 3D shape, capable of encoding, decoding, and novel sampling.
Using PointNet as the encoder, we define a VAE on the latent space of a deterministic autoencoder, similar to~\cite{achlioptas2018learning}.
Our main goal is to investigate how unsupervised geometric disentanglement using spectral information 
can be used to structure the latent space of shape in a more interpretable 
and potentially more useful manner.

\section{Point Cloud Autoencoder} \label{sec:ae}

\tikzset{%
	block/.style    = {draw, thick, rectangle, minimum height = 1.9em,
		minimum width = 1.9em},
	fm/.style      = {draw, circle, node distance = 1.7cm}, 
	fm2/.style      = {draw, circle, node distance = 1.7cm}, 
	input/.style    = {coordinate}, 
	output/.style   = {coordinate}, 
	fitting node/.style={
		inner sep=0pt,
		fill=none,
		draw=none,
		reset transform,
		fit={(\pgf@pathminx,\pgf@pathminy) (\pgf@pathmaxx,\pgf@pathmaxy)}
	},
	reset transform/.code={\pgftransformreset}
}
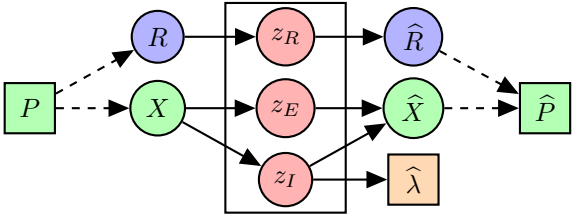
\begin{figure}
	\centering
	\begin{tikzpicture}[auto, thick, node distance=1.3cm, >=triangle 45] 
	\draw 
	node at (0,0)[block,name=p,fill=green!30] {$P$}
	node [fm,right of=p,fill=green!30] (x) {$X$}
	node [fm2,above of=x,node distance=0.95cm,fill=blue!30] (r) {$R$}
	node [fm,right of=x,fill=red!30] (ze) {$z_E$}
	node [fm,below of=ze,node distance=0.95cm,fill=red!30] (zi) {$z_I$}
	node [fm,right of=r,fill=red!30] (zr) {$z_R$}
	node [fm,right of=zr,fill=blue!30] (rhat) {$\widehat{R}$}
	node [fm2,right of=ze,fill=green!30] (xhat) {$\widehat{X}$}
	node [block,right of=zi,node distance=1.7cm,fill=orange!30] (lambda) {$\widehat{\lambda}$}
	node [block,right of=xhat,fill=green!30,node distance=1.75cm] (phat) {$\widehat{P}$};
	\draw[->,dashed](p) -- (r);
	\draw[->,dashed](p) -- (x);
	\draw[->](x) -- (ze);
	\draw[->](x) -- (zi);
	\draw[->](r) -- (zr);
	\draw[->](zr) -- (rhat);
	\draw[->,dashed](rhat) -- (phat);
	\draw[->,dashed](xhat) -- (phat);
	\draw[->](ze) -- (xhat);
	\draw[->](zi) -- (xhat);
	\draw[->](zi) -- (lambda);
	\draw (2.6, -1.39) rectangle (4.2, 1.4);
	\end{tikzpicture}
	\caption{
		A schematic overview of the combined two-level architecture used as the generative model.
		A point cloud $P$ is first encoded into $(R,X)$ by a \emph{deterministic} AE based on PointNet,
		$R$ being the quaternion representing the rotation of the shape, and $X$ the compressed 
		representation of the input shape.
		$(R,X)$ is then further compressed into a latent representation $z=(z_R,z_E,z_I)$ of a VAE.
		The hierarchical latent variable $z$ has disentangled subgroups in red 
		(representing rotation, extrinsics, and intrinsics, respectively).
		The intrinsic latent subgroup $z_I$ is used to predict the LBO spectrum $\hat{\lambda}$.
		Both the extrinsic $z_E$ and intrinsic $z_I$ are utilized to compute the shape $\hat{X}$ in the AE's latent space.
		The latent rotation $z_R$ is used to predict the quaternion $\hat{R}$.
		Finally, the decoded representation $(\hat{R},\hat{X})$ is used to reconstruct the original point cloud $\hat{P}$.
		The deterministic AE mappings are shown as dashed lines; 
		VAE mappings are represented by solid lines.
	}
	\label{fig:desc}
\end{figure}

Similar to prior work~\cite{achlioptas2018learning}, we utilize a two-level 
architecture, where the VAE is learned on the latent space of an AE.
This architecture is shown in~\reffig{fig:desc}.
Throughout this work, we use the following notation: 
$P$ denotes a point cloud, $(R,X)$ is the latent AE representation, 
and $\hat{P}$ is the reconstructed point cloud.
Although rotation is a strictly extrinsic transformation, we separate them because 
(1) rotation is intuitively different than other forms of non-rigid extrinsic pose (e.g., articulation),
(2) having separate control over rotations is commonly desirable in applications 
(e.g., \cite{kazhdan2003rotation,farfade2015multi}), 
and
(3) our quaternion-based factorization provides a straightforward way to do so.

\subsection{Point Cloud Losses} \label{sec:ae:losses}
	Following previous work on point cloud AEs~\cite{achlioptas2018learning,li2018so,dovrat2019learning}, 
	we utilize a form of the Chamfer distance as our main measure of similarity.
	We define the max-average function 
	\begin{equation}
		M_\alpha(\ell_1,\ell_2) = \alpha\max\{\ell_1,\ell_2\} + (1-\alpha)(\ell_1 + \ell_2)/2,
		\label{eq:maxaverage} 
	\end{equation}
	where $\alpha$ is a hyper-parameter that controls the relative weight of the two values.
	It is useful to weight the larger of the two terms higher, 
	so that the network does not focus on only one term \cite{yang2018foldingnet}.
	We then use the point cloud loss
	\begin{equation}
		\mathcal{L}_C = M_{\alpha_C}\left( 
		\frac{1}{|P|}\sum_{p\in P} 
		\widehat{d}(p) ,
		\frac{1}{|\hat{P}|}\sum_{\hat{p}\in \hat{P}} 
		d(\widehat{p})
		\right),
		\label{eq:max_average_loss}
	\end{equation}
	where 
	$d(\widehat{p}) = \min_{ {p}\in {P}} || p - \hat{p} ||_2^2 $ and 
	$\widehat{d}(p) = \min_{\hat{p}\in\hat{P}} || p - \hat{p} ||_2^2 $.
	In an effort to reduce outliers, we add a second term, 
	as a form of approximate Hausdorff loss:
	\begin{equation}
		\mathcal{L}_H = M_{\alpha_H}\left( 
		\max_{p\in P} 
		d(\widehat{p}),
		\max_{\hat{p}\in \hat{P}} 
		\widehat{d}(p)
		\right). 		
		\label{eq:hausdorff_loss}
	\end{equation}
	The final reconstruction loss is therefore
	$
	\mathcal{L}_R = r_C \mathcal{L}_C + r_H \mathcal{L}_H
	$
	for constants $r_C,r_H$. 

\subsection{Quaternionic Rotation Representation} \label{sec:ae:quat}
	We make use of quaternions to represent rotation in the AE model.
	The unit quaternions form a double cover of the rotation group $SO(3)$~\cite{huynh2009metrics}; 
	hence, any vector $R\in\real^4$ can be converted to a rotation via normalization. 
	We can then differentiably convert any such quaternion $R$ to a rotation 
	matrix $R_M$. 
	To take the topology of $SO(3)$ into account, 
	we use the distance metric~\cite{huynh2009metrics}
	$ 
	\mathcal{L}_Q = 1 - |q\cdot\widetilde{q}| 
	$
	between unit quaternions $q$ and $\widetilde{q}$. 

\subsection{Autoencoder Model} \label{sec:ae:model}
	The encoding function 
	$ f_E(P) = (R,X) $ 
	maps a point cloud $P$ to a vector $(R,X)\in \real^{D_A}$, 
	which is partitioned into a quaternion $R$ 
	(representing the rotation) 
	and a vector $X$, which is a compressed representation of the shape. 
	The mapping is performed by a PointNet model~\cite{qi2017pointnet}, 
	followed by fully connected (FC) layers.
	The decoding function works by rotating the decoded shape vector: 
	$ f_D(R,X) = g_D(X) R_M = \hat{P} $, 
	where $g_D$ was implemented via FC layers and $R_M$ is the matrix form of $R$.
	The loss function for the autoencoder is the reconstruction loss $\mathcal{L}_R$.

	Note that the input can be a point cloud of arbitrary size, but the output 
	is of fixed size, and is determined by the final network layer 
	(though alternative architectures could be dropped in to avoid this 
	limitation~\cite{li2018point,groueix2018atlasnet}).
	Our data augmentation scheme during training consists of random rotations of 
	the data about the height axis, and using randomly sampled points 
	from the shape as input (see~\refsec{sec:experiments}). 
	For architectural details, see Supplementary Material.

\section{Geometrically Disentangled VAE} \label{sec:vae}
Our generative model, the geometrically disentangled VAE (GDVAE), is defined on top of the latent space of the AE; 
in other words, 
it encodes and decodes between its own latent space 
(denoted $z$) 
and that of the AE 
(i.e., $(R,X)$).
The latent space of the VAE is represented by a vector that is hierarchically 
decomposed into sub-parts, $z = (z_R, z_E, z_I)$,
representing the rotational, extrinsic, and intrinsic components, respectively.
In addition to reconstruction loss, we define the following loss terms:
(1) a probabilistic loss 
that matches the latent encoder distribution to the prior $p(z)$, 
(2) a spectral loss, 
which trains a network to map $z_I$ to a spectrum $\lambda$,
and
(3) a disentanglement loss 
that penalizes the sharing of information between $z_I$ and $z_E$ in the latent space.
Note that the first (1) and third (3) terms are based on the 
Hierarchically Factorized VAE (HFVAE) defined by Esmaeili et al.~\cite{esmaeili2018structured}, 
but the third term also includes a covariance penalty motivated by the 
Disentangled Inferred Prior VAE (DIP-VAE)~\cite{kumar2017variational} and another 
penalty based on the Jacobian between latent subgroups.
In the next sections, we discuss each term in more detail.

\subsection{Latent Disentanglement Penalties} \label{sec:vae:penalties}
	To disentangle intrinsic and extrinsic geometry in the latent space, we 
	consider three different hierarchical penalties.
	In this section, we define the latent space $z$ to consist of $|G|$ subgroups, 
	i.e., $z=(z_1,\ldots,z_{|G|})$, with each subset $z_i$ being a vector-valued 
	variable of length $g_i$.
	We wish to disentangle each subgroup from all the others. 
	In this work, $z=(z_R,z_E,z_I)$ and $|G|=3$.

	\paragraph{Hierarchically Factorized Variational Autoencoder.}
	Recent work by Esmaeili et al.~\cite{esmaeili2018structured}
	showed that the prior-matching term of the VAE objective (i.e., 
	$\dkl\left[q_\phi(z|x) \,\middle|\middle|\, p(z)\right]$) 
	can be hierarchically decomposed as
	\begin{equation}
		\mathcal{L}_\text{HF}
		=
		\beta_1 P_{\text{intra}} 
		+
		\beta_2 \, P_{\text{KL}} 
		+
		\beta_3 \, \mathcal{I}[x;z]
		+
		\beta_4 \, TC(z), 
	\end{equation}
	where 
	$TC(z)$ is the inter-group TC, 
	$\mathcal{I}[x;z]$ is the mutual information between the data and its latent 
	representation, and $P_{\text{intra}}$ and $P_{\text{KL}}$ are the intra-group 
	TC and dimension-wise KL-divergence, respectively, given by the following formulas: 
	$ P_{\text{intra}} = \sum_g TC(z_g) $
	and
	$ P_{\text{KL}} 	 = 
	\sum_{g,d} 
	\dkl[ q_\phi(z_{g,d}) \,||\, p(z_{g,d}) ] $.

	As far as disentanglement is concerned,
	the main term enforcing inter-group independence (via the TC) is the one 
	weighted by $\beta_4$.
	However, note that the other terms are essential for matching the latent 
	distribution to the prior $p(z)$, 
	which allows generative sampling from the network. 
	We use the implementation in ProbTorch~\cite{siddharth2017learning}. 

	\paragraph{Hierarchical Covariance Penalty.}
	A straightforward measure of statistical dependence is covariance.
	While this is only a measure of the linear dependence between variables, unlike the information-theoretic penalty considered above, vanishing covariance is still necessary for disentanglement. 
	Hence, we consider a covariance-based penalty to enforce independence between variable groups.
	This is motivated by Kumar et al.\ \cite{kumar2017variational}, 
	who discuss how
	disentanglement can be better controlled by introducing a penalty
	that moment-matches the inferred prior $q_\phi(z)$ to the latent prior $p(z)$.
	We perform a simple alteration to make this penalty hierarchical.
	Specifically, 
	let $\hat{C}$ denote the estimated covariance matrix over the batch
	and
	recall that $q_\phi(z|x) = \normd(z|\mu_\phi(x),\Sigma_\phi(x))$.
	Finally, denote
	$ \mu_g $ as the part of $\mu_\phi(x)$ corresponding to group $g$
	(i.e., parameterizing the approximate posterior over $z_g$)
	and define 
	\begin{equation}
		\mathcal{L}_{\text{COV}} = \gamma_I
		\sum_{g\ne \widetilde{g}} \sum_{i, j} 
		\left| \hat{C}(\mu_{g}, \mu_{\widetilde{g}})_{ij} \right|
		\label{eq:hierarchical_covariance_penalty}
	\end{equation}
	as a penalty on inter-group covariance, 
	where the first sum is taken over all non-identical pairings. 
	We ignore the additional moment-matching penalties on the diagonal and intra-group covariance from \cite{kumar2017variational}, since they are not related to intrinsic-extrinsic disentanglement and a prior-matching term is already present within $\mathcal{L}_\text{HF}$.


	\begin{figure}
		\centering
		\begin{tikzpicture}[auto, thick, node distance=1.3cm, >=triangle 45] 
		\definecolor{mucolor}{RGB}{51, 249, 255}
		\draw 
		node at (0,0)[fm,name=x,fill=green!30] {$X$}
		node [fm,above right of=x,node distance=1.7cm,yshift=-0.5cm,fill=mucolor!30] (muE) {$\mu_E$}
		node [fm,below right of=x,node distance=1.7cm,yshift=0.5cm,fill=mucolor!30] (muI) {$\mu_I$}
		node [fm,right of=muE,fill=red!30] (zE) {$z_E$}
		node [fm,right of=muI,fill=red!30] (zI) {$z_I$}
		node [fm,right of=x,node distance=4cm,fill=green!30] (xhat) {$\hat{X}$}
		node [fm,above right of=xhat,node distance=1.7cm,yshift=-0.5cm,fill=mucolor!30] (muEhat) {$\hat{\mu}_E$}
		node [fm,below right of=xhat,node distance=1.7cm,yshift=0.5cm,fill=mucolor!30] (muIhat) {$\hat{\mu}_I$}
		;
		\draw[->](x) -- (muE);
		\draw[->](x) -- (muI);
		\draw[->]([yshift= -3pt] muE.east) -- ([yshift= -3pt] zE.west);
		\draw[dashed,red,line width=0.5mm]([yshift= 3pt] muE.east) -- ([yshift= 3pt] zE.west);
		\draw[dashed,blue,line width=0.5mm] ([yshift=-3pt] muI.east) -- ([yshift=-3pt] zI.west);
		\draw[->] ([yshift=3pt] muI.east) -- ([yshift=3pt] zI.west);
		\draw[dashed,red,line width=0.5mm](zE.-15) -- (xhat.130);
		\draw[->]    					  (zE.-50) -- (xhat.165);
		\draw[dashed,blue,line width=0.5mm](zI.15) -- (xhat.230);
		\draw[->]    					   (zI.50) -- (xhat.195);
		\draw[dashed,red,line width=0.5mm](xhat.-15) -- (muIhat.130);
		\draw[->]    					  (xhat.-50) -- (muIhat.165);
		\draw[dashed,blue,line width=0.5mm](xhat.15) -- (muEhat.230);
		\draw[->]    					(xhat.50) -- (muEhat.195);	
		\end{tikzpicture}
		\caption{
			Diagram of the pairwise Jacobian norm penalty computation within a VAE.
			The red and blue dashed paths show the computation graph paths 
			utilized to compute the Jacobians.
		}
		\label{fig:descJacobian}
	\end{figure}
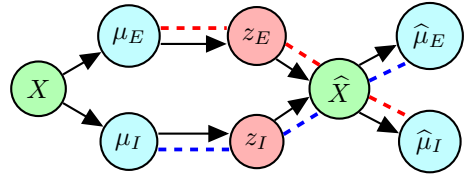

	\paragraph{Pairwise Jacobian Norm Penalty.}
	Finally, we follow the intuition that changing the value 
	of one latent group should not affect the expected value of any other group.
	We derive a loss term for this by considering how the variables change 
	if the decoded shape is re-encoded into the latent space.
	This approach to geometric disentanglement is visualized in~\reffig{fig:descJacobian}.
	Unlike the TC and covariance-based penalties, this does not disentangle 
	$z_R$ from $z_E$ and $z_I$. 

	Formally, we consider the Jacobian of a latent group with respect to another.
	The norm of this Jacobian can be viewed as a measure of how much 
	one latent group can affect another group, through the decoder. 
	This measure is
	\begin{equation}
		\mathcal{L}_J = 
		\max_{g\ne \widetilde{g}} 
		\left|
		\left|
		\frac{\partial \hat{\mu}_g}{\partial \mu_{\widetilde{g}}}
		\right|
		\right|^2_F,
		\label{eq:jacobian_norm_penalty}		
	\end{equation}
	where 
	$\hat{X}$ is the decoded shape,
	$\hat{\mu}_g$ represents group $g$ from $ \mu_\phi(\hat{X})$, 
	and 
	we take the maximum over pairs of groups.

\subsection{Spectral Loss} \label{sec:vae:spectral}
    Mathematically, the intrinsic differential geometry of a shape can be viewed as 
    those properties dependent only on the metric tensor, i.e., independent of the
    embedding of the shape~\cite{corman2017functional}.
    Such properties depend only on geodesic distances on the shape rather than how 
    the shape sits in the ambient 3D space.
    The Laplace-Beltrami operator (LBO) is a popular way of capturing intrinsic shape.
    Its spectrum $\lambda$ can be formally described by viewing a shape as a 2D Riemannian manifold
    $(\mathcal{M},g)$ embedded in 3D, with point clouds being viewed as random samplings from this surface.
	
	Given the spectrum $\lambda$ of a shape, we wish to compute a loss with 
	respect to a predicted spectrum $\hat{\lambda}$, treating each as a vector with $N_\lambda$ elements.
	The LBO spectrum has a very specific structure, with 
	$ \lambda_i\geq 0\;\forall\;i $ and $\lambda_j\geq\lambda_k\;\forall\;j>k$.
	Analogous to frequency-space signal processing, 
	larger elements of $\lambda$ correspond to ``higher frequency'' properties 
	of the shape itself: i.e., finer geometric details, as opposed to coarse overall shape.
	This analogy can be formalized by the ``manifold harmonic transform'', a 
	direct  generalization of the Fourier transform to non-Euclidean domains based on the LBO \cite{vallet2008spectral}.
	Due to this structure, a naive vector space loss function on $\lambda$ 
	(e.g., $L_2$) 
	will over-weight learning the higher frequency elements of the spectrum. 
	We suggest that the lower portions of $\lambda$ not be down-weighted, as 
	they are less susceptible to noise and convey larger-scale, 
	``low-frequency'' global information about the shape, which is more useful 
	for coarser shape reconstruction.

	Given this, we design a loss function that avoids over-weighting the higher 
	frequency end of the spectrum:
	\begin{equation}
		\mathcal{L}_S(\lambda,\hat{\lambda})
		= 
		\frac{1}{N_\lambda} 
		\sum_{i=1}^{N_\lambda} 
		\frac{|\lambda_i - \hat{\lambda}_i|}{i},
		\label{eq:spectral_loss}
	\end{equation}
	where the use of the $L_1$ norm and the linearly increasing element-wise 
	weight of $i$ decrease the disproportionate effect of the larger magnitudes 
	at the higher end of the spectrum.
	The use of linear weights is theoretically motivated by Weyl's law 
	(e.g., \cite{reuter2006laplace}), which asserts that spectrum elements 
	increase approximately linearly, for large $i$.

\subsection{VAE Model} \label{sec:vae:model}
	Essentially, the latent space is divided into three parts, for rotational, 
	extrinsic, and intrinsic geometry, denoted $z_R$, $z_E$, and $z_I$, respectively. 
	We note that, while rotation is fundamentally extrinsic, we can take 
	advantage of the AE's decomposed representation to define $z_R$ on the AE 
	latent space over $R$, and use $z_E$ and $z_I$ for $X$. 
	The encoder model can be written as
	$
	(z_E,z_I) = \mu_\phi(X) + \Sigma_\phi(X)\xi 
	$, 
	where $\xi\sim\normd(0,I)$, 
	while the decoder is written
	$ \hat{X} = h_D(z_E,z_I) $. 
	A separate encoder-decoder pair is used for $R$.
	The spectrum is predicted from the latent intrinsics alone:
	$ \hat{\lambda} = f_S(z_I) $.

	The reconstruction loss, used to compute the log-likelihood,
	is given by the combination of the quaternion metric and a Euclidean loss 
	between the vector representation of the (compressed) shape and its reconstruction:
	\begin{equation}
		\mathcal{L}_V = \frac{1}{D}||X - \hat{X}||^2_2 + w_Q \mathcal{L}_Q,
		\label{eq:reconstruction_loss}
	\end{equation}
	where $ \mathcal{L}_Q $ is the metric over quaternion rotations and $D=\dim(X)$.
	We now define the overall VAE loss:
	\begin{equation}
		\mathfrak{L} = \eta \mathcal{L}_V 
		+ \mathcal{L}_\text{HF}
		+ \mathcal{L}_\text{COV}
		+ w_J \mathcal{L}_J
		+ \zeta \mathcal{L}_S.
		\label{eq:vae_loss}
	\end{equation}
	The VAE needs to be able to 
	(1) autoencode shapes, 
	(2) sample novel shapes, and 
	(3) disentangle latent groups. 
	The first term of $\mathfrak{L}$ encourages (1), while the second term enables (2); 
	the last four terms of $\mathfrak{L}$ contribute to task (3).
\section{Experiments} \label{sec:experiments}

For our experiments, we consider four datasets of meshes: 
shapes computed from the MNIST dataset~\cite{lecun1998gradient}, 
the MPI Dyna dataset of human shapes~\cite{dyna}, 
a dataset of animal shapes from the Skinned Multi-Animal Linear model 
(SMAL)~\cite{Zuffi:CVPR:2017}, and a dataset of human shapes from the 
Skinned Multi-Person Linear model (SMPL)~\cite{SMPL:2015} via the SURREAL 
dataset~\cite{varol17_surreal}. 
For each, we generate point clouds of size $N_T$ via area-weighted sampling. 

For SMAL and SMPL we generate data from 3D models using a 
modified version of the approach in Groueix et al.~\cite{groueix2018atlasnet}.
During training, the input of the network is a uniformly random subset of 
$N_S$ points from the original point cloud.	
We defer to the Supplemental Material for details concerning dataset 
processing and generation.

We compute the LBO spectra directly from the triangular meshes using the 
cotangent weights formulation~\cite{meyer2003discrete}, as it provides a more 
reliable result than algorithms utilizing point clouds 
(e.g., \cite{belkin2009constructing}). 
We thus obtain a spectrum $\lambda$ as a $N_\lambda-$dimensional vector, 
associated with each shape.
We note that our algorithm requires only a point cloud as input data 
(or a Gaussian random vector, if generating samples). 
LBO spectra are utilized only at training time, while triangle meshes are 
used only for training set generation.
Hence, our method remains applicable to pure point cloud data.

\subsection{Generative Shape Modeling} \label{sec:experiments:generative}

 	\begin{figure}
 		\centering
 		\includegraphics[width=0.4797\textwidth]{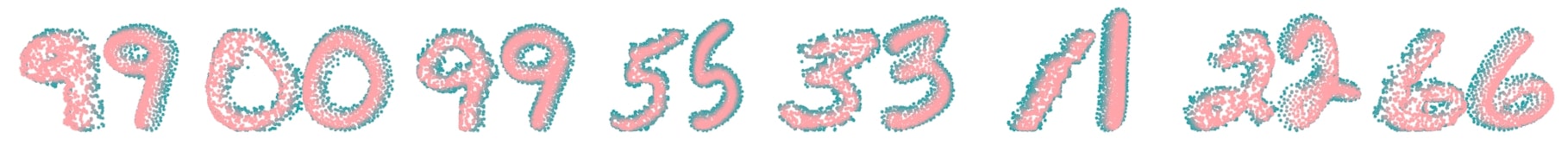} 
 		\includegraphics[width=0.4797\textwidth]{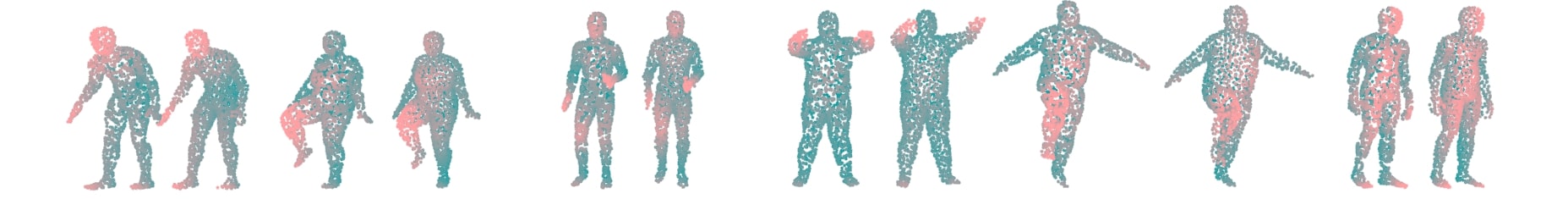} %
 		\includegraphics[width=0.4797\textwidth]{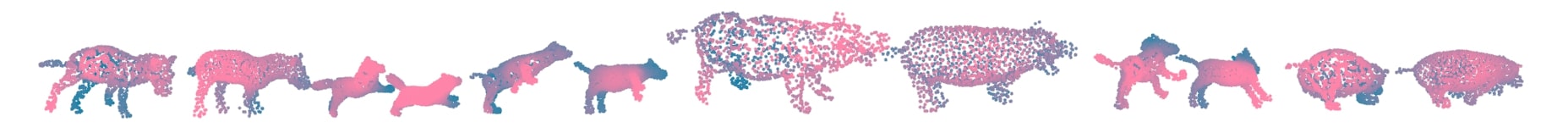} %
 		\includegraphics[width=0.47947\textwidth]{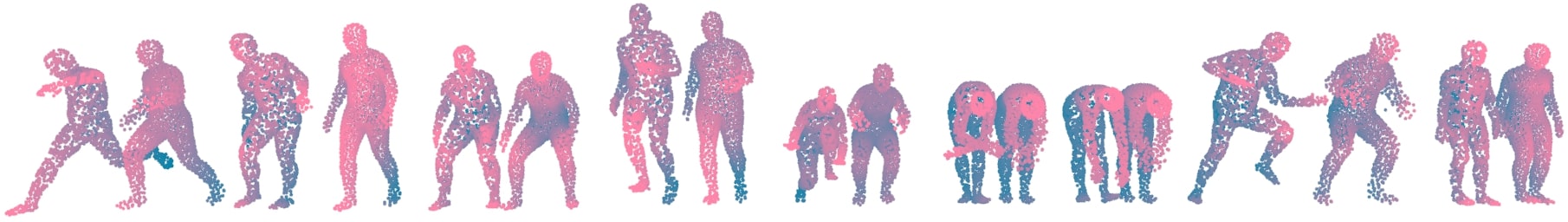} %
 		\caption{\textbf{Reconstructions} of random samples, passed through both the AE and VAE. 
 		    For each pair, the left shape is the input and the right shape is the reconstruction. 
 			Colors denote depth (i.e., distance from the camera). Rows: MNIST, Dyna, SMAL, SMPL.
 		}
 		\label{fig:recon}
 	\end{figure}

 	\begin{figure}
 		\centering
 		\includegraphics[width=0.48197\textwidth]{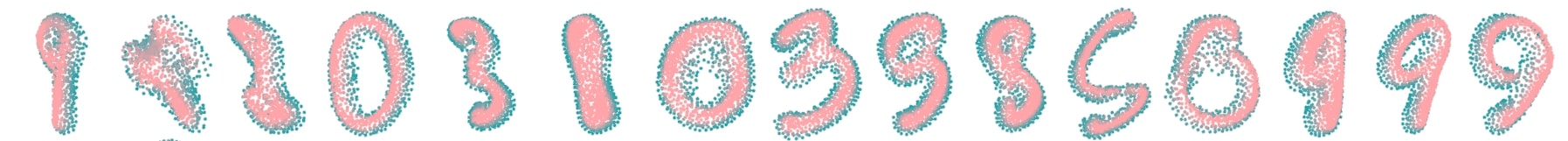} 
 		\includegraphics[width=0.48197\textwidth]{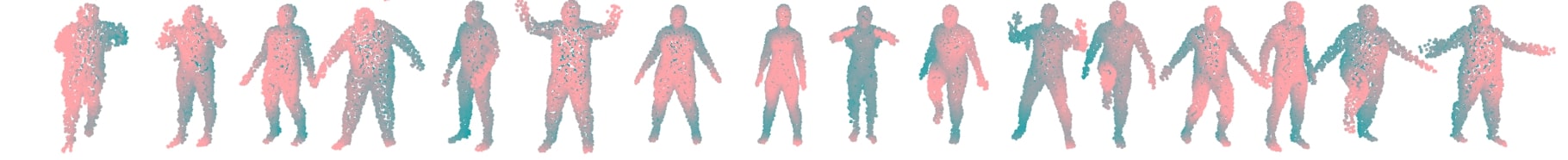} %
 		\includegraphics[width=0.48197\textwidth]{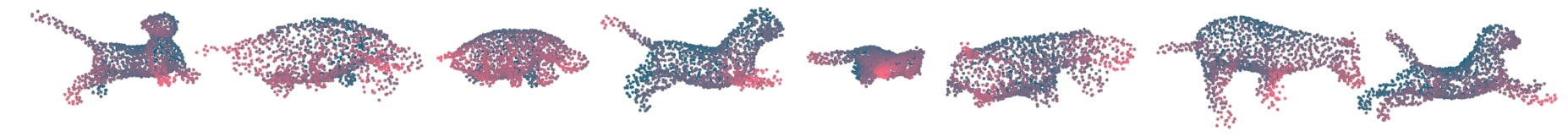} %
 		\includegraphics[width=0.481927\textwidth]{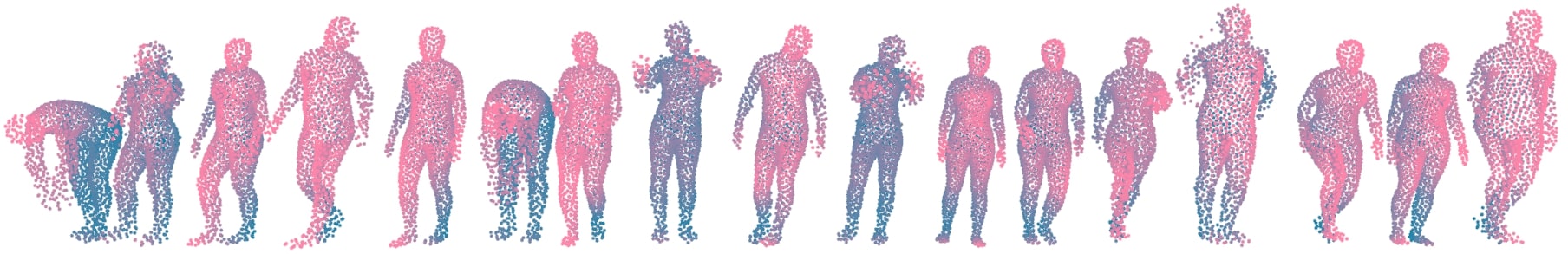} %
 		\caption{\textbf{Samples drawn} from the latent space of the VAE by decoding $z\sim\normd(0,I)$ with $z_R = 0$. Colors denote depth (i.e., distance from the camera). Rows: MNIST, Dyna, SMAL, SMPL.}
 		\label{fig:sample}
 	\end{figure}

	Ideally, our model should be able to disentangle intrinsic and extrinsic 
	geometry without losing its capacity to 
		(1) reconstruct point clouds 
		and 
		(2) generate random shape samples.  
	We show qualitative reconstruction results in Figure~\ref{fig:recon}. 
	Considering the latent dimensionalities 
	($|z_E|, |z_I|$ are 5, 5; 10, 10; 8, 5; and 12, 5, for MNIST, Dyna, SMAL, and SMPL, respectively),
	it is clear that the model is capable of 
	decoding from significant compression. 
	However, thin or protruding areas (e.g., hands or legs) have a lower point density (a known problem with the Chamfer distance~\cite{achlioptas2018learning}). 

	We also consider the capacity of the model to generate novel shapes from 
	randomly sampled latent $z$ values, as shown in~\reffig{fig:sample}.
	We can see a diversity of shapes and poses; 
	however, not all samples belong to the data distribution (e.g., invalid MNIST samples, or extra protrusions from human shapes).
	VAEs are known to generate blurry
	images~\cite{tomczak2017vae,larsen2015autoencoding};
	in our case, ``blurring'' implies a perturbation in the latent space, 
	rather than in the 3D point positions, 
	explaining the unintuitive artifacts in Figures~\ref{fig:recon} and \ref{fig:sample}.

	A standard evaluation method in generative modeling is testing the usefulness of the representation in downstream tasks (e.g., \cite{achlioptas2018learning}).
	This is also useful for illustrating the role of the latent disentanglement. 
	As such, we utilize our encodings for classification on MNIST, recalling that our representation was learned without access to the labels.
	To do so, we train a linear support vector classifier (from scikit-learn \cite{scikit-learn}, with default parameters and no data augmentation) on the parts of the latent space defined by the GDVAE (see Table \ref{table:mnistAcc}).
	Comparing the drop from $S=(R,X)$ to $z$ shows the effect of compression and KL regularization; we can also see that $z_R$ is the least useful component, but that it still performs better than chance, suggesting a correlation between digit identity and the orientation encoded by the network. 
    In the Supplemental Material, we include confusion matrices showing that mistakes on $z_I$ or $(z_R,z_I)$ 
    are similar to those incurred when using $\lambda$ directly.

	\setlength{\tabcolsep}{0.5em}
	\begin{table}[t]
		\centering
		\begin{tabular}{cccccccc}
			$z_R$ & $z_E$ & $z_I$ & $z_{RE}$ & $z_{RI}$ & $z_{EI}$ & $z$ & $S$ \\
			\hline 
			$ 0.32 $ & $ 0.47 $ & $ 0.60 $ & $ 0.64 $ & $ 0.68 $ & $ 0.88 $ & $ 0.88 $ & $ 0.98 $ \\
		\end{tabular}
		\caption{Accuracies of a linear classifier on various segments of the latent space from the MNIST test set. We denote $z_{RE}=(z_R,z_E)$, $z_{RI}=(z_R,z_I)$, $z_{EI}=(z_E,z_I)$, and $S=(R,X)$.}
		\label{table:mnistAcc}
	\end{table}
    		
    Lastly, our AE naturally disentangles rigid pose (rotation) and the rest of the representation.
    Ideally, the network would not learn disparate $X$ representations for a single shape under rotation; rather, it should map them to the same shape representation, with a different accompanying quaternion.
    This would allow rigid pose normalization via derotations: for instance, rigid alignment of shapes could be done by matching $z_R$, which could be useful for pose normalizing 3D data.
    We found that the model is robust to small rotations, but it often learns separate representations under larger rotations (see Supplemental Material).
    In some cases, this may be unavoidable (e.g., for MNIST, 9 and 6 are often indistinguishable after a 
    180\textdegree\  rotation).
    	
    \begin{figure*}[h]
    	\includegraphics[height=1.937in]{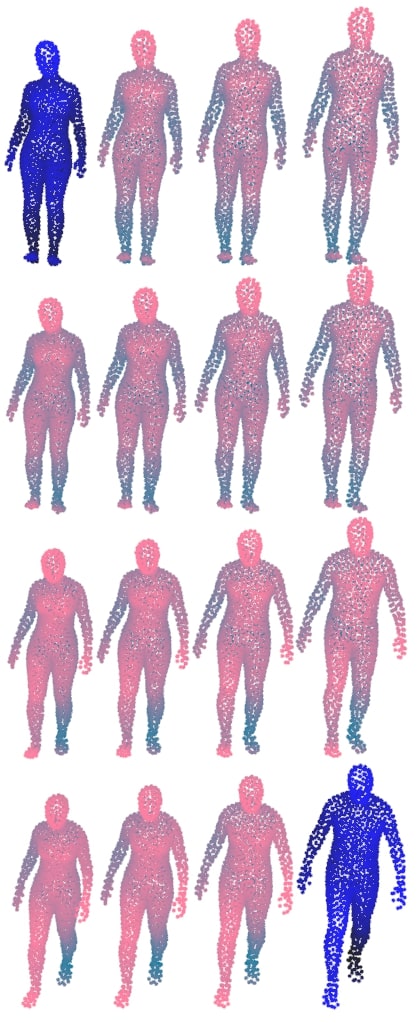} \hfill
    	\includegraphics[height=1.937in]{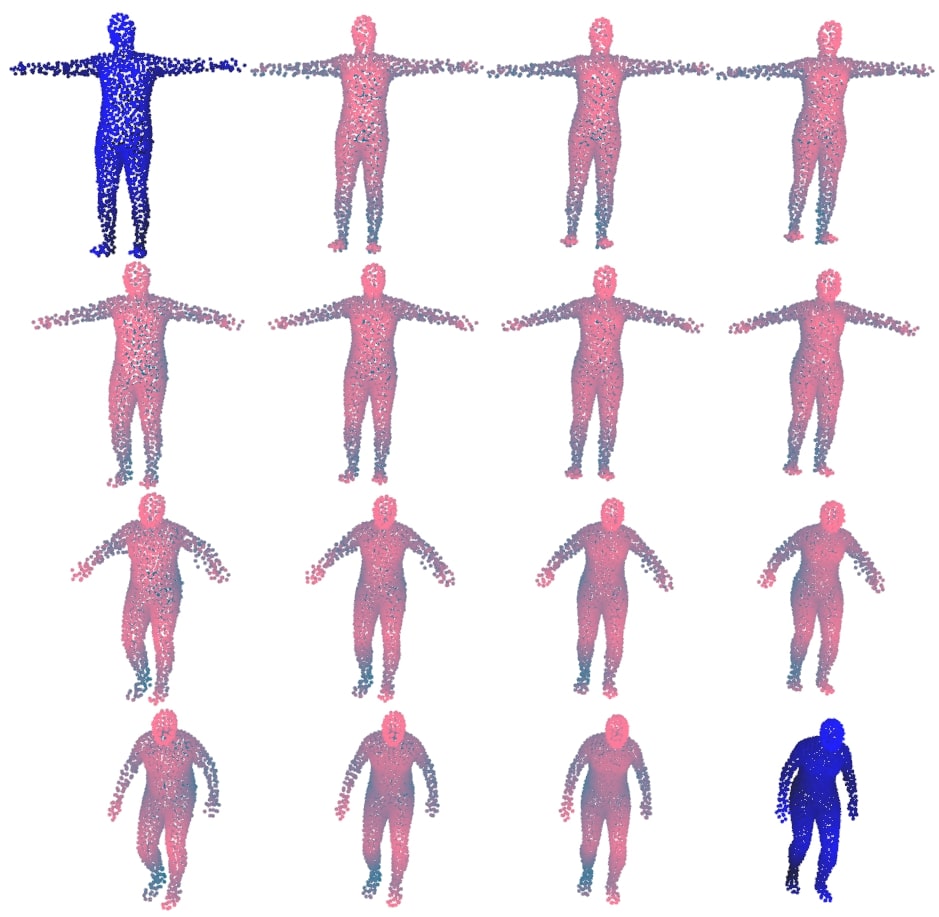} \hfill
    	\includegraphics[height=1.937in]{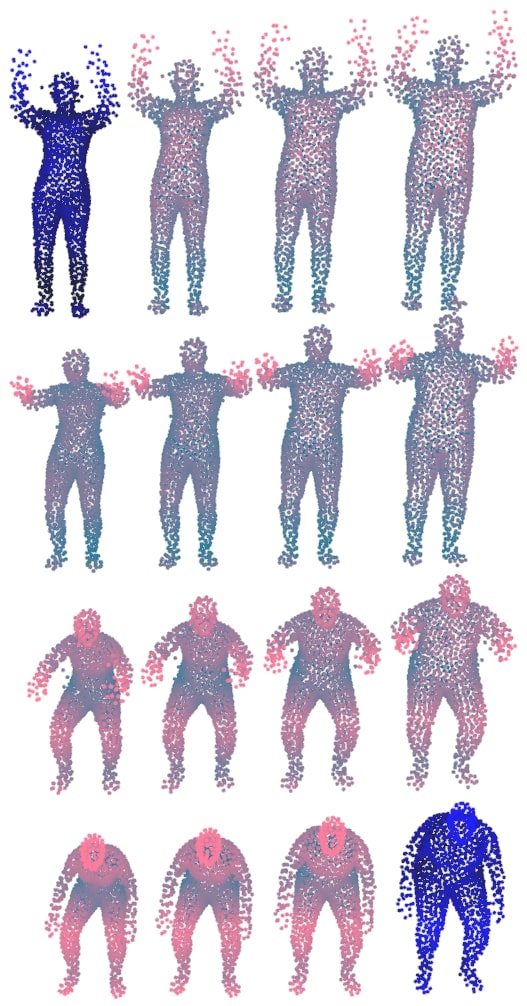} \hfill
    	\includegraphics[height=1.937in]{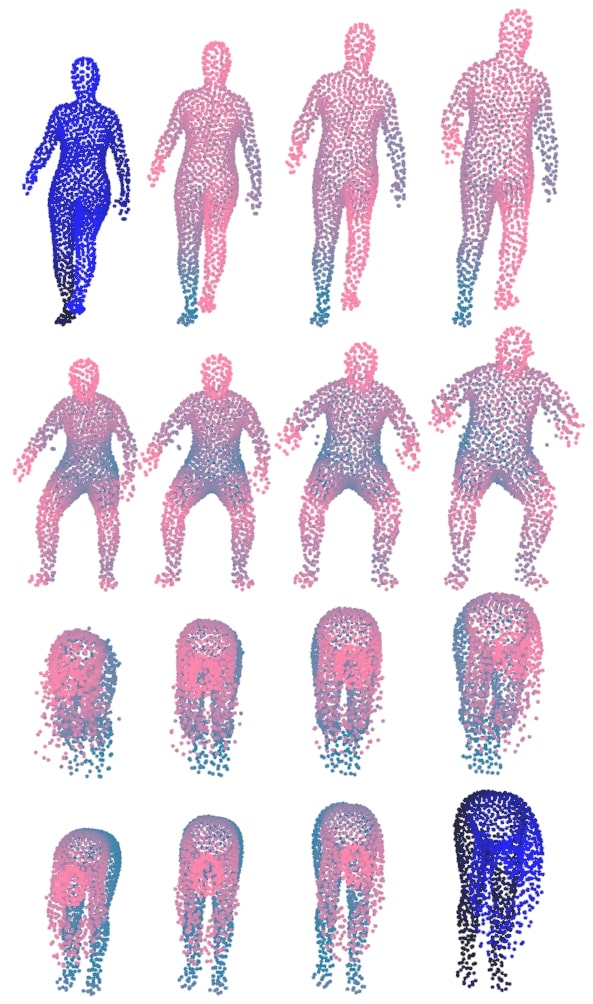}\hfill
    	\includegraphics[height=1.937in]{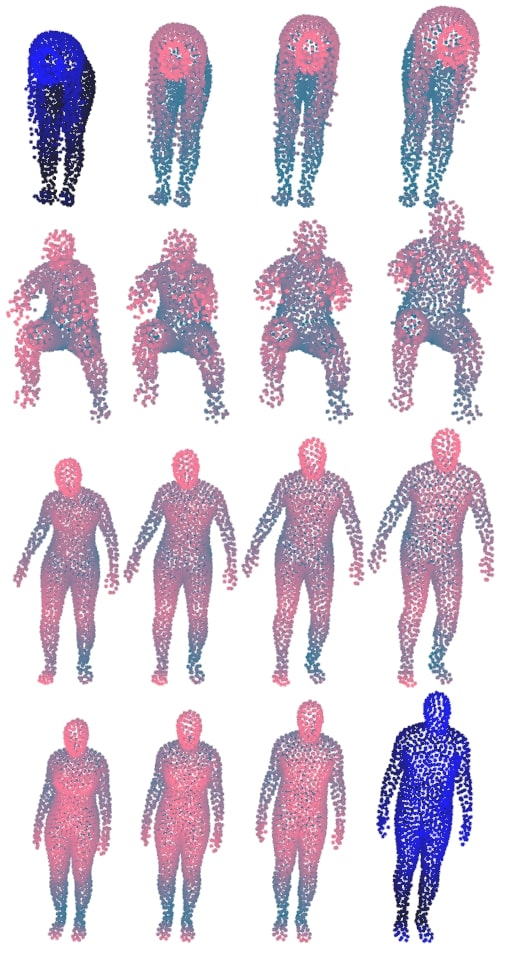} \\
    	\includegraphics[height=1.0795in]{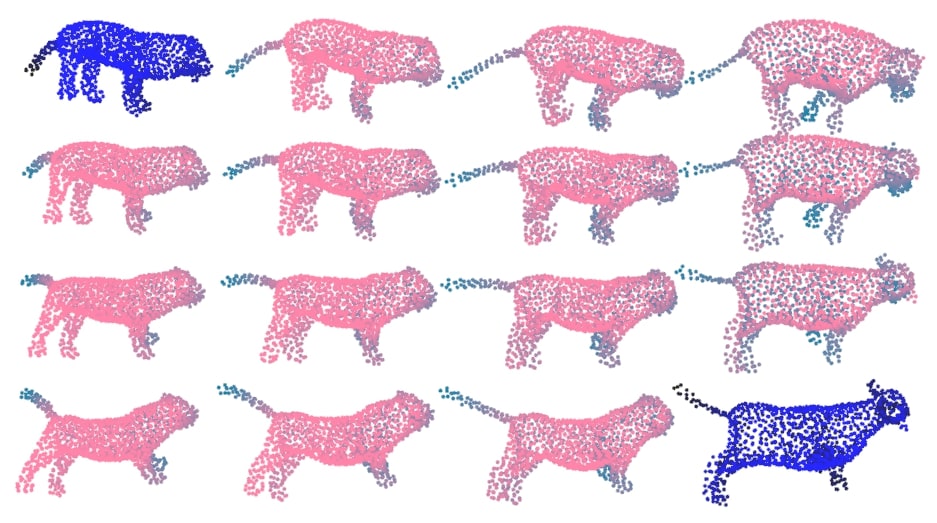} \hfill
    	\includegraphics[height=1.0795in]{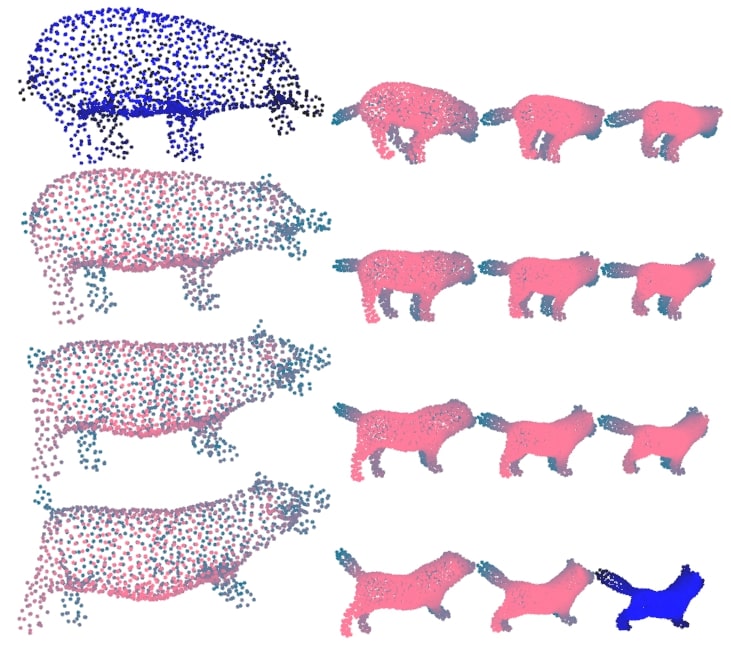} \hfill
    	\includegraphics[height=1.0795in]{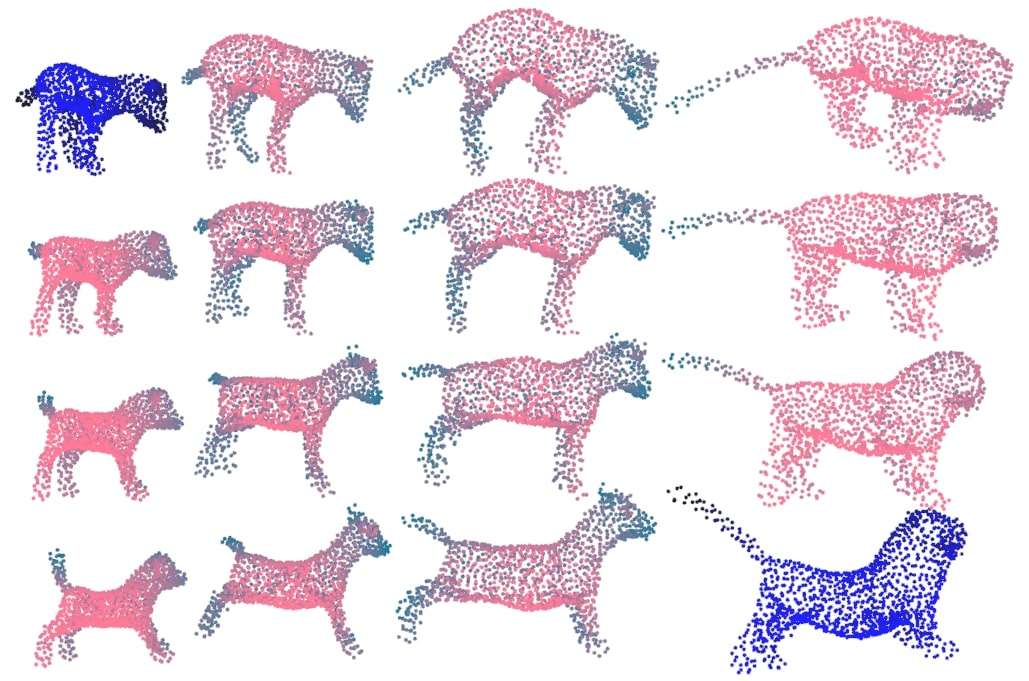} \hfill
    	\includegraphics[height=1.0795in]{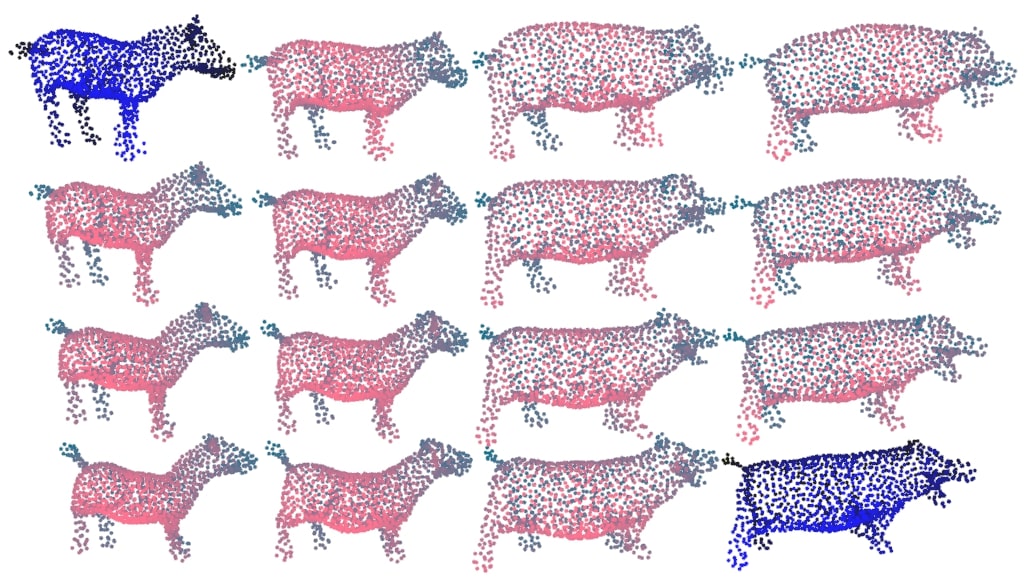}
    	\caption{\textbf{Latent space interpolations} between SMPL (row 1) and SMAL (row 2) shapes. 
    		Each inset interpolates $z$ between the upper-left and lower right shapes, with $z_E$ changing along the vertical axis and $z_I$ changing along the horizontal one.
    		Per-shape colours denote depth.
    	}
    	\label{fig:interp}
    \end{figure*}
    
    \subsection{Disentangled Latent Shape Manipulation}	
    
    \begin{figure}
    	\centering
    	\begin{tabular}{l|c|c|r}
    		\includegraphics[height=1.31in]{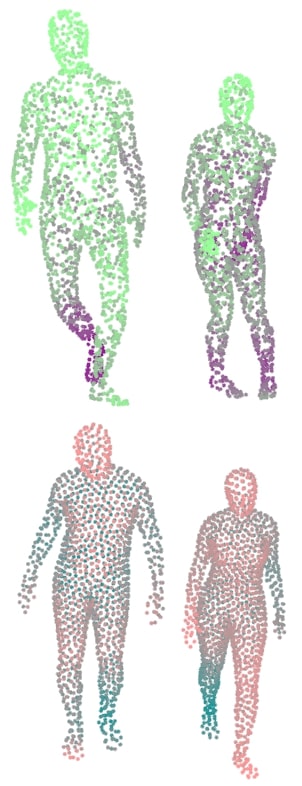} &
    		\includegraphics[height=1.31in]{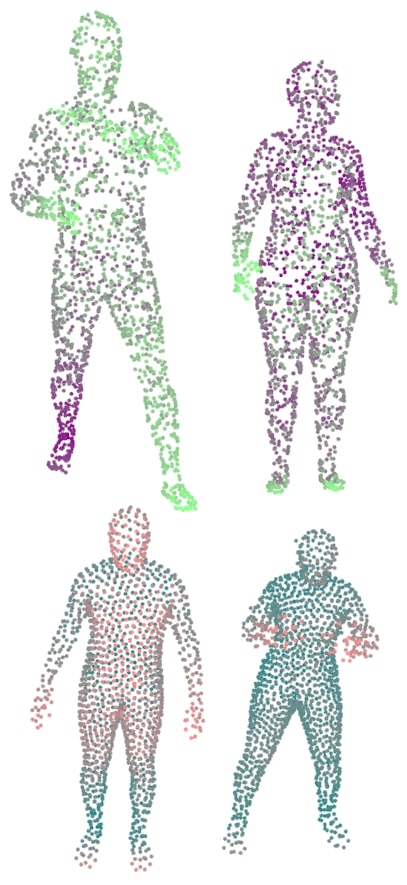} &
    		\includegraphics[height=1.31in]{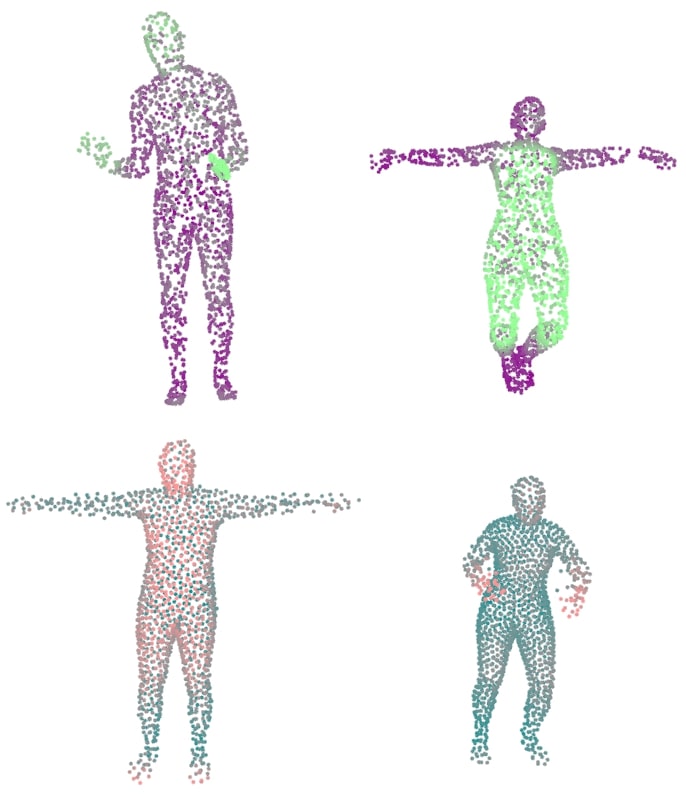} &
    		\includegraphics[height=1.31in]{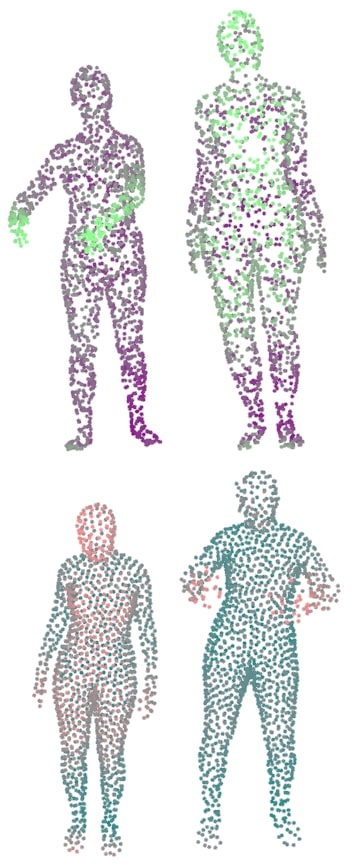} 
    	\end{tabular} \\
    	\begin{tabular}{l|c|r}
    		\includegraphics[height=0.5275in]{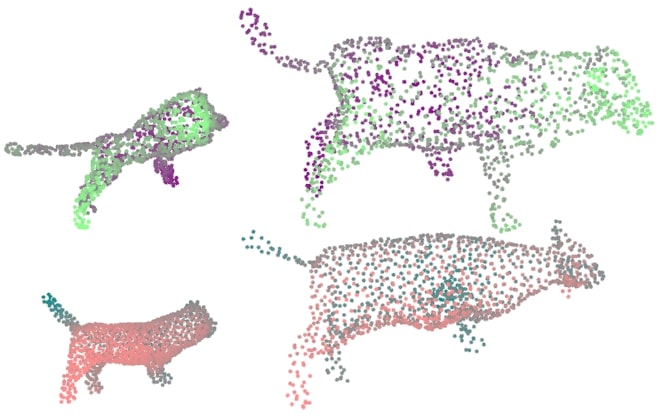} &
    		\includegraphics[height=0.5275in]{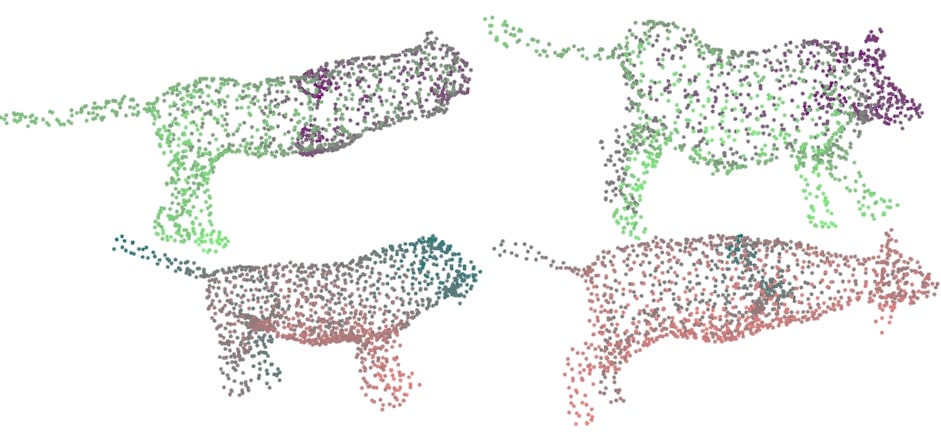} &
    		\includegraphics[height=0.5275in]{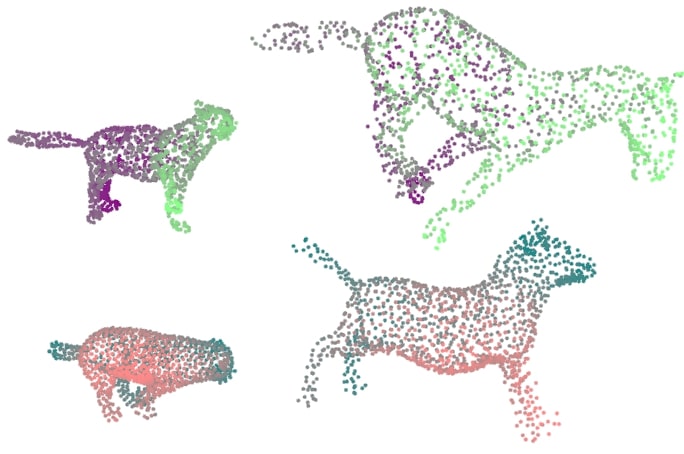}
    	\end{tabular} \\
    	\begin{tabular}{l|c|c|r}		
    		\includegraphics[height=1.21in]{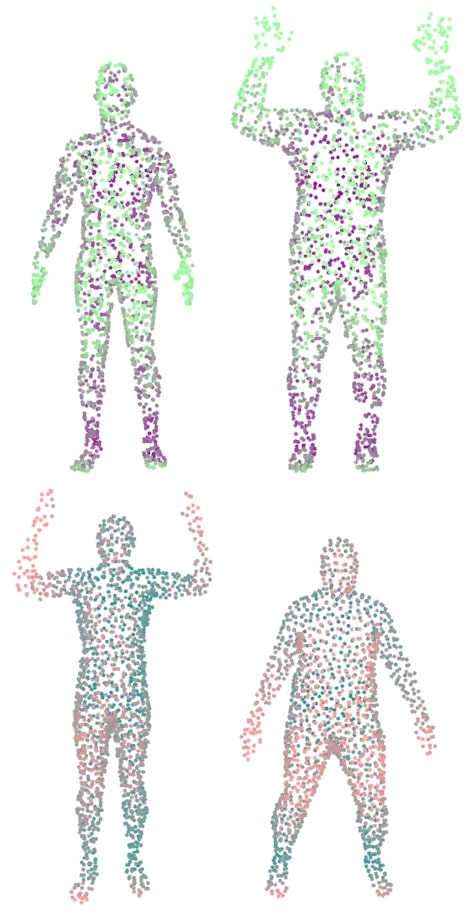} &
    		\includegraphics[height=1.21in]{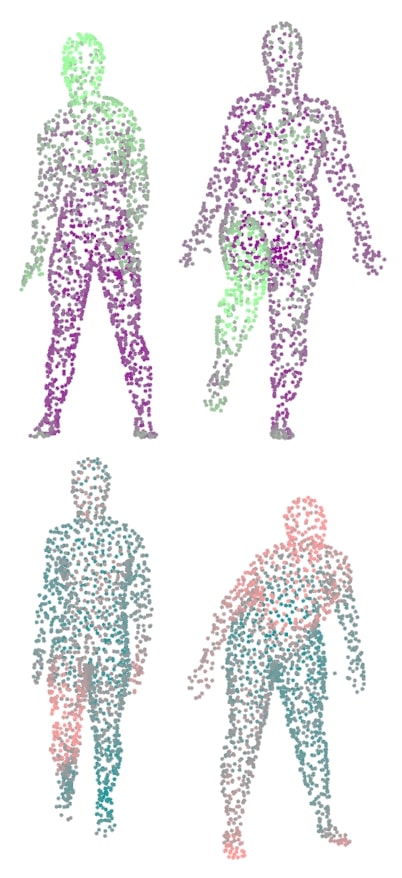} &
    		\includegraphics[height=1.21in]{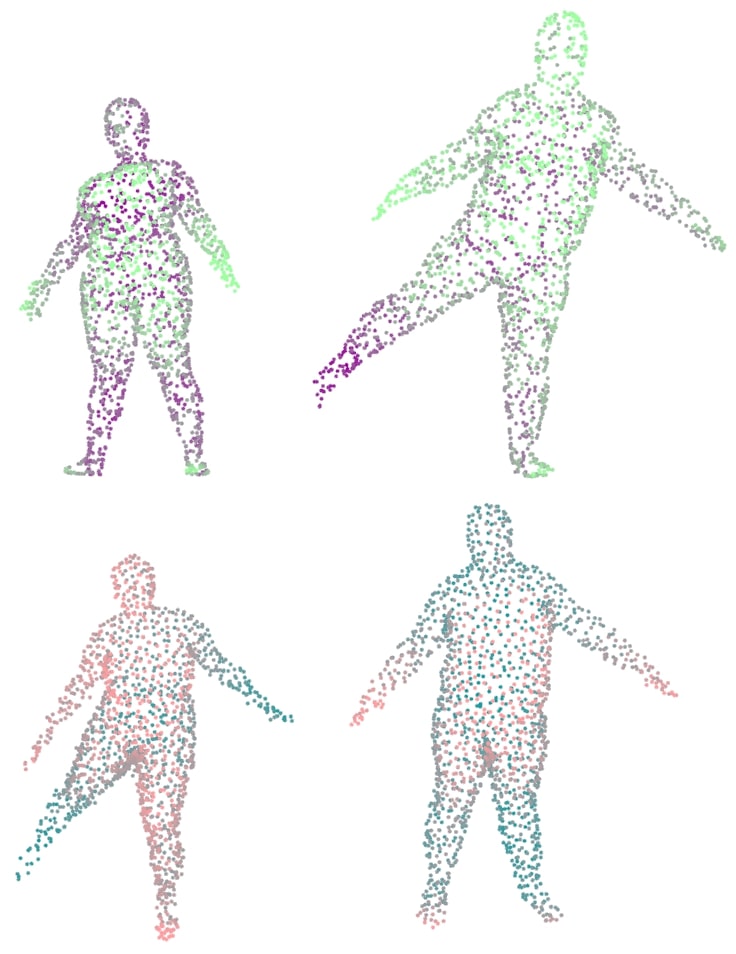} &
    		\includegraphics[height=1.21in]{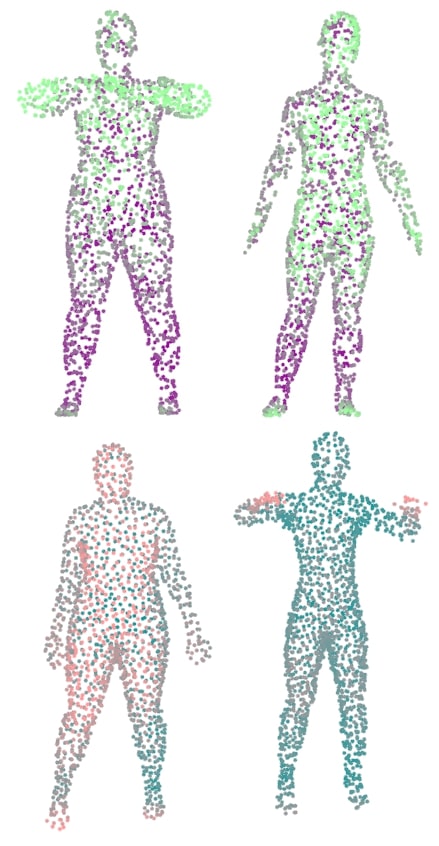}
    	\end{tabular} 
    	\caption{		
    		\textbf{Pose transfer} via exchanging latent extrinsics. 
    		Per inset of four shapes, 
    		the bottom shapes have the $z_R$ and $z_I$ of the shape directly above, 
    		but the $z_E$ of their diagonally opposite shape in the top row.
    		Per-shape colors denote depth.
    		Upper shapes are real point clouds; lower ones are reconstructions after latent transfer.
    		Rows: SMPL, SMAL, and Dyna examples.
    	}
    	\label{fig:transfer}
    \end{figure}
    
    We provide a qualitative examination of the properties of the geometrically disentangled latent space. 
    For human and animal shapes, we expect $z_E$ to control the articulated pose, while $z_I$ should independently control the intrinsic body shape.
    We show the effect of traversing the latent space within its intrinsic and extrinsic components separately, via linear interpolations between shapes in Figure~\ref{fig:interp} (fixing $z_R=0$). 
    We observe that moving in $z_I$ (horizontally) largely changes the body type of the subject, 
    	associated with identity in humans or species among animals, 
    	whereas moving in $z_E$ (vertically) mostly controls the articulated pose.
    Moving in the diagonal of each inset is akin to latent interpolation in a non-disentangled representation.
    
\begin{figure*}
    	\includegraphics[height=1.096724in]{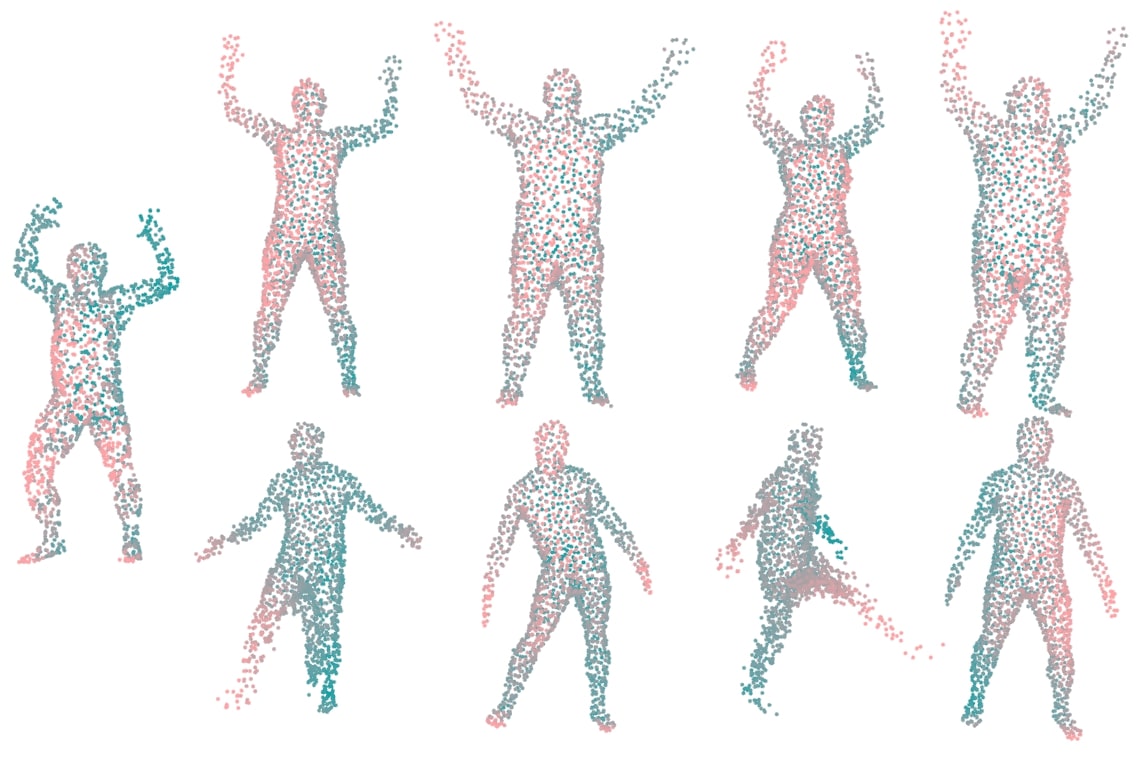} 
    	\hfill
    	\includegraphics[height=1.096724in]{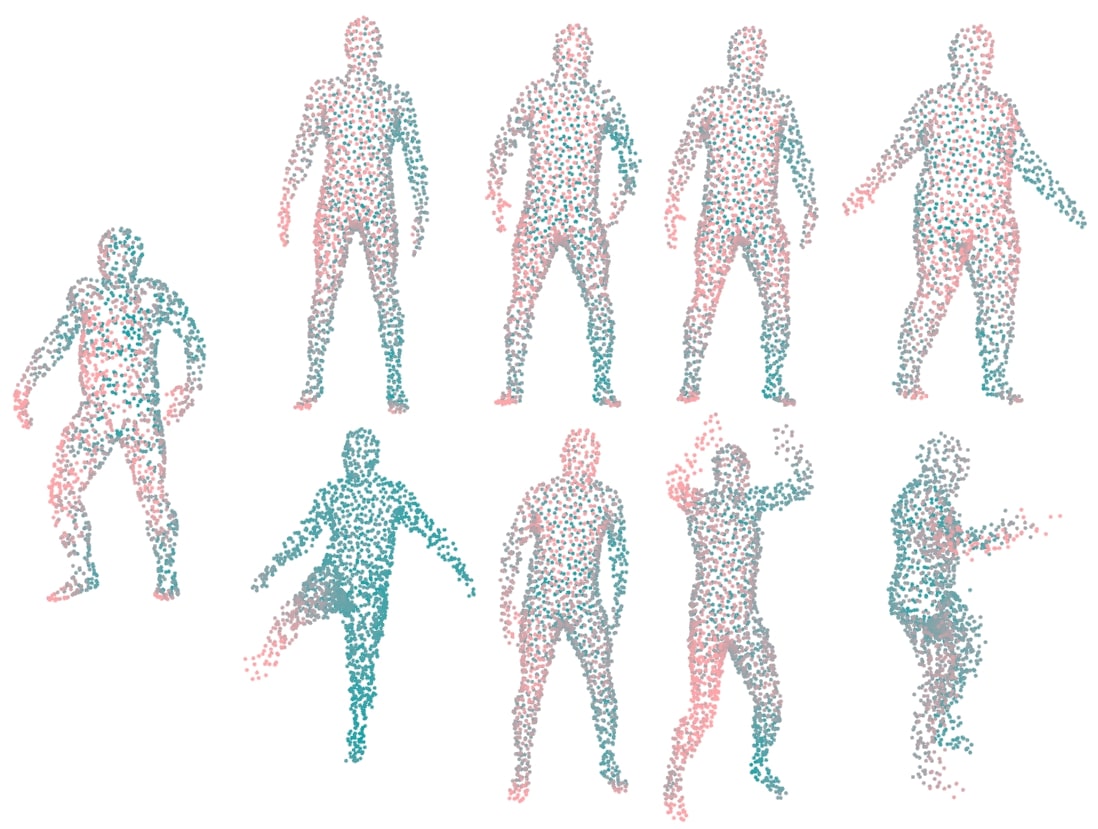} 
    	\hfill
    	\includegraphics[height=1.096724in]{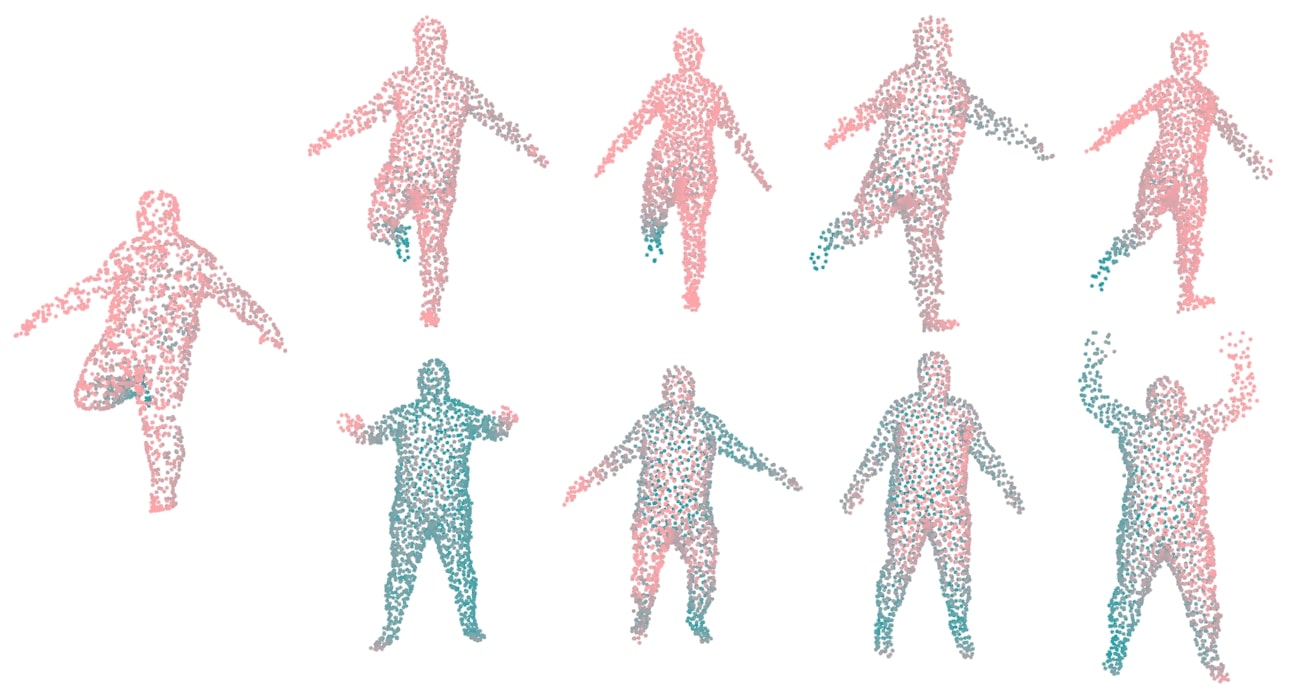} 
    	\hfill
    	\includegraphics[height=1.0724in]{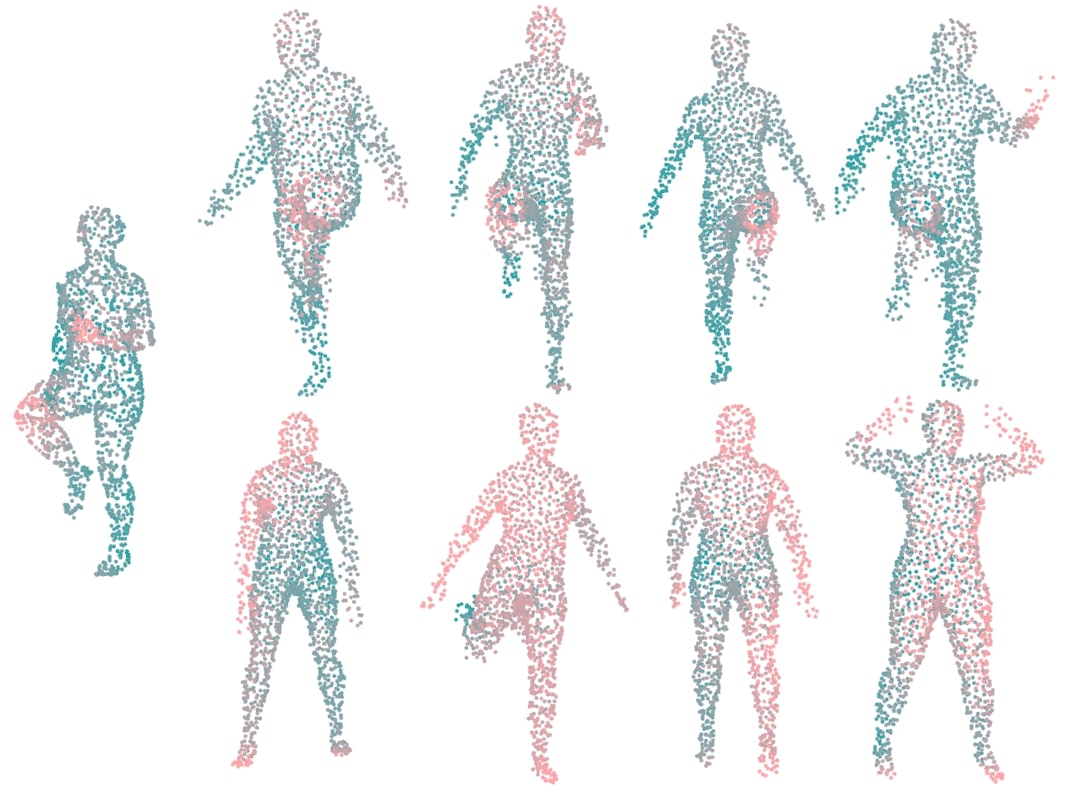} 
    	\caption{Effect of \textbf{randomly sampling either the intrinsic or extrinsic components} of four Dyna shapes. 
    	\textbf{Leftmost shape}: original input; \textbf{upper row:} $z_I\sim\normd(0,I)$, fixed $z_E$ ; \textbf{lower row:} $z_E\sim\normd(0,I)$, fixed $z_I$. Colors denote depth (distance from the camera).}
    	\label{fig:intextSampling}
    \end{figure*}

    We can also consider the viability of our method for pose transfer, 
    by transferring latent extrinsics between two shapes.
    Although the the analogous pose is often exchanged (see \reffig{fig:transfer}), 
    there are some failure cases: 
    for example, on SMPL and Dyna, the transferred arm positions tend to be similar, but not exactly the same.
    This suggests a failure in the disentanglement, since the articulations are tied to the latent instrinsics $z_I$.
    In general, we found that latent manipulations starting from real data (e.g., interpolations or pose transfers between real point clouds) gave more interpretable results than those from latent samples,
    suggesting the model sometimes struggled to match the approximate posterior to the prior,
    particularly for the richer datasets from SMAL and SMPL. 
    Nevertheless, on the Dyna set, we show that randomly sampling $z_E$ or $z_I$ can still give intuitive alterations to pose versus intrinsic shape (Figure~\ref{fig:intextSampling}).
    
\subsection{Pose-Aware Shape Retrieval} \label{sec:experiments:retrieval}
    We next apply our model to a classical computer vision task: 3D shape retrieval.
    Note that our disentangled representation also affords retrieving shapes based exclusively on intrinsic shape (ignoring isometries) or articulated pose (ignoring intrinsics). 
    While the former can be done via spectral methods 
    (e.g., \cite{bronstein2011shape,pickup2016shape}), 
    the latter is less straightforward. 
    Our method also works directly on raw point clouds.
    
    We measure our performance on this task using the synthetic datasets from SMAL and SMPL. 
    Since both are defined by intrinsic shape variables ($\beta$) and articulated pose parameters (Rodrigues vectors at joints, $\theta$), we can use knowledge of these to validate our approach quantitatively. 
    Note that our model only ever sees raw point clouds (i.e., it cannot access $\beta$ or $\theta$ values).
    Our approach is simple: after training, we encode each shape in a held-out test set, and then use the $L_2$ distance in the latent spaces ($X$, $z$, $z_E$, and $z_I$) to retrieve nearest neighbours. 
    We measure the error in terms of how close the $\beta$ and $\theta$ values of the query $P_Q$ ($\beta_Q$, $\theta_Q$) are to those of a retrieved shape $P_R$ ($\beta_R$, $\theta_R$). 
    We define the distance $E_\beta(P_Q,P_R)$ between the shape intrinsics as the mean squared error $MSE(\beta_Q, \beta_R)$.
    To measure extrinsic pose error, 
    we first transform the axis-angle representation $\theta$ to the equivalent unit quaternion $q(\theta)$, and then compute
    $ E_\theta(P_Q,P_R) = \mathcal{L}_Q(q(\theta_Q),q(\theta_R)) $.
    We also normalize each error by the average error between all shape pairs,
    thus measuring our performance compared to a uniformly random retrieval algorithm.
    Ideally, retrieving via $z_E$ should have a high $E_\beta$ and a low $E_\theta$, while using $z_I$ should have a high $E_\theta$ and a low $E_\beta$.
    
\begin{figure}[t]
    	\centering
    	\includegraphics[width=0.47\textwidth]{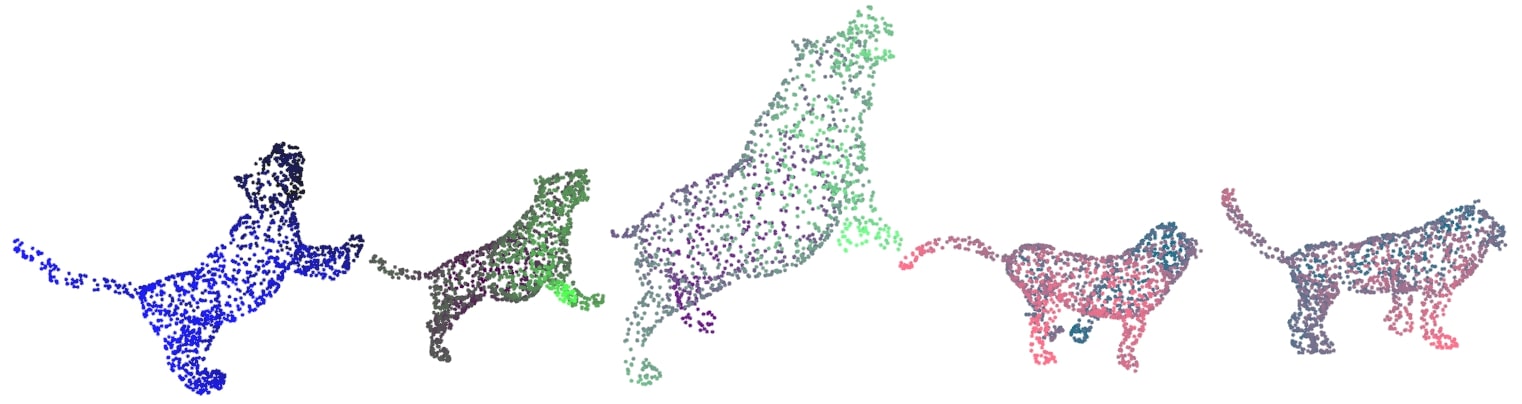} \\
    	\includegraphics[width=0.175\textwidth]{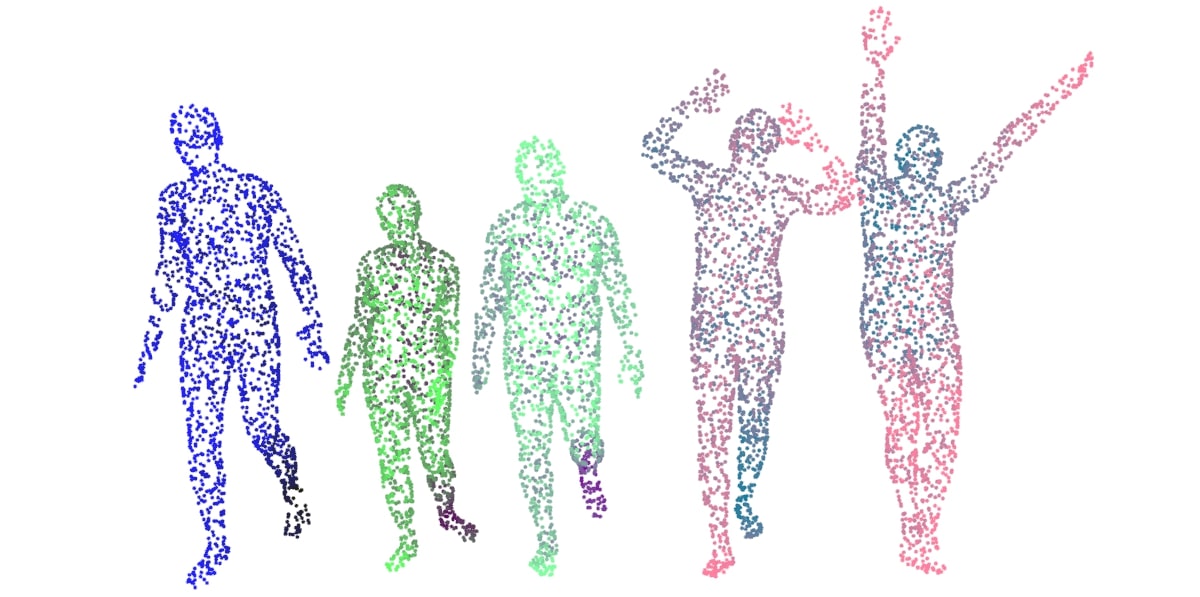} \hfill
    	\includegraphics[width=0.289\textwidth]{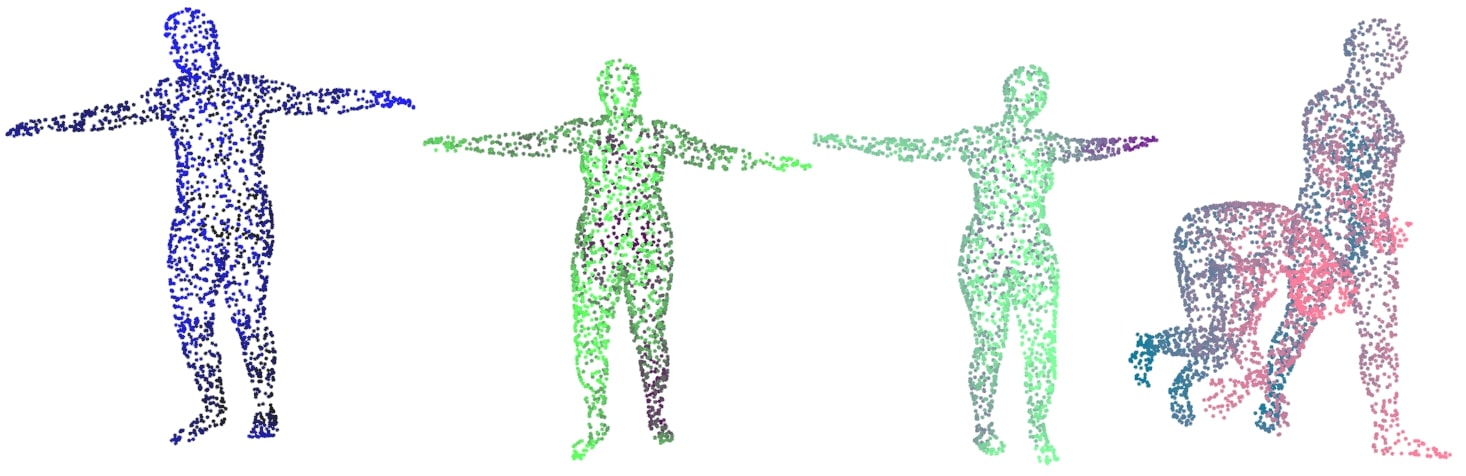} 
    	\caption{\textbf{Shape retrieval}. Per inset: leftmost shape is query, middle two shapes are retrieved via $z_E$, and rightmost two shapes are retrieved via $z_I$. 
    	Color gradients per shape denote depth.}
    	\label{fig:retEgs}
    \end{figure}

    Table~\ref{table:retrievals} shows the results. 
    Each error is computed using the mean error over the top three matched shapes per query, averaged across the set.
    As expected, the $E_\beta$ for $z_I$ is much lower than for $ z_E $ (and $z$ on SMAL), while the $E_\theta$ for $z_E$ is much lower than that of $z_I$ (and $z$ on SMPL).
    Just as importantly, from a disentanglement perspective, we see that the $E_\beta$ of $z_E$ is much higher than that of $z$, as is the $E_\theta$ of $z_I$.
    We emphasize that $E_\beta$ and $E_\theta$ measure different quantities, and should not be directly compared; instead, each error type should be compared across the latent spaces. 
    In this way, $z$ and $X$ serve as non-disentangled baselines, where both error types are low.
    This provides a quantitative measure of geometric disentanglement which shows that our unsupervised representation is useful for generic tasks, such as retrieval. 
    Figure~\ref{fig:retEgs} shows some examples of retrieved shapes
    using $z_E$ and $z_I$.
    The high error rates, however, do suggest that there is still much room for improvement. 	

    \begin{table} 
    	\centering
    	\begin{tabular}{cc|cccc}
    		$\,$ & $\,$  & $X$   & $z$   & $z_E$  & $z_I$ \\\hline        
    		\multirow{2}{*}{SMAL} 
    		& $E_\beta$  & 0.641 & 0.743 & 0.975  & 0.645 \\
    		& $E_\theta$ & 0.938 & 0.983 & 0.983  & 0.993 \\\hline    
    		\multirow{2}{*}{SMPL} 
    		& $E_\beta$  & 0.856 & 0.922 & 0.997  & 0.928 \\
    		& $E_\theta$ & 0.577 & 0.726 & 0.709  & 0.947
    	\end{tabular}
    	\caption{
    	    Error values for retrieval tasks, using various latent representations.
    	    Values are averaged over three models trained with the same hyper-parameters, 
    	        with each model run three times to account for randomness in
    	        the point set sampling of the input shapes.
    	    (See Supplemental Material for standard errors).
    	}
    	\label{table:retrievals}
    \end{table}
    
\subsection{Disentanglement Penalty Ablations} \label{sec:experiments:ablation}
    We use three disentanglement penalties to control the structure of the latent space, based on the inter-group total correlation (TC), covariance (COV), and Jacobian (J).
    To discern the contributions of each, we conduct the following experiments 
    (details and figures are in the Supplemental).
    	
    We first train several models on MNIST, monitoring the loss curves while we vary the strength of each penalty.
    We find that higher TC penalties substantially reduce COV and J, 
    while COV and J are less effective in reducing TC.
    This suggests TC is a ``stronger'' penalty than COV and J, which is intuitive, 
    given that it directly measures information, rather than linear relationships (as COV does) or local ones (as J does).
    Nevertheless, it does not remove the entanglement measured in COV and J as effectively as direct penalties on them, 
    and using higher TC penalties quickly leads to lower reconstruction performance.
    Using all three penalties achieves the lowest values for all measures.
    
    We then perform a more specific experiment on the SMAL and SMPL datasets, ablating the COV and/or J penalties, 
    and examining both the loss curves and the retrieval results. 
    Particularly on SMPL, the presence of a direct penalty on COV and J is very useful in reducing 
    their respective values.
    Regarding retrieval, the $E_\beta$ using $z_I$ on SMAL and the $E_\theta$ using $z_E$ on SMPL were lowest using all three penalties. 
    Interestingly, $E_\beta$ using $z_I$ on SMPL and $E_\theta$ using $z_E$ on SMAL could be improved without COV and J; however, such decreases were concomitant with reductions in $E_\theta$ using $z_I$ and $E_\beta$ using $z_E$, which suggests increased entanglement.
    While not exhaustive, these experiments suggest the utility of applying all three terms.

    We also considered the effect of noise in the spectra estimates 
        (see Supplemental Material). 
    The network tolerates moderate spectral noise, 
        with decreasing disentanglement performance as the noise increases.
    In practice, one may use meshes with added noise for data augmentation,
        to help generalization to noisy point clouds at test time.
   

\section{Conclusion} \label{sec:conclusion}

We have defined a novel, two-level unsupervised VAE with a disentangled latent space, using \emph{purely geometric} 
information (i.e., without semantic labels).
We have considered several hierarchical disentanglement losses, 
including a novel penalty based on the Jacobian of the latent variables of the reconstruction with respect to the original latent groups, and have examined the effects of the various penalties via ablation studies.
Our disentangled architecture can effectively compress vector representations via encoding and perform generative sampling of new shapes.
Through this factored representation, our model permits several downstream tasks on 3D shapes (such as pose transfer and pose-aware retrieval), which are challenging for entangled models, without any requirement for labels.

\vspace{2pt plus 2pt minus 2pt}
\noindent
\textbf{Acknowledgments}
We are grateful for support from NSERC (CGS-M-510941-2017) and Samsung Research.

{\small
	\bibliographystyle{ieee_fullname}
	\bibliography{ttaa}
}

\clearpage
\newpage 

\appendix

\noindent{\normalfont\Large\bfseries Appendix}

\section{Dataset Details}

\subsection{MNIST Dataset}
\label{section:data:mnist}

As a simple dataset on which to test our method, we generate point clouds from the greyscale MNIST images.
We first produce triangle meshes by placing vertices at each pixel, with the $x$ and $y$ coordinates normalized by image width, and the $z$ coordinate given by 0.1 times the normalized pixel value. 
Edges are added between each horizontal and vertical neighbour, as well as one diagonal, thus defining 2 triangles per set of four vertex neighbours.
This is essentially the mesh of the height field defined by the image of the digit.
We then threshold the mesh, deleting any vertices with a $z$ height value less than 0.01. 
Finally, viewing the mesh as a graph, we take the largest connected component, giving us the final triangle mesh from which we sample our point clouds.
This resulted in 59483 training and 9914 testing meshes with spectra.
For the MNIST dataset, we fixed $N_T = 2000$, $N_\lambda = 40$, and $N_S = 1000$.

\subsection{MPI Dyna Dataset}
\label{section:data:dyna}

The Dyna dataset \cite{dyna} consists of 3D scans of 10 individuals performing various sequences of simple actions (e.g., holding up their arms). 
In total, the dataset contains approximately 40K triangle meshes of multiple body types in a large variety of poses.
During training, we fixed the total number of samples to $N_T=6000$, the spectrum length as $N_\lambda=100$, and the size of the input point clouds as $N_S=2000$ for the Dyna dataset.

\subsection{SMAL-derived Dataset}
\label{section:data:smal}

Using the SMAL model \cite{Zuffi:CVPR:2017}, we generated a dataset of animal meshes, including random body types and articulated motions. In detail, we use the fitted multivariate Gaussians computed by the authors of SMAL, which each act as a distribution over clusters of the shape parameters of the same animal species. We generate 3200 meshes for each of the five categories by sampling from each cluster distribution. Following the generation procedure in other works \cite{groueix20183d}, we then sample the pose (here, the joint angles) via a Gaussian distribution with a standard deviation of 0.2. We then split the resulting dataset into $15000$ training and $1000$ testing meshes (each comprised of equal numbers of meshes per species).
For SMAL, the total number of samples was $N_T=8000$, the spectrum length was $N_\lambda=50$, and the size of the input point clouds during training was $N_S=1500$.

\subsection{SMPL-derived Dataset}
\label{section:data:smpl}

Similar to the dataset derived from SMAL, we generate a dataset of human meshes with random body types and articulations via the SMPL model \cite{SMPL:2015}.
We largely follow the protocol from 3D-CODED \cite{groueix20183d}.
Briefly, we sampled 20500 meshes from each of the male and female models using random samples from the SURREAL dataset \cite{varol17_surreal}. 
We augmented this data with 3100 meshes of ``bent'' humans per gender, using the alterations from Groueix et al \cite{groueix20183d}.
We then assigned 500 unbent and 100 bent meshes (per gender) to the held-out test set, and the remaining meshes to the training set. This resulted in 45992 meshes and 1199 testing meshes after spectral calculations.
Using these meshes, we derived point clouds with $N_T=8000$, $N_\lambda=40$, and $N_S=2000$.


\section{Spectral Geometry Intuition}

In this section, we provide some intuition for the geometric meaning of the spectrum and why it can be used as a geometrically disentangled prior for shape representation, particularly for shapes that undergo isometric articulated pose transforms.

\begin{figure}
	\centering
	\includegraphics[width=0.497\textwidth]{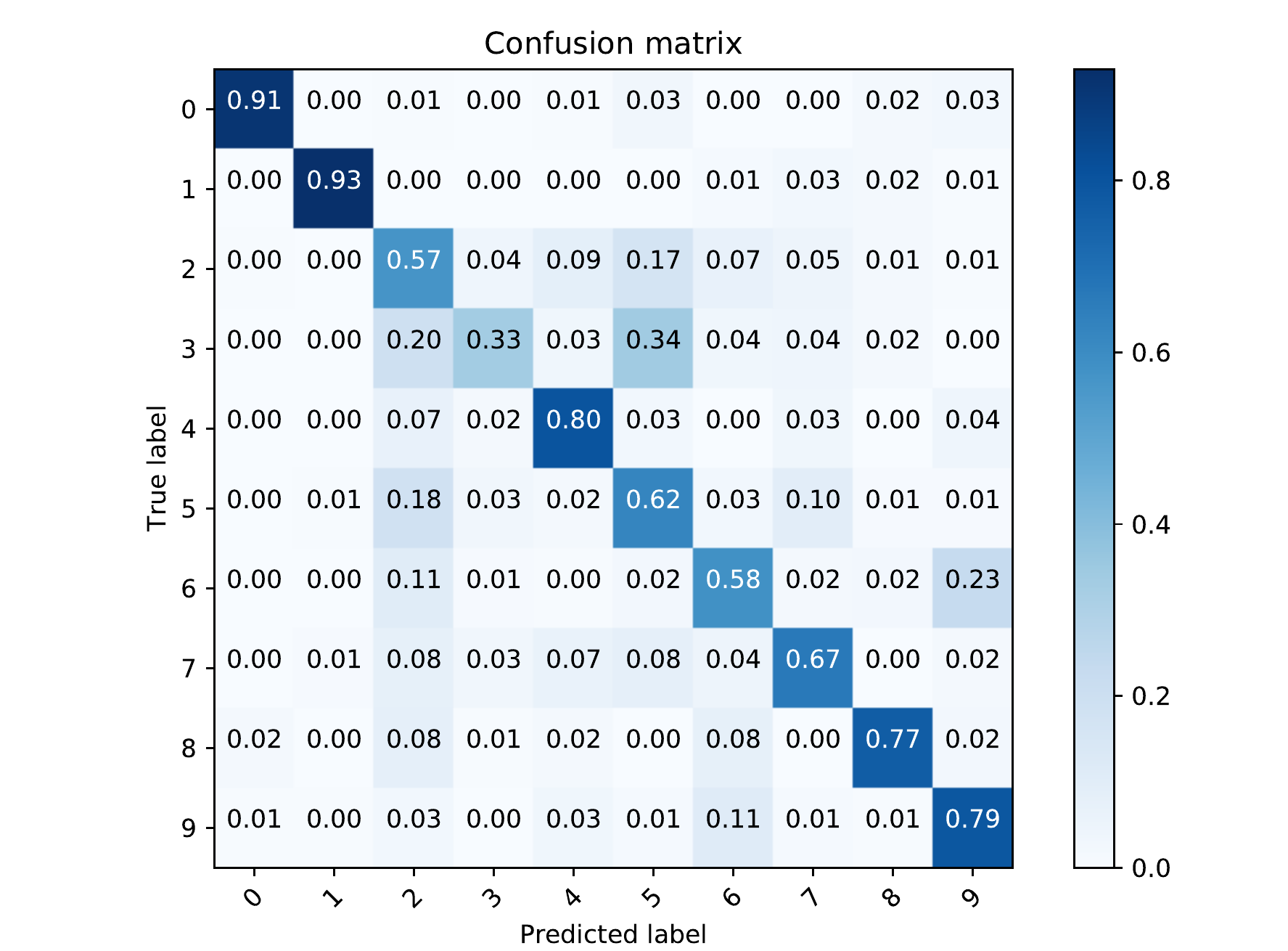}
	\caption{Confusion matrix of a shallow classifier, mapping mesh spectra to digit identity. Note the main confusions, e.g., between 9 and 6, match our intuition for the shape of the digit geometrically.}
	\label{fig:mnistSpectra}
\end{figure}

For MNIST, an obvious question is how the intrinsic geometry of a digit's shape captured by its LBO spectrum is related to its semantic label (i.e., numeric value).
Intuitively, 
a natural notion of the shapes of the digits should be closely related to their numeric identity, 
while minor perturbations that change the style of the digits should have less of an effect on the intrinsic shape.
We examine this relation by training a simple classifier, 
which learns to map from the spectrum vector to the digit identity.
We use a neural network with one hidden layer (size 100), 
using the ReLU non-linearity and otherwise default parameters from the scikit-learn library \cite{scikit-learn}.
This obtains an accuracy of 0.69, 
suggesting that the intrinsic digit shape alone is capable of significant discriminatory power, 
but not as much as the complete shape information.
The confusion matrix is shown in Figure~\ref{fig:mnistSpectra}.
The observed results agree with our intuition:
for instance, 9 and 6 are close to rotations of each other, 
while 2 and 5 also clearly have very similar intrinsic shapes; 
unsurprisingly, these are relatively more common misclassifications. 
The most misclassified shapes, however, are 3 and 5, 
which differ only in the placement of the ``bridge'' between the lower curved part and the upper line making up the digits.


\begin{figure}[t]
	\centering
	\includegraphics[width=0.497\textwidth]{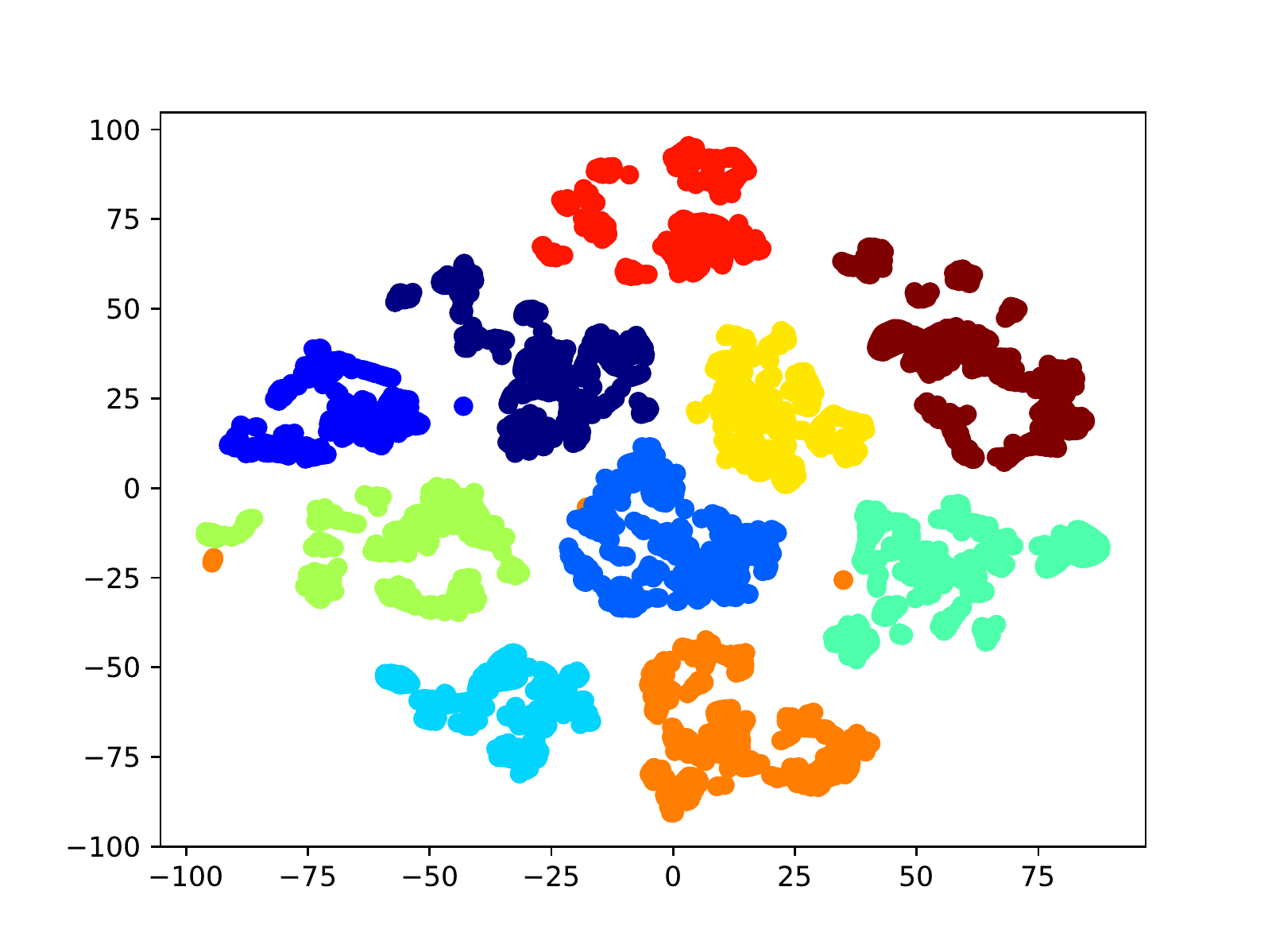}
	\caption{A t-SNE plot of the LBO spectra of the human shapes in the Dyna dataset. 
		Each color corresponds to a different individual. Note the natural 
		clusters formed by individuals. }
	\label{fig:tsnedyna}
\end{figure}

For datasets of some objects, such as people, articulations (i.e., non-rigid changes in pose for a single person) are nearly isometric transformations.
Just as one naturally divides rigid and non-rigid deformations of objects, so too can one separate isometries from non-isometric deformations.
To illustrate, 
in Figure~\ref{fig:tsnedyna}, a t-Distributed Stochastic Neighbor Embedding 
(t-SNE)~\cite{maaten2008visualizing} plot (via scikit-learn~\cite{scikit-learn}) 
shows the embeddings of the spectra of a portion of the Dyna dataset 
(every sixth shape in each activity sequence). 
Notice that the embedding naturally clusters by individual, as the intrinsic 
shape of each individual's body is largely unchanged by the pose articulations 
of each action sequence. 
This corresponds to an intuitive view of intrinsic shape as being both 
continuous (as opposed to labelling each individual) and, in this case, 
articulation invariant.


\section{Architectures}
\label{appendix:architectures}

All models were implemented in PyTorch \cite{paszke2017automatic}.
For exact values for each dataset, see Table \ref{supp:table:hyperparams}.

\subsection{Autoencoder Architecture}
\label{appendix:ae_architecture}

The encoder consists of a PointNet model \cite{qi2017pointnet}, followed by fully-connected (FC) layers.
The convolutional layers were of sizes $S_{\text{AE,C}}$.
There were two affine transformers (``T-networks'') after the first and third layers, with convolutional layers of size 64, 128, and 512, followed by an FC layer with hidden size 
256.
After the convolutional layers and affine transforms, 1D max pooling was applied, followed by processing by a FC network with hidden layers of sizes $S_{\text{AE,FE}}$.
The final latent representation of the AE was of dimensionality $L_\text{AE}$.
The decoder directly generated a fixed size point cloud (with $N_\text{AE,OUT}$ points), via a series of FC layers with hidden layer sizes $S_{\text{AE,FD}}$

We fixed the loss weighting parameters as follows: 
 $ \alpha_C = 0.75, \alpha_H = 0.5$. 
For MNIST and Dyna, FC and 1D convolutional layers utilized a BatchNorm layer \cite{ioffe2015batch} and the ReLU non-linearity between linear transforms.
For SMPL and SMAL, no normalization and layer normalization \cite{ba2016layer} were used, respectively, in the decoder
(similar to the encoder-only usage of batch norm in prior work \cite{achlioptas2018learning}).
Data augmentation consisted of random rotations about the ``height'' axis of the models (e.g., see Figure \ref{ae_rot_fig}),
with rotations in $[0,2\pi]$ for MNIST and Dyna and in $[-\pi/6,\pi/6]$ for SMPL and SMAL.
A batch-size of $S_B$
and a learning rate of $\eta_\text{AE,LR}$
was used, with Adam \cite{kingma2014adam} as the optimizer, 
for $N_\text{EP}$ epochs.

\begin{table*}
    \centering
    \begin{tabular}{cc|c|c|c|c}
       $\,$ & $\,$ & MNIST & Dyna & SMAL & SMPL \\\hline
       \multirow{8}{*}{AE} & $S_{\text{AE,C}}$ 
            & 50, 100, 200, 300, 400  
            & 50, 100, 200, 300, 400 
            & 50, 100, 200, 400, 500 
            & 50, 100, 200, 400, 500 \\
       $\,$ & $S_{\text{AE,FE}}$
            & 300
            & 300
            & 800
            & 500 \\
       $\,$ & $L_\text{AE}$
            & 250 
            & 250
            & 400
            & 300 \\
       $\,$ & $S_{\text{AE,FD}}$
            & 300, 500, 800
            & 300, 500, 800
            & 800, 800, 2000
            & 500, 800, 2000 \\
       $\,$ & $r_C,\, r_H$ 
            & 10, 1
            & 10, 1
            & 20, 0.1
            & 20, 0.1 \\
       $\,$ & $N_\text{AE,OUT}$
            & 1000
            & 2000
            & 1500
            & 2000 \\ 
       $\,$ & $S_B, N_\text{EP}$
            & 32, 100 
            & 32, 200
            & 30, 500
            & 20, 100 \\
       $\,$ & $\eta_\text{AE,LR} $
            & 0.00005
            & 0.00005
            & 0.00025
            & 0.00025 \\\hline 
       \multirow{5}{*}{VAE} & $S_{\lambda}$
            & 500, 400
            & 500, 400
            & 300, 300
            & 300, 400 \\
       $\,$ & $S_\text{V,B}$
            & 200
            & 200
            & 250
            & 250 \\
       $\,$ & $ N_\text{V,EP} $
            & 150
            & 150
            & 200
            & 100  \\
       $\,$ & $ |z_E|,|z_I| $ 
            & 5, 5
            & 10, 10 
            & 8, 5
            & 12, 5 \\
       $\,$ & $\beta_4, \gamma_I, w_J $
            & 50, 1, 1
            & 25, 5, 5 
            & 50, 100, 10
            & 50, 10, 10 \\
    \end{tabular}\vspace{0.25mm}
    \caption{Architectural parameters for the models across different datasets. See Sections \ref{appendix:ae_architecture}
and \ref{appendix:vae_architecture} for details concerning the AE and VAE parameters, respectively. }
    \label{supp:table:hyperparams}
\end{table*}
\subsection{GDVAE Architecture}
\label{appendix:vae_architecture}

The encoder consists of two parts: one for the rotation $R$ and one for the shape $X$.
Each part consists of two sub-parts: 
	a set of shared layers, followed by a separate network for the variational parameters $\mu$ and $\Sigma$ 
	(i.e., 
	$z_R\sim
	\normd( \mu(g_\text{shared,r}(R)),
			\Sigma(g_\text{shared,r}(R)) )$ 
	and 
	$(z_E,z_I)\sim
	\normd( \mu(g_\text{shared,x}(X)),
			\Sigma(g_\text{shared,x}(X)) )$). 
For the rotation mapping, 
	the shared layers consisted of FC layers with hidden sizes
	$ 300, 200 $,
	while the mean and variance mappings are each given by a single affine transform.
For the $X$ mapping,
	the shared layers consisted of FC layers with hidden sizes 
	1000, 750, and 500,
	while the mean and variance are parameterized by a FC network with one hidden layer of size 
	250 (MNIST/Dyna) or 300 (SMPL/SMAL). 
	
The decoder also consisted of two parts:
	the rotation decoder 
	(FC layers with sizes 200 and 300) 
	and the  shape decoder 
	(FC layers of sizes
	500, 750, and 1000).
The spectral predictor defined on the latent intrinsic $z_I$ space was an FC network with 
layer sizes $ S_{\lambda} $. 

Models used a batch-size of 
$S_\text{V,B}$
for 
$ N_\text{V,EP} $
epochs, using Adam with a learning rate of 
0.0001. 
A small $L_2$ weight decay with coefficient $5\times 10^{-6}$ was used for MNIST and Dyna.
Data was again augmented by random rotations about the height axis, but limited to an angular magnitude within $[-\pi/5, \pi/5]$ for MNIST and Dyna, and $[-\pi/12, \pi/12]$ for SMPL and SMAL.

Concerning hyper-parameters, unless otherwise specified, we used 
a spectral weight of $\zeta = 1000$,
a relative quaternionic loss weight of $w_Q = 10$,
hierarchically factorized VAE loss coefficients of
$\beta_1 = \beta_2 = \beta_3 = 1.0$,
and a reconstruction loss weight of 
$\eta = \dim(X)$ (on MNIST and Dyna),
$\eta = 200$ (on SMPL),
and $\eta = 1$ (on SMAL).
On SMPL, we used $\zeta = 500$, however.
Recall that the weights on the disentanglement penalties were written
$\beta_4$ for the inter-group TC,
$\gamma_I$ for the inter-group covariance,
and 
$w_J$ for the Jacobian.

	\begin{table}
		\centering
		\begin{tabular}{c|ccccc}
			$\,$ 		& Beta-VAE & Jacobian & Cov   & TC   & All \\
			\hline 
			\rule{0pt}{1.01\normalbaselineskip}
			$\beta_4$	& 1		   & 1		  & 1     & 100  & 100 \\
			$\gamma_I$	& 0		   & 0		  & 100   & 0    & 100 \\
			$w_J$		& 0		   & 100	  & 0     & 0    & 100 \\
		\end{tabular}
		\caption{Hyper-parameter values varied in the test of various loss weightings. (See Figure \ref{hyperParams} for visualization of results).}
		\label{table_hypertest}
	\end{table}


\begin{figure*}
	\includegraphics[height=0.914in]{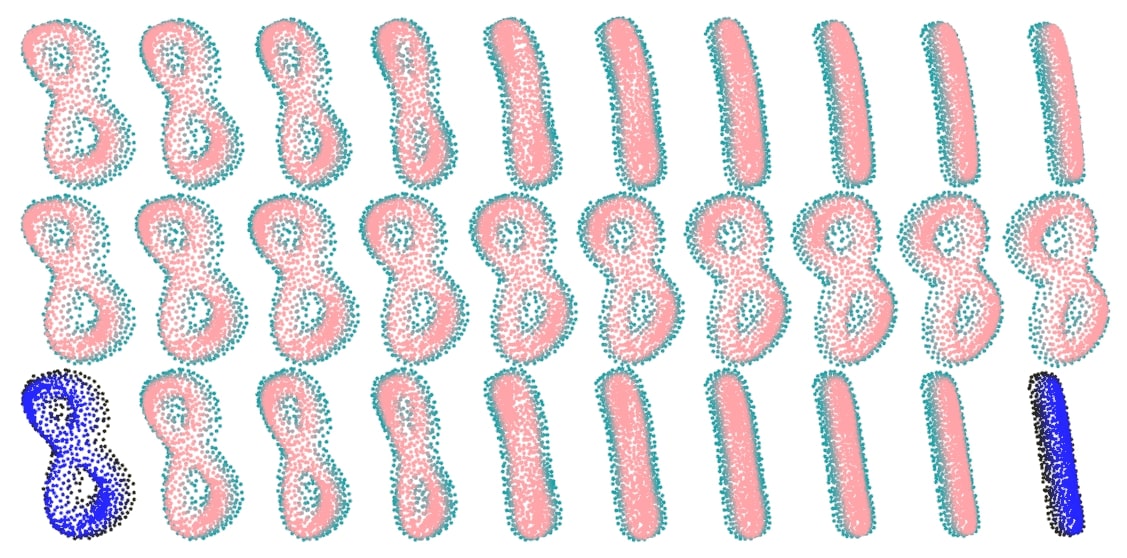}
	\hfill
    \includegraphics[height=0.914in]{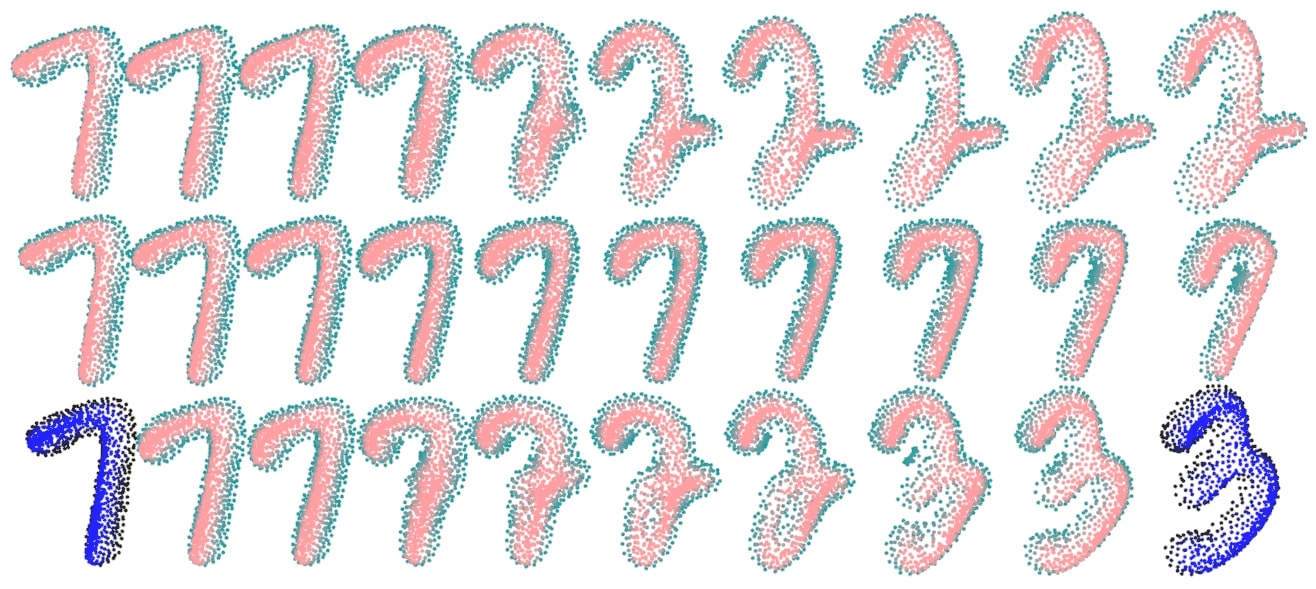}
    \hfill 
    \includegraphics[height=0.914in]{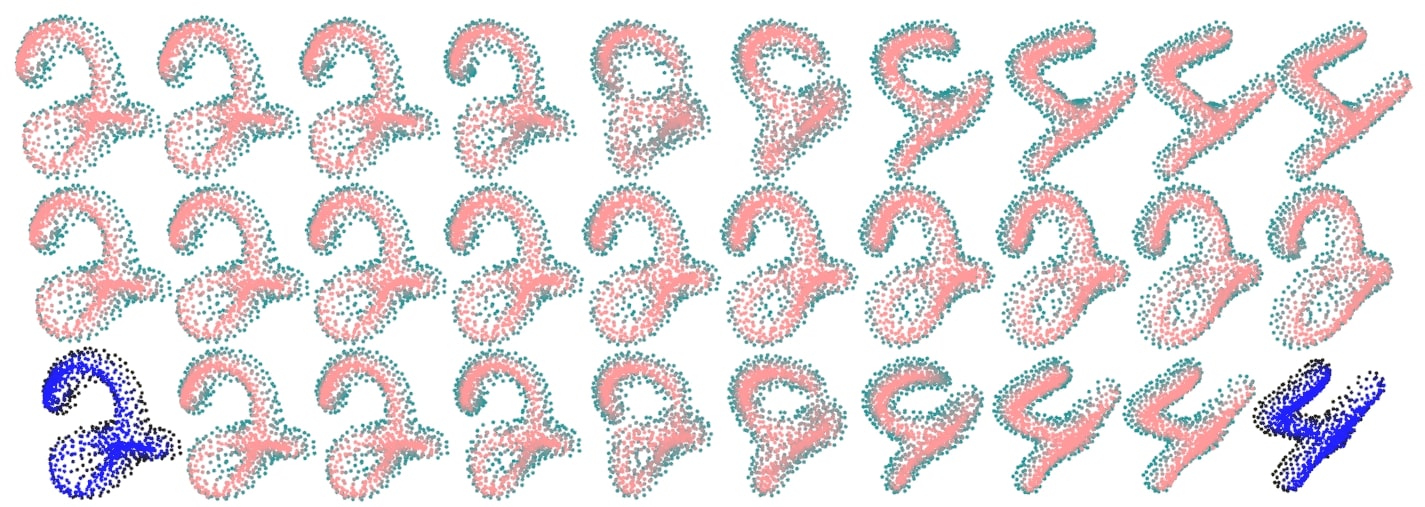} \\
    \includegraphics[height=0.91in]{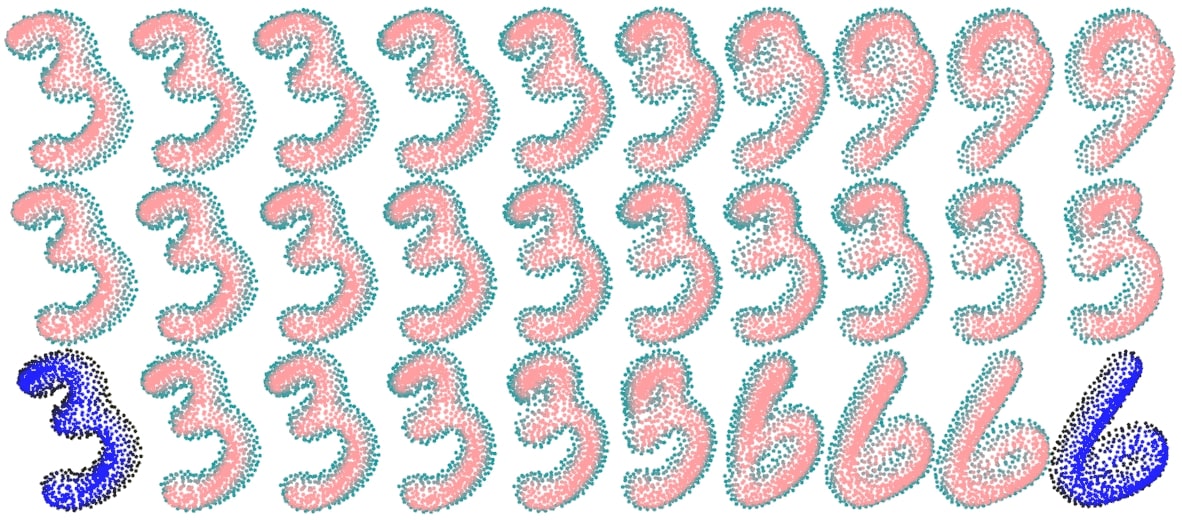}
	\hfill
    \includegraphics[height=0.91in]{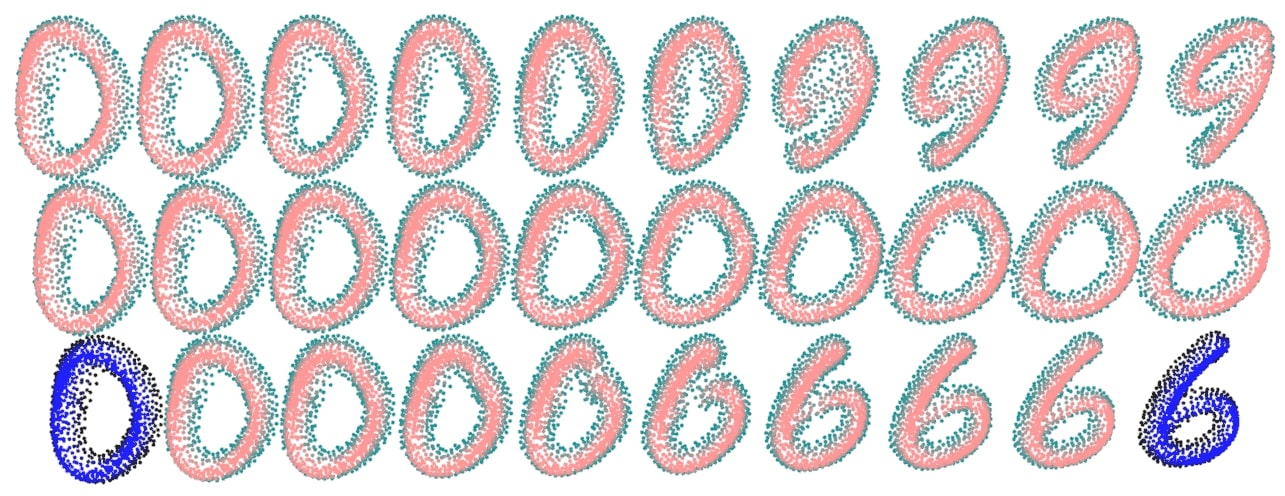}
    \hfill 
    \includegraphics[height=0.91in]{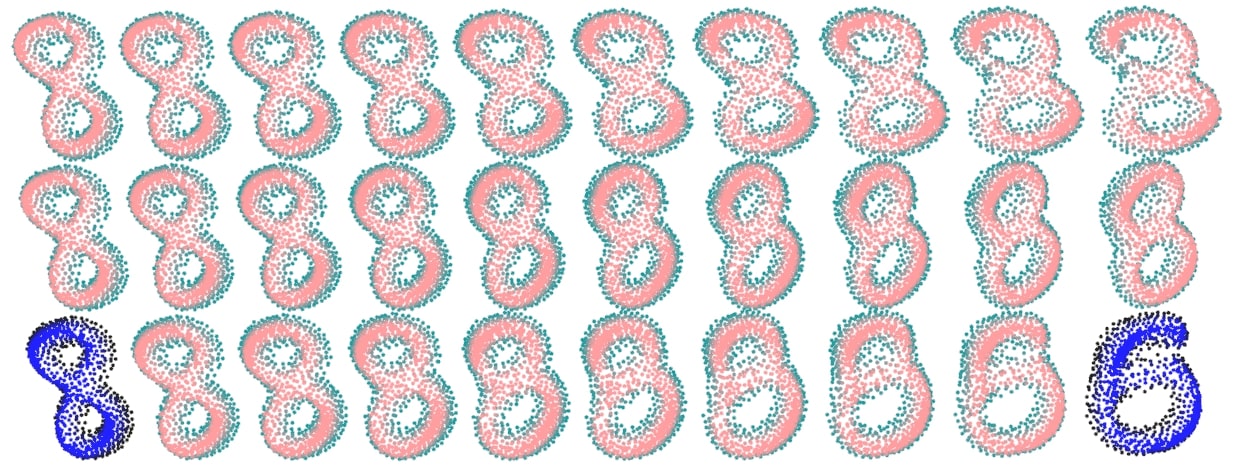}
	\caption{Disentangled latent interpolations for the MNIST dataset. 
	Colours per digit denote depth.
	For each inset, we are interpolating between the blue-black digits in the bottom row.
	The first row corresponds to only moving through $z_I$ (with $z_E$ constant), 
	while the second corresponds to only moving through $z_E$ (holding $z_I$ fixed).
	The third row traverses the full latent $z$ space.
	In all cases, $z_R$ is set to zero.
	}
\label{mnist_interps}
\end{figure*}

	\begin{figure*}
		\centering
		\includegraphics[width=0.3253\textwidth]{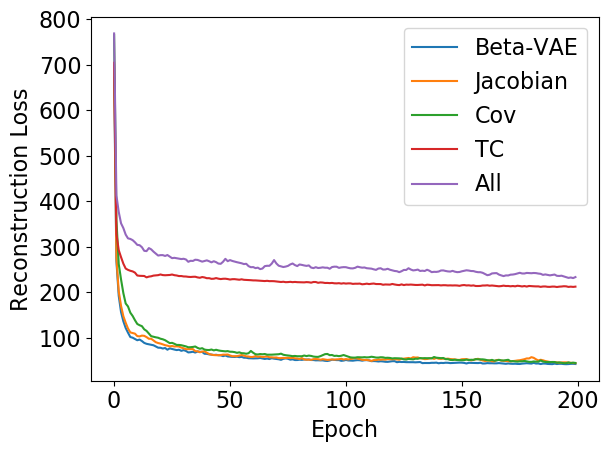}
		\includegraphics[width=0.3253\textwidth]{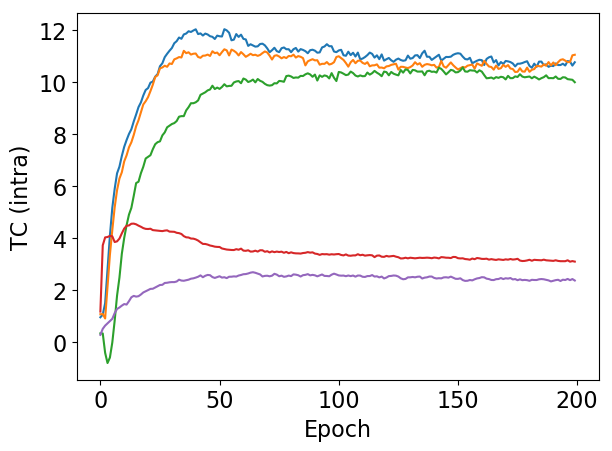}
		\includegraphics[width=0.3253\textwidth]{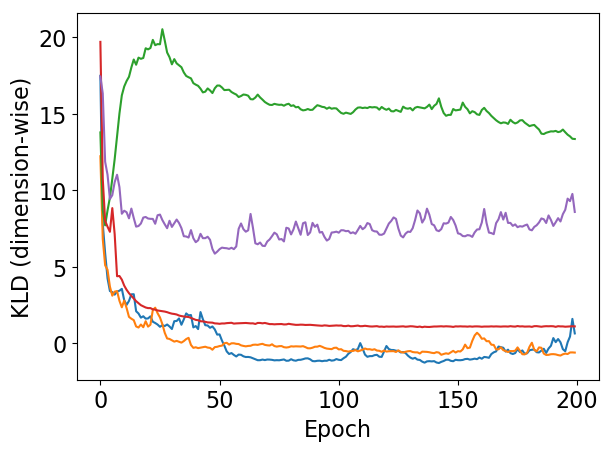}\\
		\includegraphics[width=0.3253\textwidth]{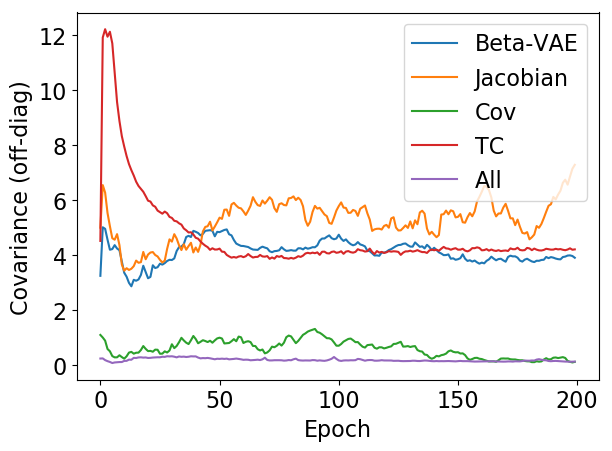}
		\includegraphics[width=0.3253\textwidth]{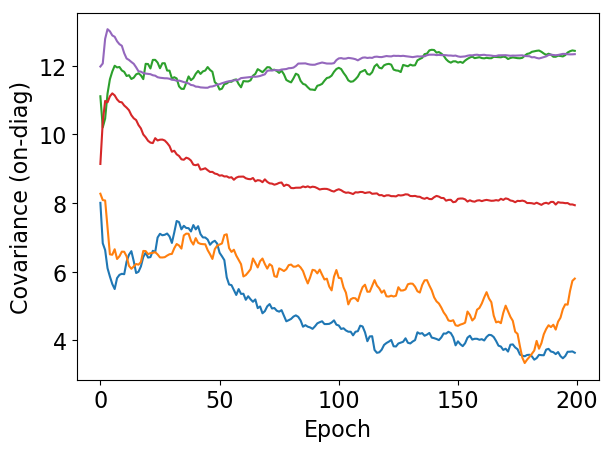}
		\includegraphics[width=0.3253\textwidth]{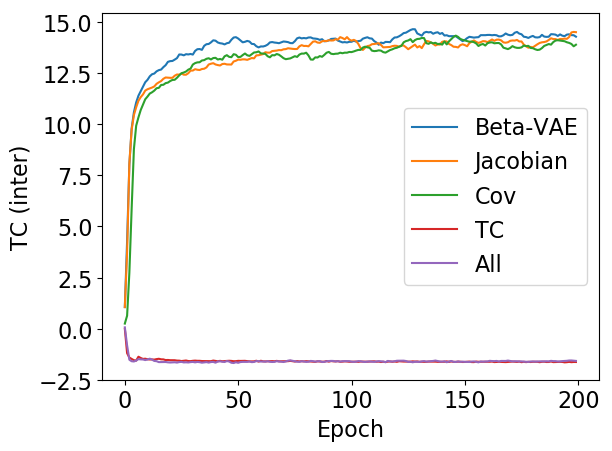}\\
		\includegraphics[width=0.3253\textwidth]{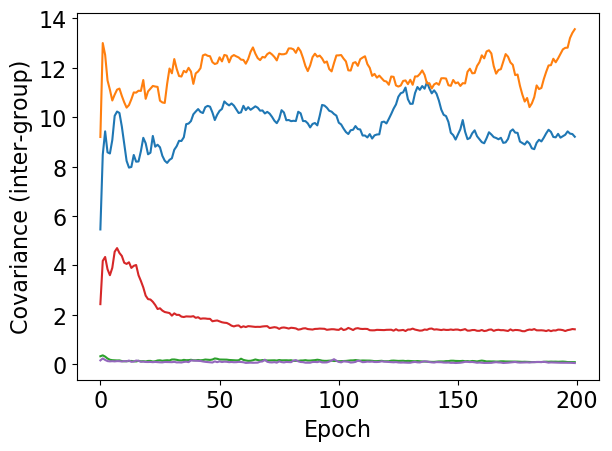}
		\includegraphics[width=0.3253\textwidth]{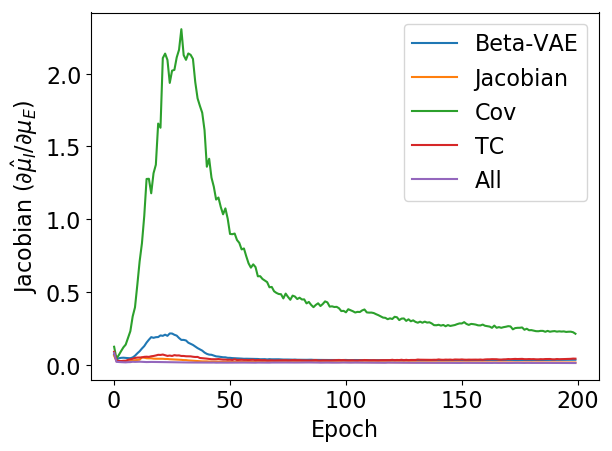}
		\includegraphics[width=0.3253\textwidth]{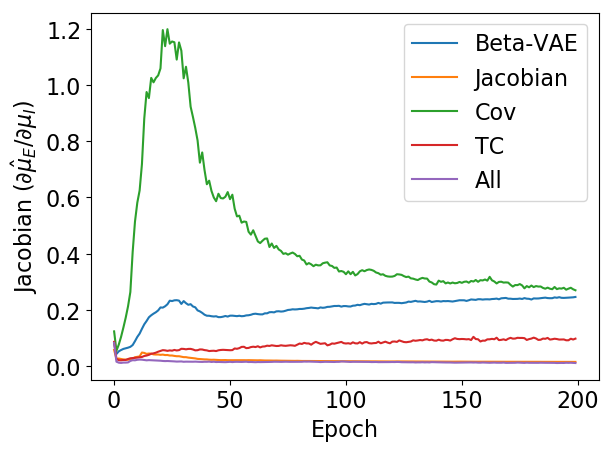}
		\caption{Empirical curves of loss terms during training across weight hyper-parameters on model learning. 
				See Table \ref{table_hypertest} for weight values.
				Each colored curve represents a different set of weight values (i.e., model hyper-parameters).
				Top row: log-likelihood reconstruction loss, intra-group TC, dimension-wise KL divergence.
				Middle row: off-diagonal intra-group covariance, on-diagonal covariance terms (i.e., variances), inter-group TC.
				Bottom row: inter-group covariance, Jacobian penalty (intrinsics with respect to extrinsics), Jacobian penalty (extrinsics with respect to intrinsics).
			}
		\label{hyperParams}
	\end{figure*}


\section{MNIST Disentanglement}
\label{supp_section:mnist_disent}

\subsection{MNIST Latent Interpolations}

Compared to that of articulating shapes (e.g., humans or other animal), the geometric disentanglement of the latent MNIST digit representation is less intuitive.
Often, however, we found that moving in $z_E$ would tend to deform the digit in ``style'', 
    while tranversing $z_I$ would more affect digit scale/thickness and identity.
Some examples are shown in Figure \ref{mnist_interps}.
For instance, in the first inset, moving in $z_E$ simply shifts around the lines of the `8' without changing its digit identity (deforming it stylistically), while moving in $z_I$ horizontally squishes the `8' into a `1'.

\subsection{MNIST Classification Results}

\begin{figure*}
	\centering
	\includegraphics[width=0.33\textwidth,trim={1.65cm 0.75cm 0 0},clip]{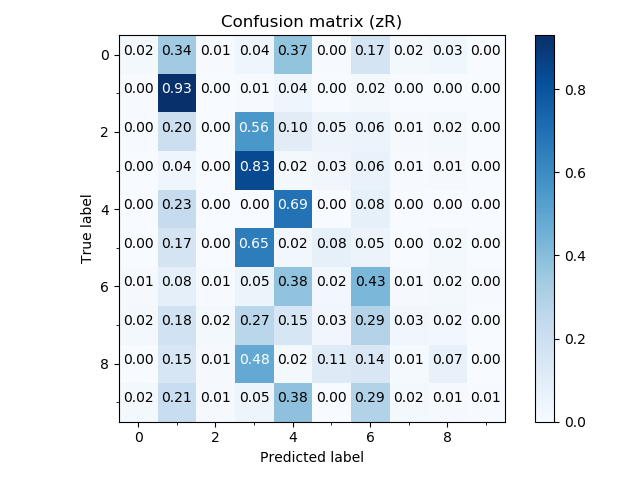}
	\includegraphics[width=0.33\textwidth,trim={1.65cm 0.75cm 0 0},clip]{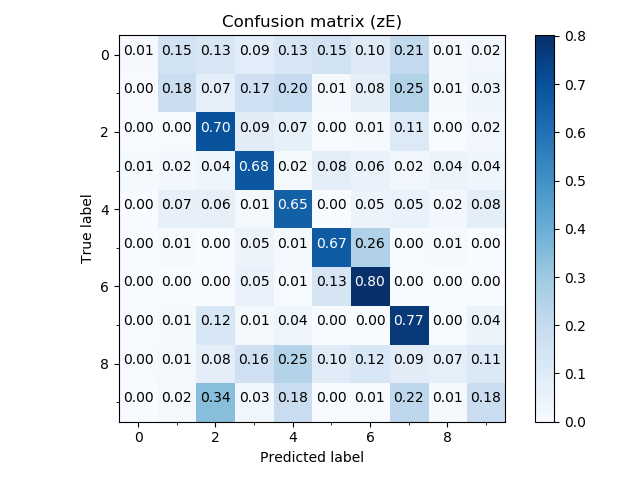}
	\includegraphics[width=0.33\textwidth,trim={1.65cm 0.75cm 0 0},clip]{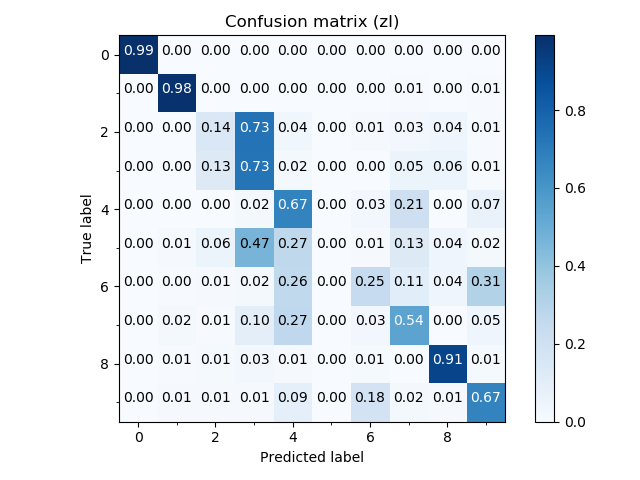} \\
	\includegraphics[width=0.33\textwidth,trim={1.65cm 0.75cm 0 0},clip]{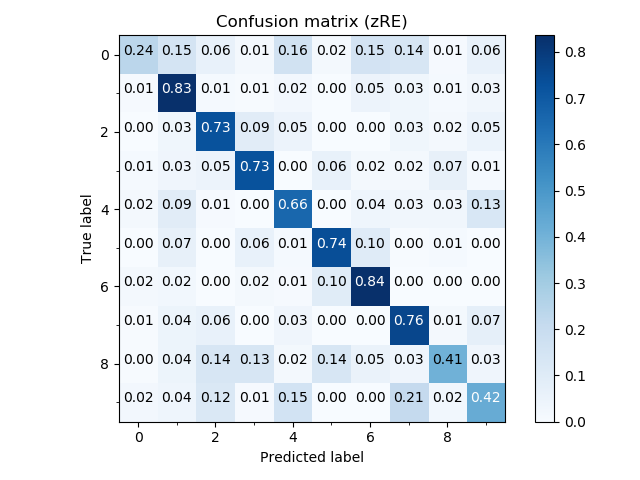} 
	\includegraphics[width=0.33\textwidth,trim={1.65cm 0.75cm 0 0},clip]{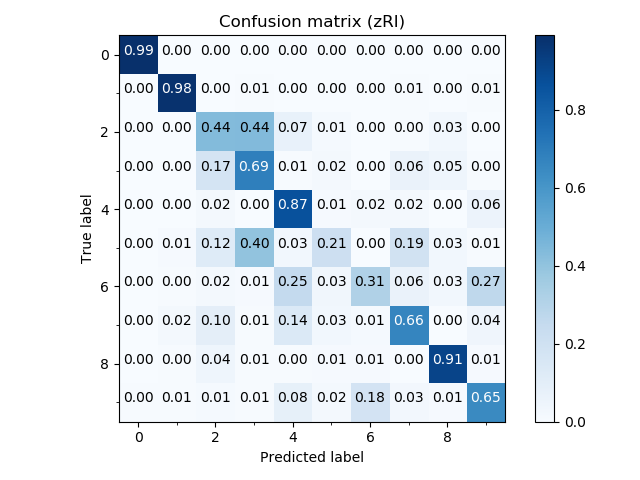}
	\includegraphics[width=0.33\textwidth,trim={1.65cm 0.75cm 0 0},clip]{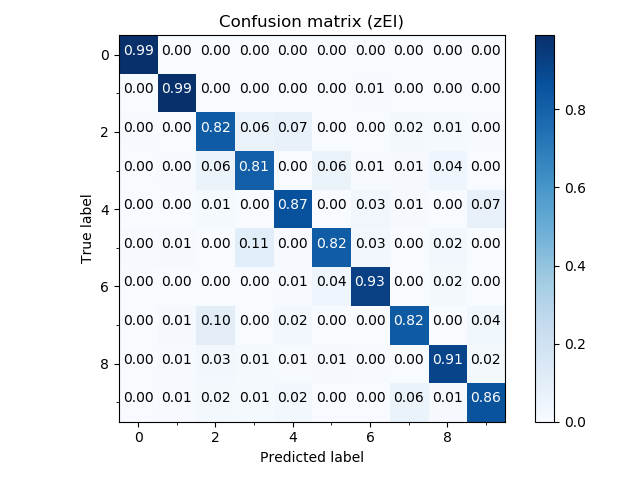} \\
	\includegraphics[width=0.33\textwidth,trim={1.65cm 0.75cm 0 0},clip]{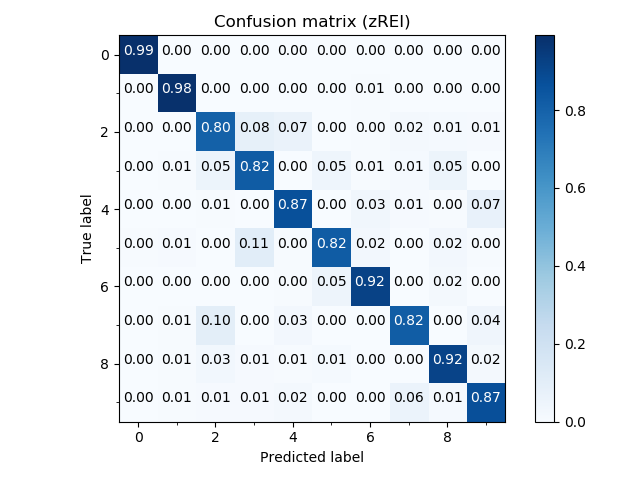}
	\includegraphics[width=0.33\textwidth,trim={1.65cm 0.75cm 0 0},clip]{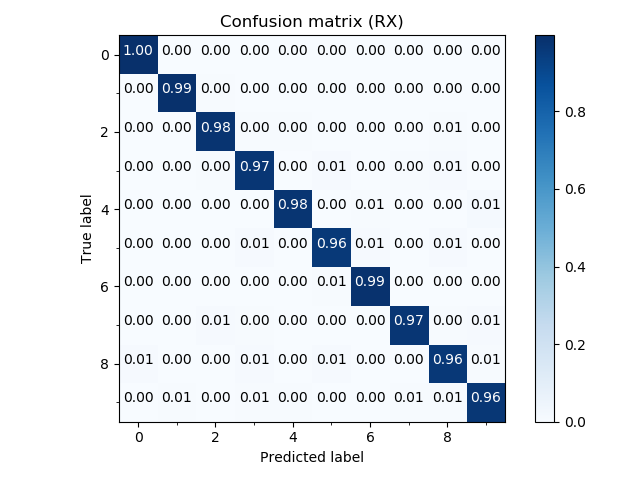}
	\caption{
	    Confusion matrices for a linear SVM classifier accuracy on the MNIST test set. 
		Top row: testing with $z_R$, $z_E$, $z_I$.
		Middle row: testing with $z_{RE}$, $z_{RI}$, $z_{EI}$.
		Bottom row: testing with, $z_{REI}$, $(R,X)$. 
	}
	\label{supp:fig:confmats}
\end{figure*}

We considered the information stored in the disentangled latent space segments by examining the performance of a classifier trained on various combinations of them, as representations of the digits.
We show the confusion matrices for a linear SVM classifier on these sub-components of the latent spaces in Figure~\ref{supp:fig:confmats}.
Notice that utilizing $z_R$ only has the poorest performance (and that adding it to $z_{EI}$ has little effect), and that the most prominent mistakes when using only $z_I$ are very similar to those in Figure~\ref{fig:mnistSpectra} 
(e.g., mixing up 2, 3, 4; and 6, 9).


\section{Rotation Disentanglement}
\label{supp_section:ae_rot}

\begin{figure*}
	\includegraphics[height=1.02in]{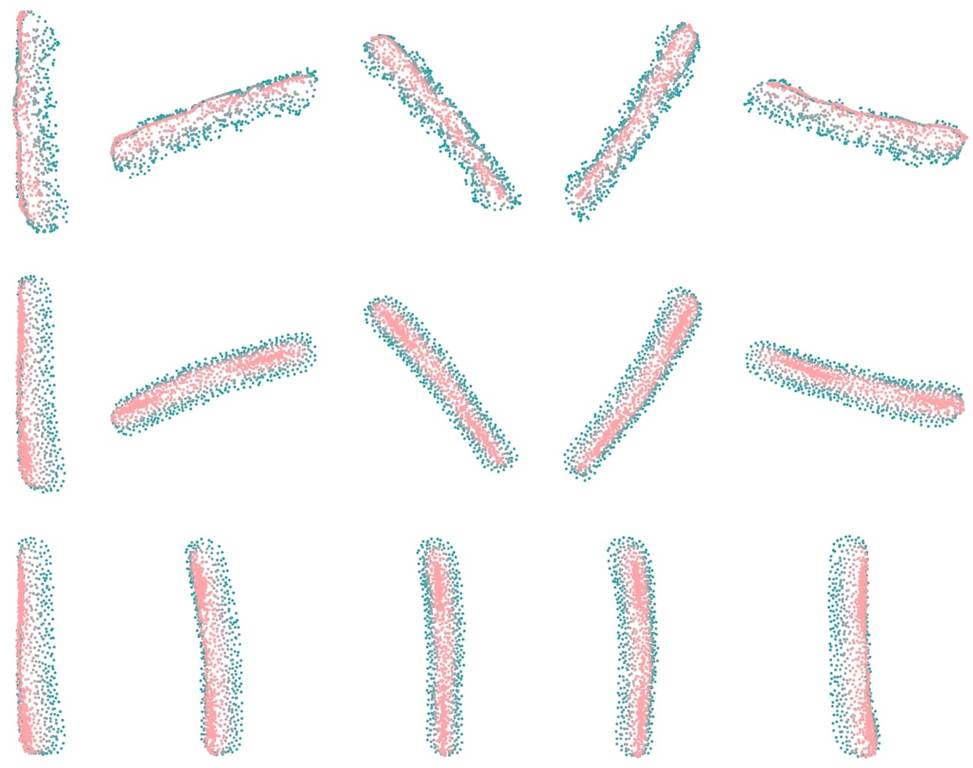}\hfill
	\includegraphics[height=1.02in]{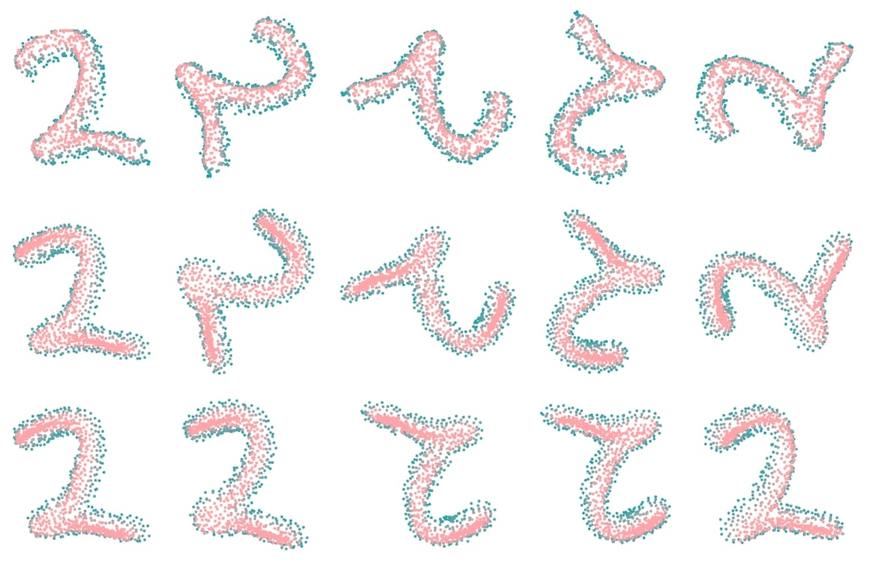}\hfill
	\includegraphics[height=1.02in]{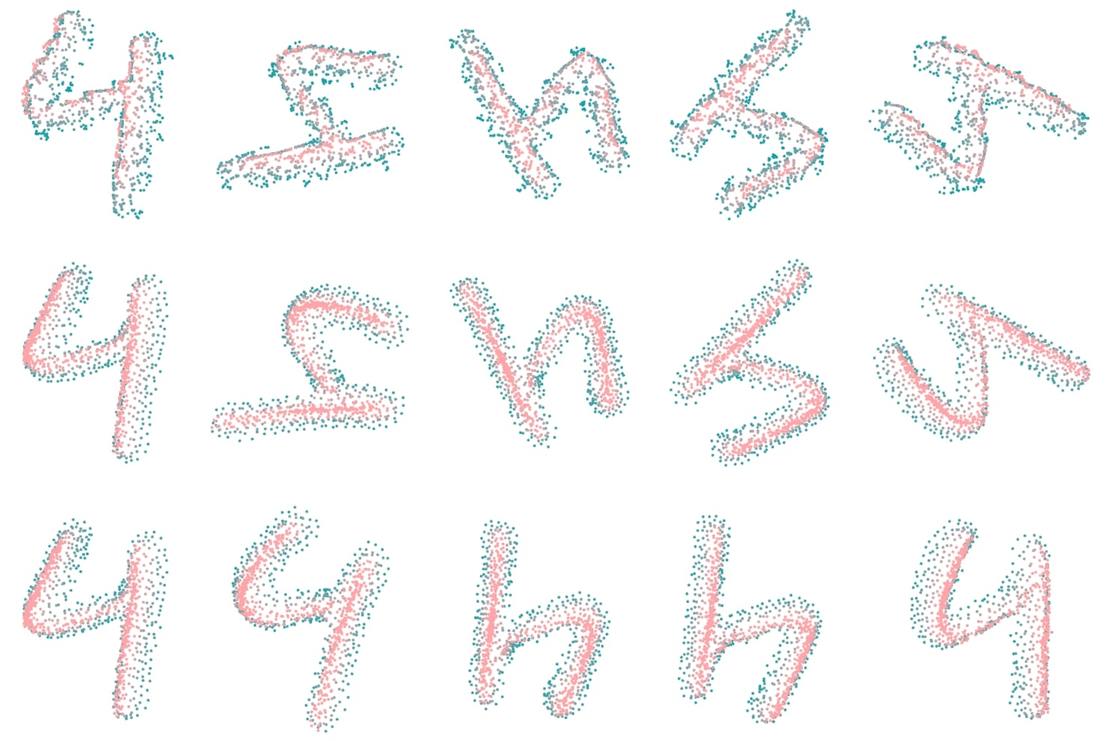}\hfill
	\includegraphics[height=1.02in]{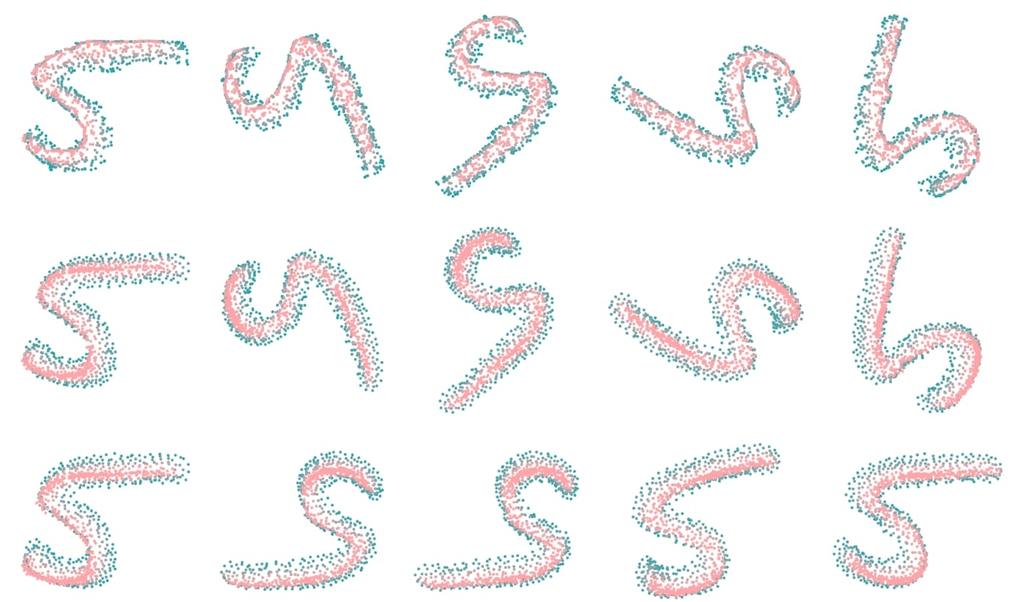} \\
	\includegraphics[height=1.02in]{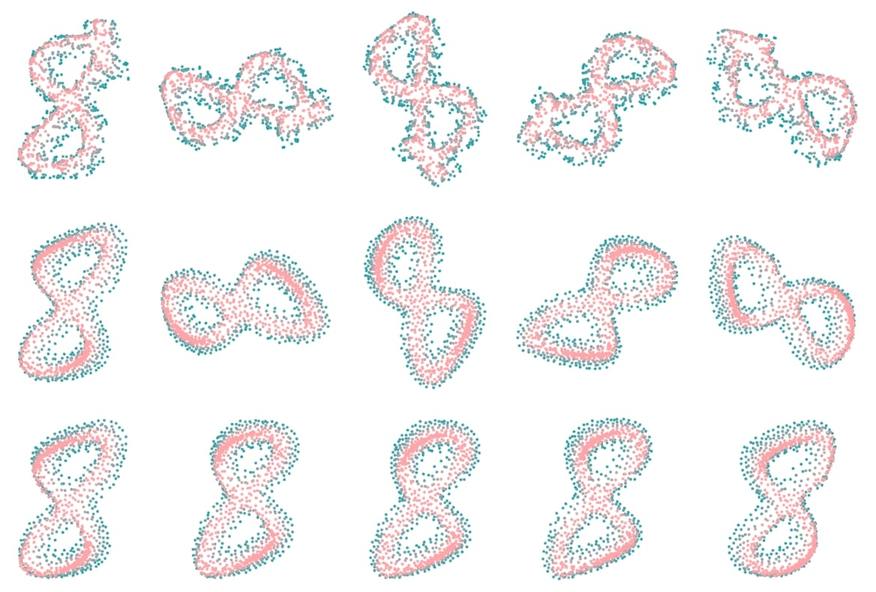}\hfill
	\includegraphics[height=1.02in]{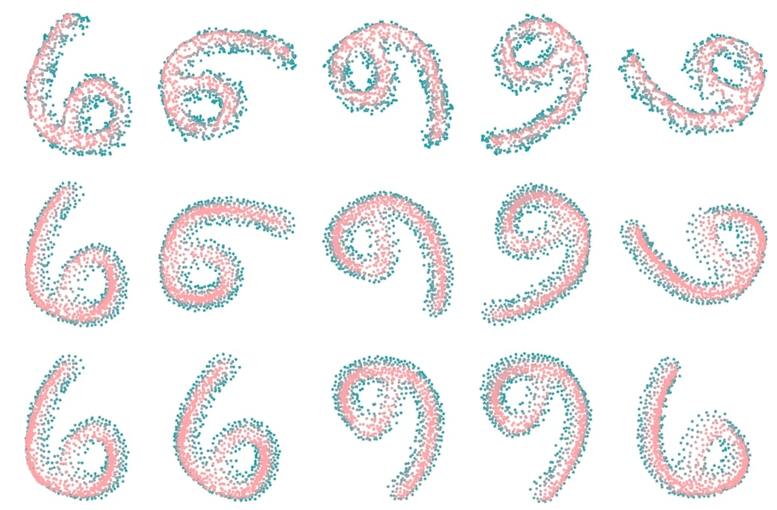}\hfill
	\includegraphics[height=1.02in]{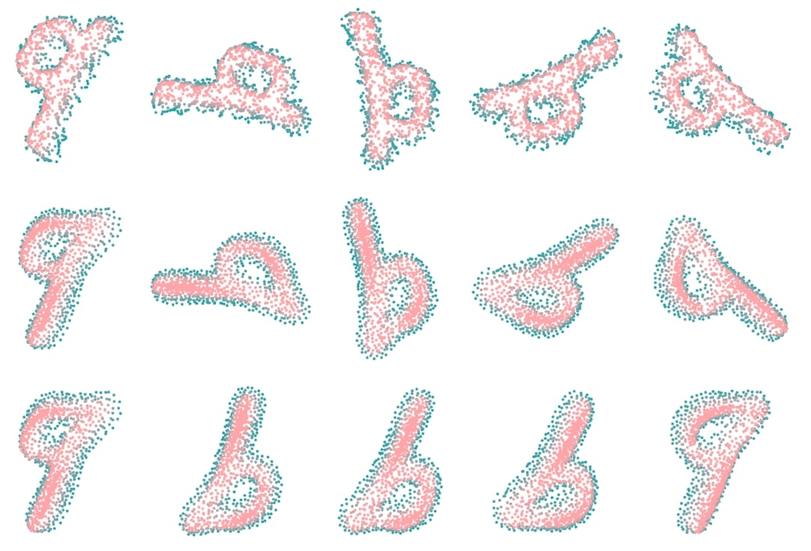}\hfill
	\includegraphics[height=1.02in]{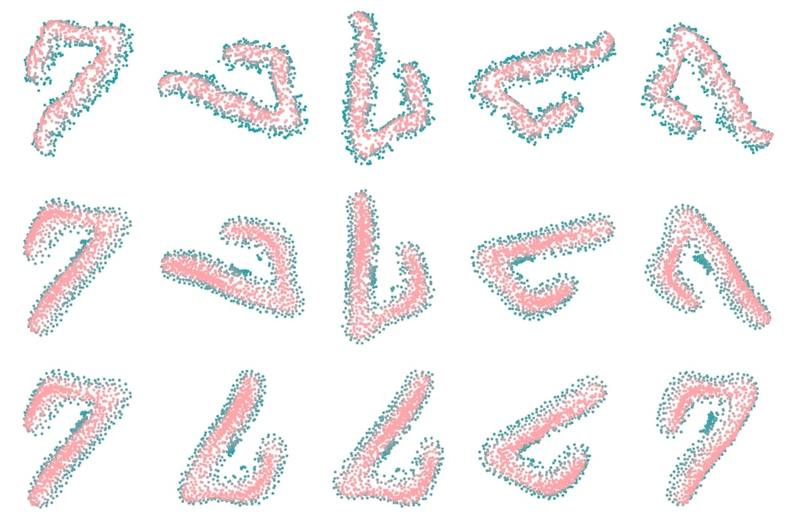} \\
	\includegraphics[height=1.34in]{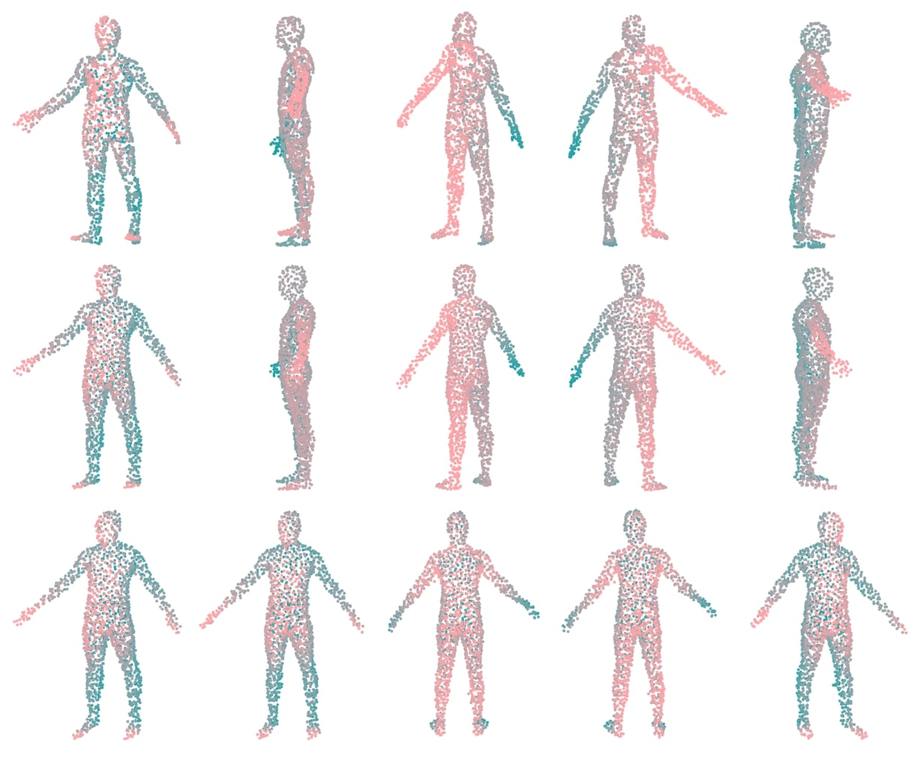}\hfill
	\includegraphics[height=1.34in]{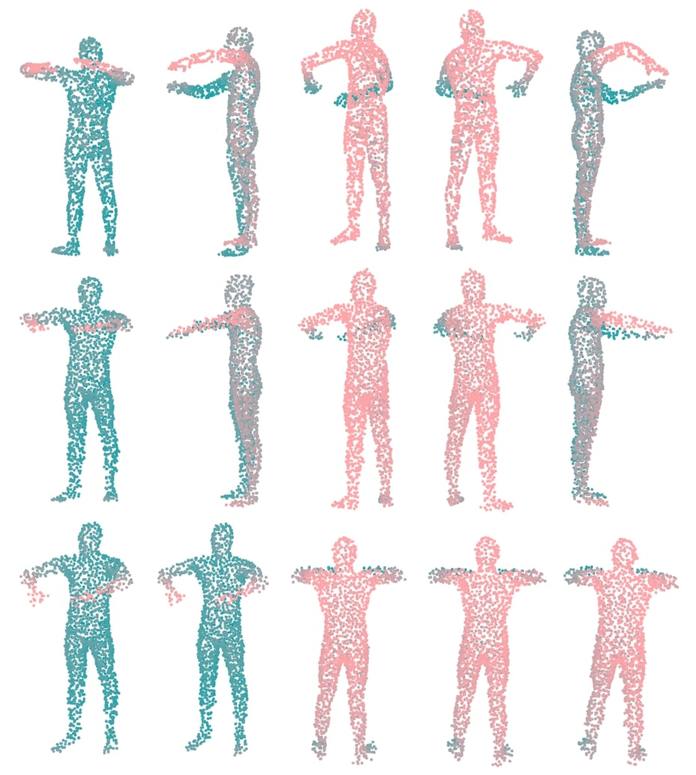}\hfill
	\includegraphics[height=1.34in]{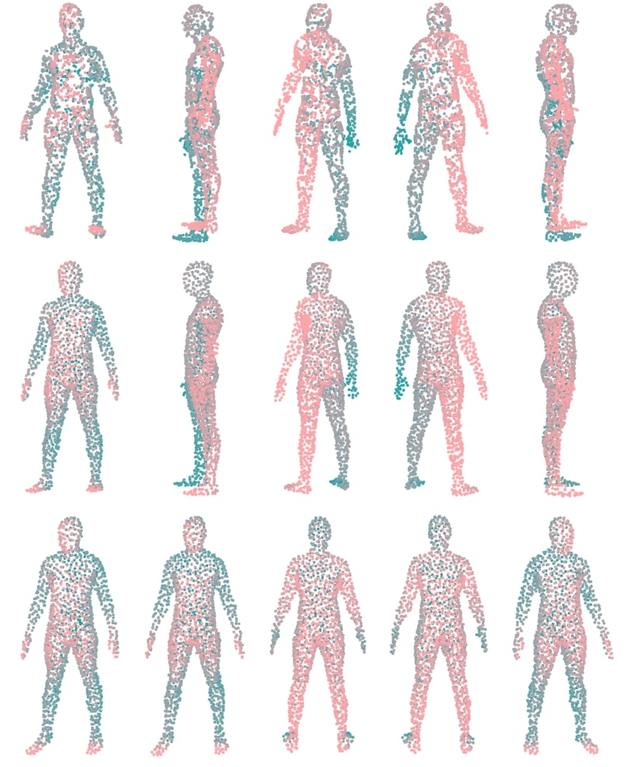}\hfill
	\includegraphics[height=1.34in]{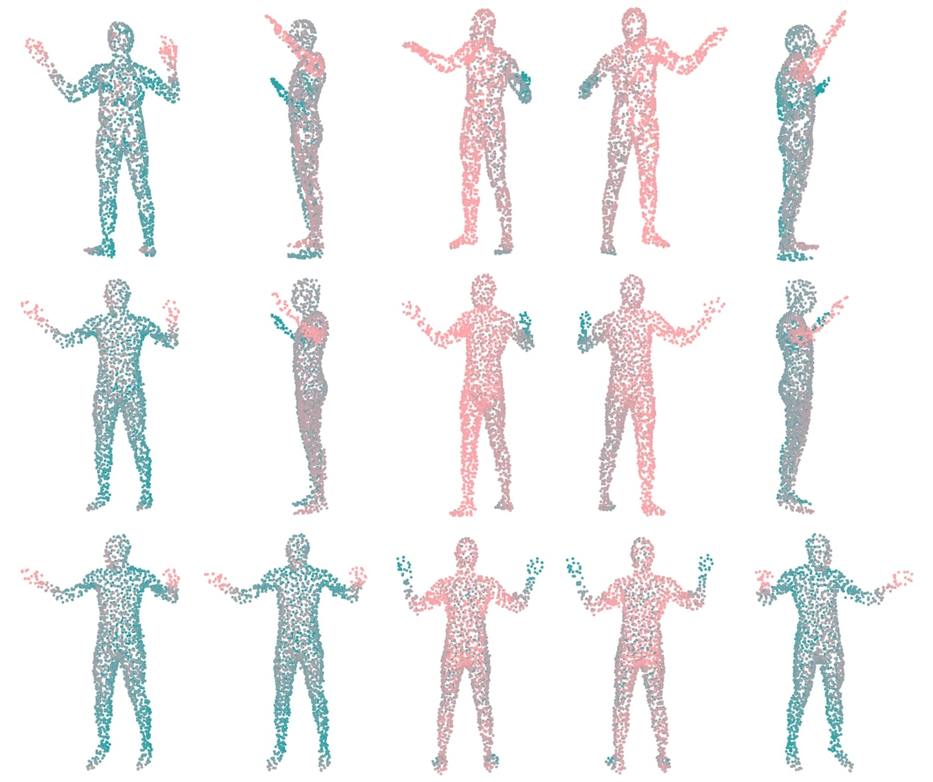}\hfill
	\includegraphics[height=1.34in]{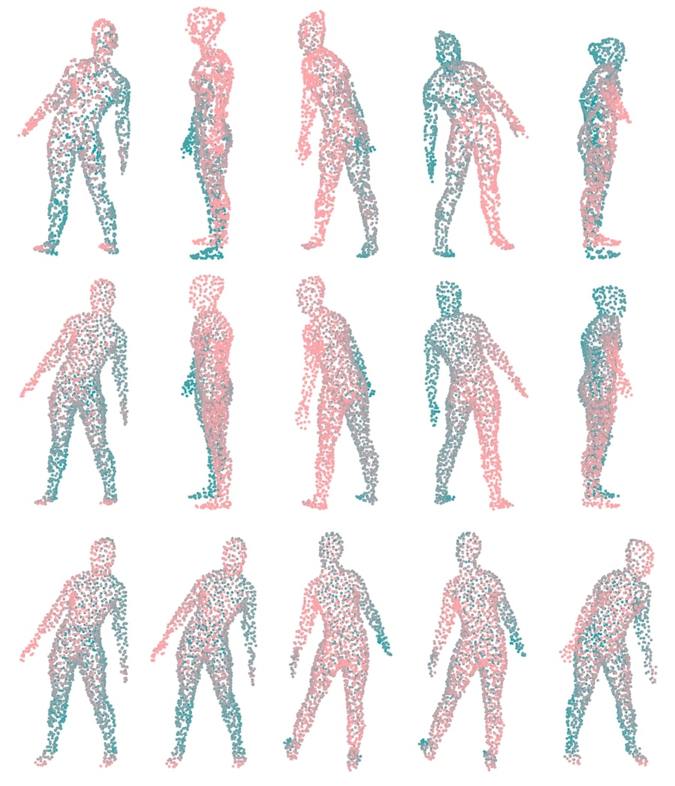} \\
	\includegraphics[height=1.42in]{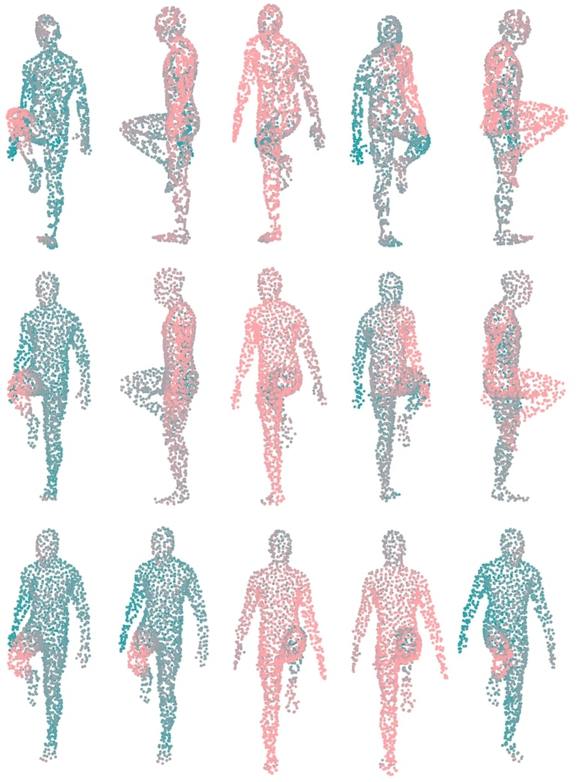}\hfill
	\includegraphics[height=1.42in]{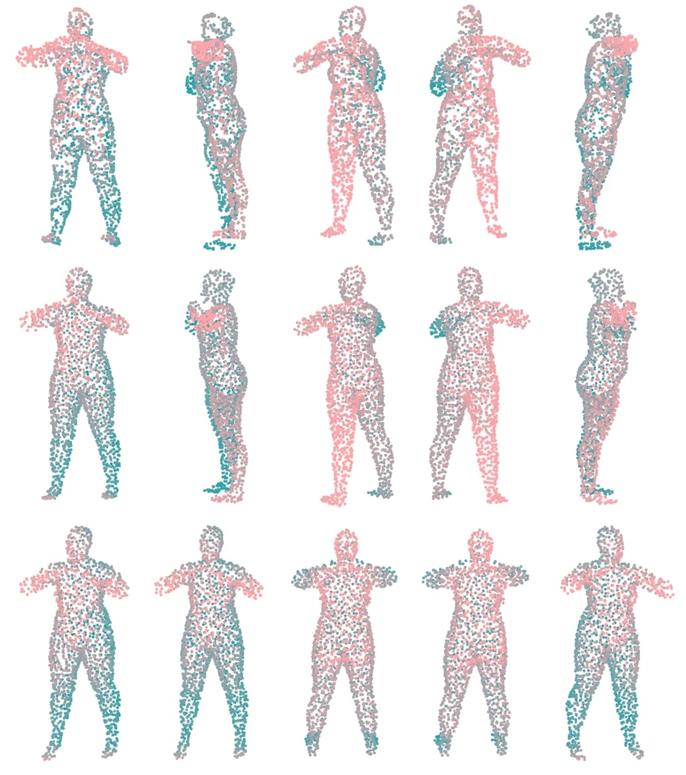}\hfill
	\includegraphics[height=1.42in]{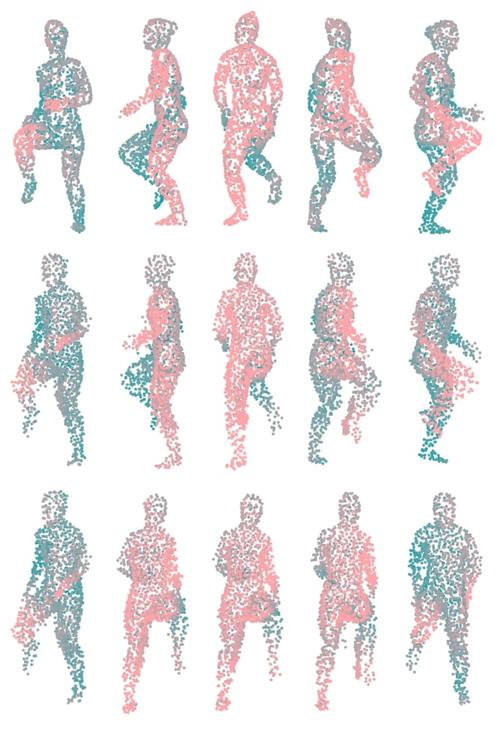}\hfill
	\includegraphics[height=1.42in]{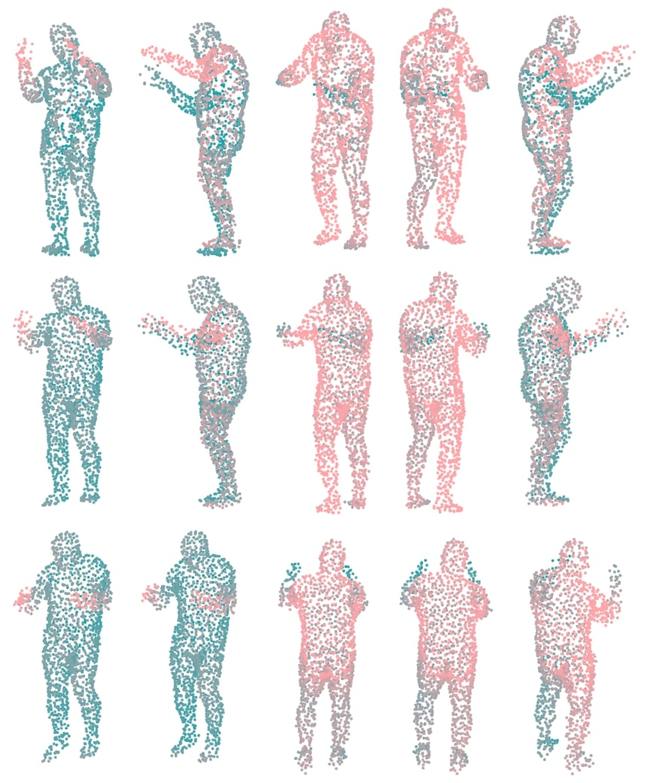}\hfill
	\includegraphics[height=1.42in]{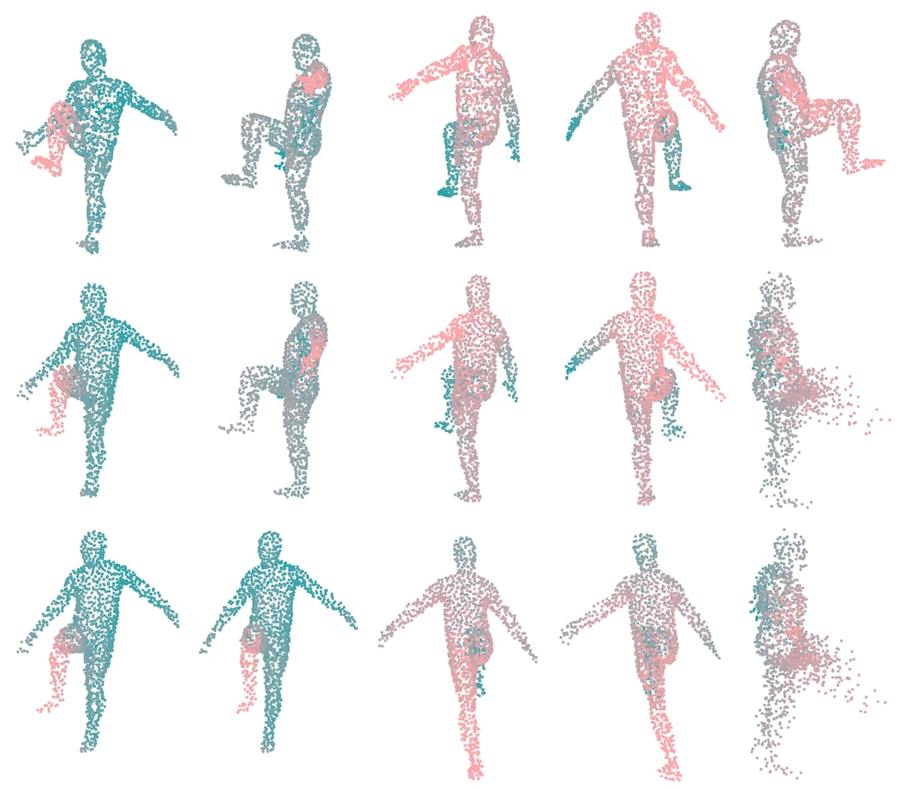} \\
	\caption{
		Examples of autoencoder reconstructions across different rotations of the same object. 
		Note that the encoding of the datum from the AE is not passed through the VAE.
		Colour denotes depth ($z$ value).
		Within each group of shapes, 
		the columns show different rotations of the left-most shape,
		while
		the top row shows input shapes (i.e., $P$, from the original data), the middle row shows the reconstruction $\hat{P}$, and the bottom row shows the derotated shape with the rotation component $R$ set to the value given by the shape in the first column. Note that these models were trained with data augmentations across all rotations about the gravity axis.	}
	\label{ae_rot_fig}
\end{figure*}

We examine the performance of the deterministic autoencoder,
visualizing a number of reconstructions in Figure \ref{ae_rot_fig}.
Qualitatively, the points sampled over the reconstructions are largely uniformly spread out. However, thin or protruding areas tend to have lower densities of points, an issue identified by other works \cite{achlioptas2018learning}.

\begin{figure*}
	\begin{minipage}[c]{0.66\textwidth}
		\vspace{0pt}
		\includegraphics[height=1.65in]{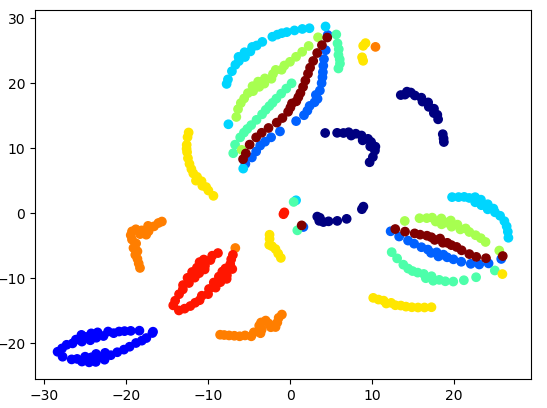}\hfill%
		\includegraphics[height=1.65in]{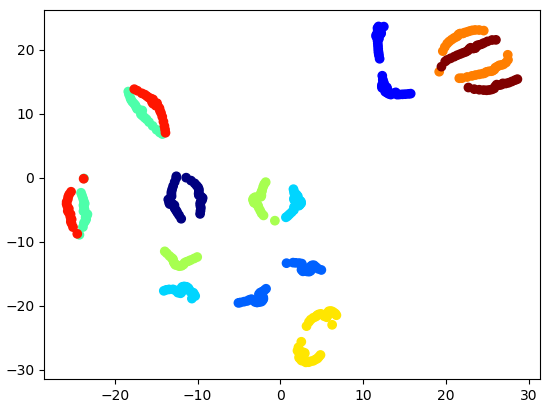}%
	\end{minipage}\hfill
	\begin{minipage}[c]{0.3329\textwidth}
		\vspace{0pt}
		\caption{
			A t-SNE visualization of the rotated shapes from the deterministic AE.
			Left: plot of MNIST.
			Right: plot of Dyna.
			In both plots, 10 random shapes are selected, and rotated 40 times, evenly spread over $[0,2\pi]$, with respect to the height axis.
			Each shape is then encoded by the AE, and only the $X$ component of its representation is plotted.
			A single color corresponds to a single shape.
		} \label{supp:rot_figs}
	\end{minipage}
\end{figure*}

One question is how the presence of the rotation quaternion affects the representation.
We visualize this by ``derotating'' the shapes (shown in the last row of each set of shapes in Figure \ref{ae_rot_fig}), where the $R$ portion of their representation is set to the same value. 
We can see that the derotated shapes tend to approximately fall into two or three groups with similar orientation.
To confirm this, we also sample random shapes, rotate them, 
and then embed their $X$ encodings via t-SNE 
in Figure \ref{supp:rot_figs}.
Qualitatively, we can see that rotating the point sets does not appear to lead to a single representation $X$; instead, it forms a small number of latent groups.

We also tried adding a ``rotational consistency'' penalty to the loss function, which was computed by placing an $L_2$ loss between the $X$ representation of pairs of shapes that differ only by a rigid rotation (i.e., $ || X - X_r ||_2^2 $, where $X$ and $X_r$ are generated by rotations of the same point cloud).
However, this seemed to lead to higher reconstruction error and did not entirely prevent the differing representations forming across rotations.
Furthermore, as noted in the main text, some shapes can naturally become a ``new'' shape, semantically speaking, due to rotation (e.g., a 9 becoming a 6), which this penalty does not allow.


%

\vspace*{40pt}

\section{Model Analysis}

\begin{table}
    \centering
    \begin{tabular}{cc|cccc}
         $\,$ & $\,$       & $X$  & $z$  & $z_E$ & $z_I$ \\\hline
         			\rule{0pt}{0.51\normalbaselineskip}
         \multirow{2}{*}{SMAL} & $E_\beta$  & 0.641 & 0.743 & 0.975  & 0.645 \\
                               & $E_\theta$ & 0.938 & 0.983 & 0.983  & 0.993 \\\hline
         \multirow{2}{*}{SMAL-NJ} & $E_\beta$  & 0.642 & 0.829 & 0.980  & 0.734 \\
                                  & $E_\theta$ & 0.938 & 0.979 & 0.980  & 0.996\\\hline
         \multirow{2}{*}{SMAL-NC} & $E_\beta$  & 0.642 & 0.670 & 0.962  & 0.656 \\
                                  & $E_\theta$ & 0.938 & 0.966 & 0.969  & 0.991 \\\hline
         \multirow{2}{*}{SMAL-NJC} & $E_\beta$  & 0.642 & 0.661 & 0.834  & 0.891 \\
                                   & $E_\theta$ & 0.938 & 0.967 & 0.978  & 0.982 \\\hline
         \multirow{2}{*}{SMPL} & $E_\beta$  & 0.856 & 0.922 & 0.997  & 0.928 \\
                                & $E_\theta$ & 0.577 & 0.726 & 0.709  & 0.947 \\\hline
         \multirow{2}{*}{SMPL-NJ} & $E_\beta$  & 0.858 & 0.907 & 1.006  & 0.895 \\
                               & $E_\theta$ & 0.578 & 0.695 & 0.812  & 0.908 \\\hline
         \multirow{2}{*}{SMPL-NC} & $E_\beta$  & 0.855 & 0.908 & 0.995  & 0.905 \\
                               & $E_\theta$ & 0.578 & 0.727 & 0.836  & 0.909 \\\hline
         \multirow{2}{*}{SMPL-NJC} & $E_\beta$  & 0.857 & 0.888 & 0.992  & 0.921 \\
                                & $E_\theta$ & 0.578 & 0.693 & 0.722  & 0.965 
    \end{tabular}
    \caption{
        Different retrieval scores under various disentanglement penalty ablation conditions.
        NJ, NC, and NJC mean no Jacobian, no covariance, and neither Jacobian nor covariance cases respectively.
        Note that differences between scores under $X$ are due to the random samplings of points from the shape; hence, each score is obtained by running the process three times (across point samplings).
        However, only the scores for SMAL and SMPL are averaged over multiple training runs.
    }
    \label{table:ablations}
\end{table}	
	
\subsection{Hyper-parameter Variation} \label{sec:experiments:hyper}
	Our generative model must balance three main terms: 
	(1) autoencoding reconstruction, 
	(2) latent prior sampling, 
	and
	(3) disentanglement. 
	We therefore trained five models on MNIST under five different hyper-parameter conditions 
	(see Table \ref{table_hypertest}) to showcase the relative effect of these loss weights.
	The choice of models was designed to check how the different  penalties affected the metrics of disentanglement based on each loss (e.g., how penalizing covariance affects the pairwise Jacobian penalty), as well as autoencoding and generation.
	
	Based on the results (loss curves are shown in Figure \ref{hyperParams}), we can make several observations. 
	Firstly, a high TC penalty results in decreased inter-group covariance and Jacobian losses.
	However, while the covariance and Jacobian penalties effectively reduce their own penalties, 
		they struggle to reduce the TC or each other.
	Nevertheless, 
		the TC penalty alone does not drive the covariance or Jacobian values as low as having a penalty on them directly does. 
	This suggests that the TC may be a more powerful penalty, 
		but it can still be complemented by the other approaches.
	One explanation for this effect is that the Jacobian penalty is fundamentally local (penalizing the expected change with respect to an infinitesimal perturbation around each data point separately) and the covariance penalty only reduces linear correlations, whereas the TC penalty is information-theoretic (i.e., able to detect non-linear relations) and considers the estimated latent probability distributions on a more global level.
	
	On the other hand, the models with high TC show the worst log-likelihood for reconstructions. 
	The model with high weights for all terms also has poor dimension-wise KL divergence, meaning the ability to generate novel samples may be compromised, though it is not as high as that of the covariance-penalized model. 
	Though our observations are limited to this dataset, hyper-parameters, and architecture, they suggest that the disentanglement term can be in conflict with reconstruction and sampling 
	(just as the latter two are known to be in conflict with each other; e.g., \cite{higgins2017beta}).

\subsection{Disentanglement Penalty Ablation}

\begin{figure*}
		\centering
		\includegraphics[width=0.3253\textwidth]{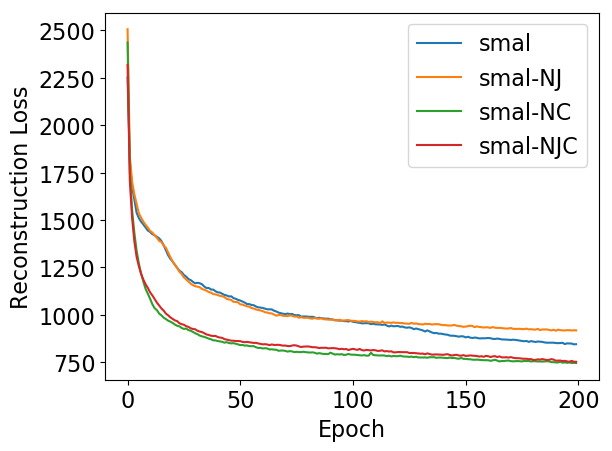}
		\includegraphics[width=0.3253\textwidth]{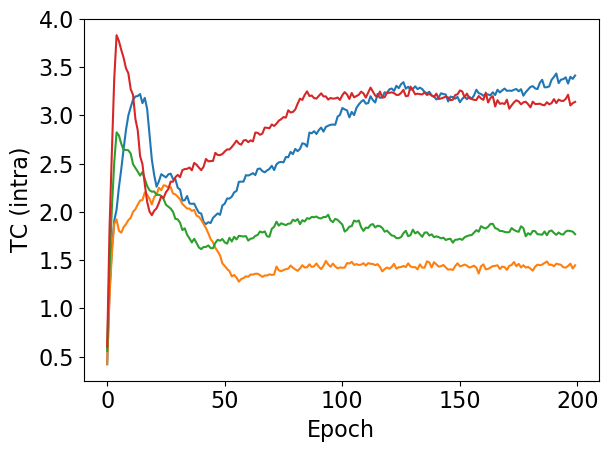}
		\includegraphics[width=0.3253\textwidth]{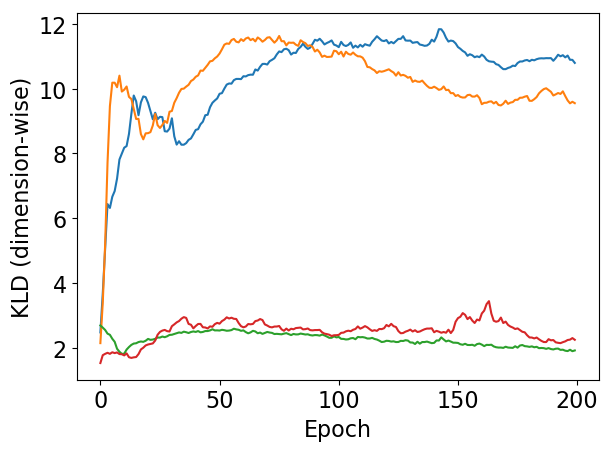}\\
		\includegraphics[width=0.3253\textwidth]{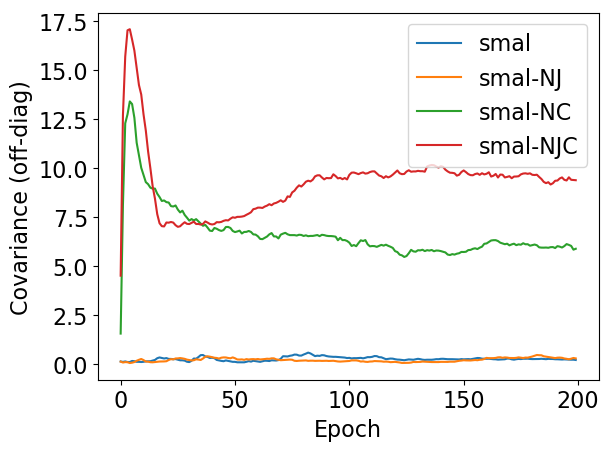}
		\includegraphics[width=0.3253\textwidth]{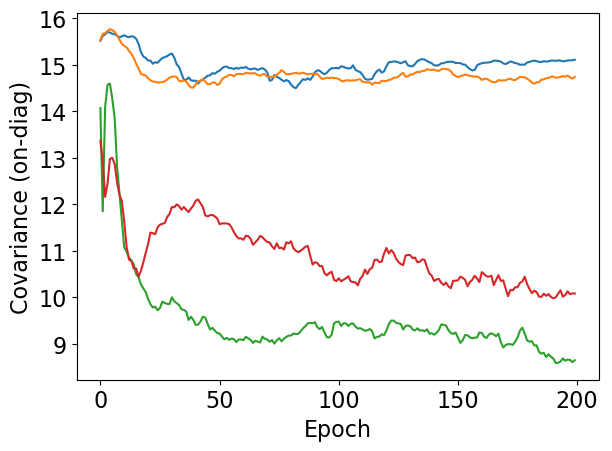}
		\includegraphics[width=0.3253\textwidth]{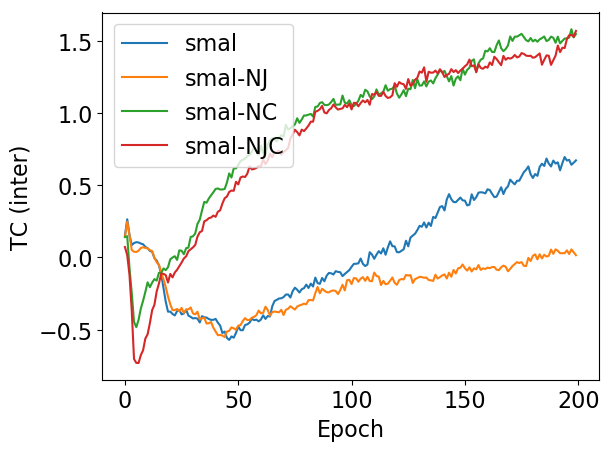}\\
		\includegraphics[width=0.3253\textwidth]{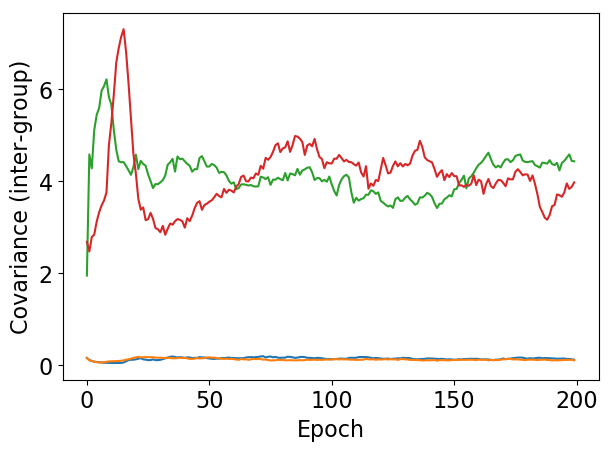}
		\includegraphics[width=0.3253\textwidth]{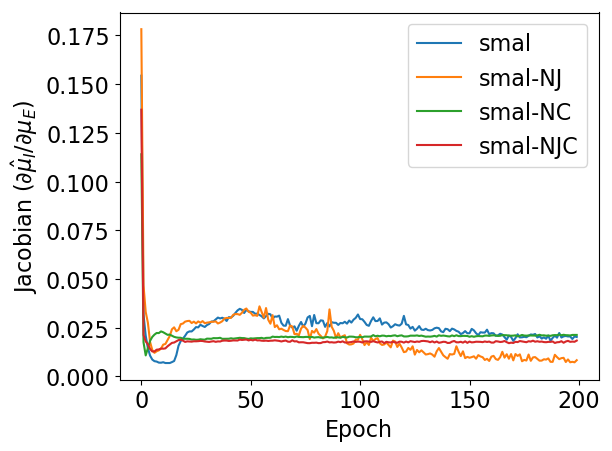}
		\includegraphics[width=0.3253\textwidth]{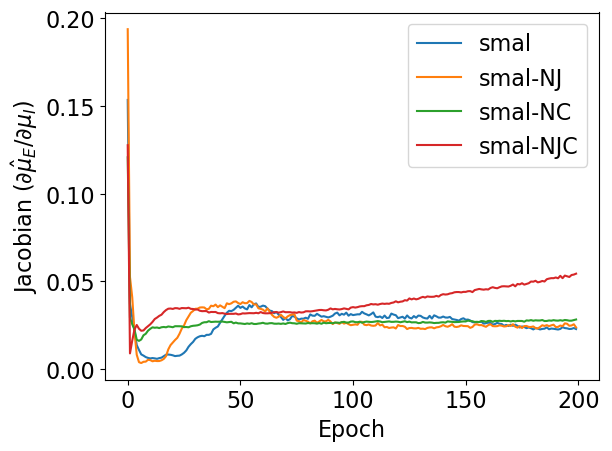}
		\caption{
		        Empirical curves of loss terms for disentanglement ablations on SMAL. 
				NC, NJ, and NJC mean no covariance, no Jacobian, or neither penalties, respectively.
				Each colored curve represents a different set of weight values (i.e., model hyper-parameters).
				Top row: log-likelihood reconstruction loss, intra-group TC, dimension-wise KL divergence.
				Middle row: off-diagonal intra-group covariance, on-diagonal covariance terms (i.e., variances), inter-group TC.
				Bottom row: inter-group covariance, Jacobian penalty (intrinsics with respect to extrinsics), Jacobian penalty (extrinsics with respect to intrinsics).
			}
		\label{smal-curves}
	\end{figure*}

	\begin{figure*}
		\centering
		\includegraphics[width=0.3253\textwidth]{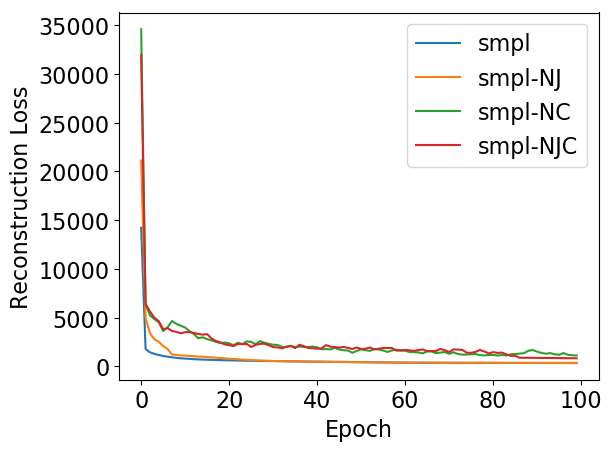}
		\includegraphics[width=0.3253\textwidth]{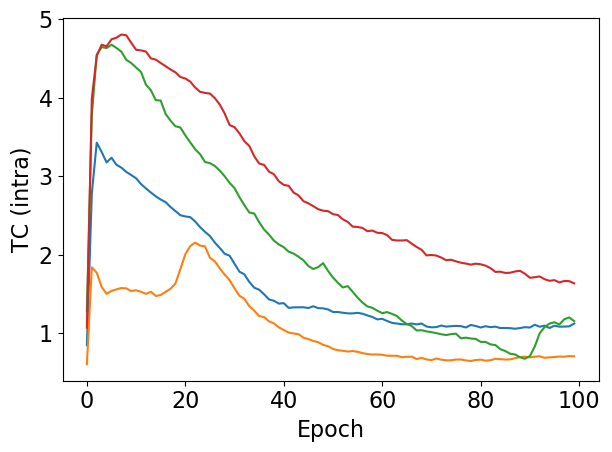}
		\includegraphics[width=0.3253\textwidth]{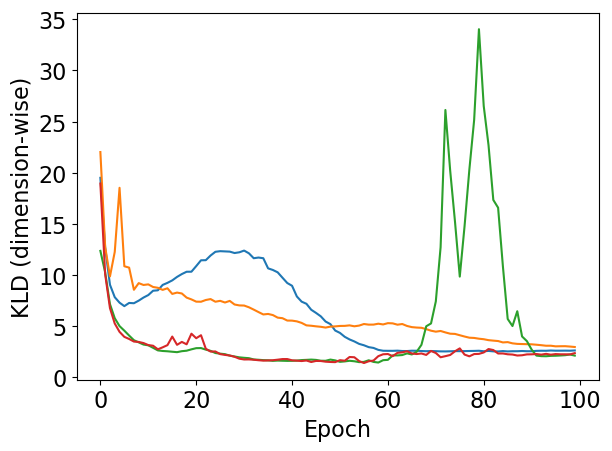}\\
		\includegraphics[width=0.3253\textwidth]{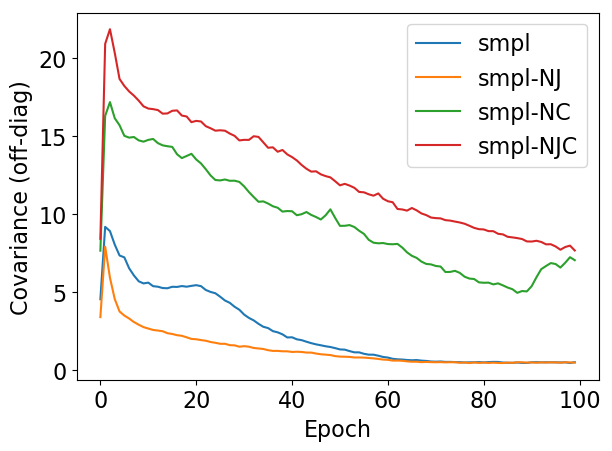}
		\includegraphics[width=0.3253\textwidth]{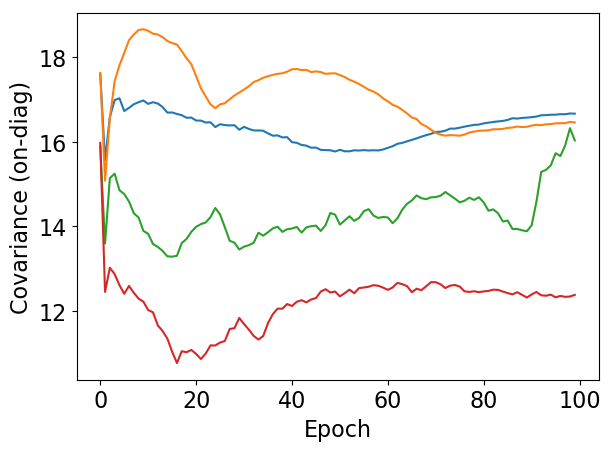}
		\includegraphics[width=0.3253\textwidth]{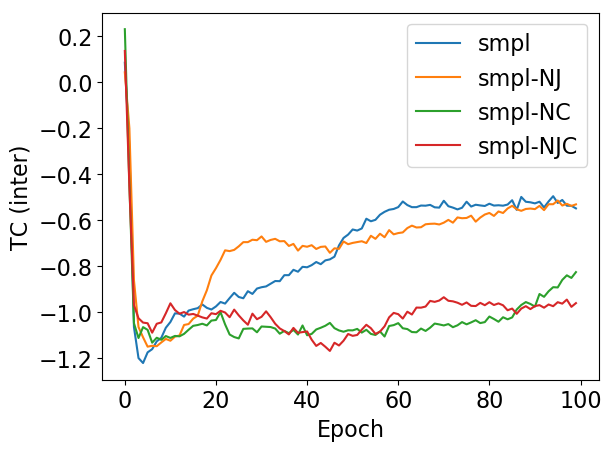}\\
		\includegraphics[width=0.3253\textwidth]{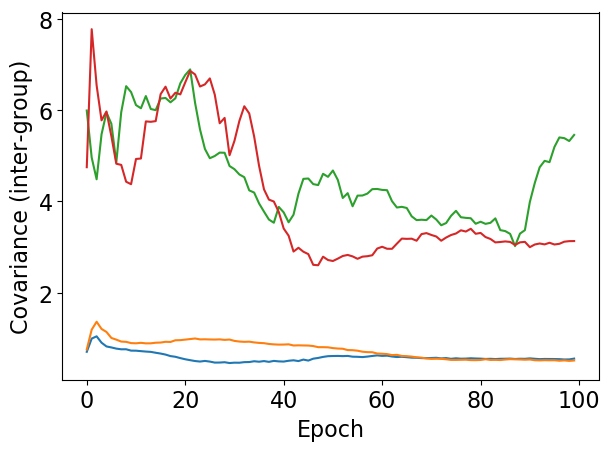}
		\includegraphics[width=0.3253\textwidth]{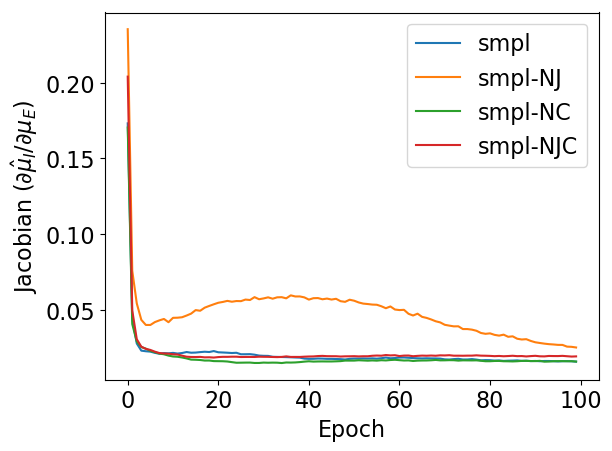}
		\includegraphics[width=0.3253\textwidth]{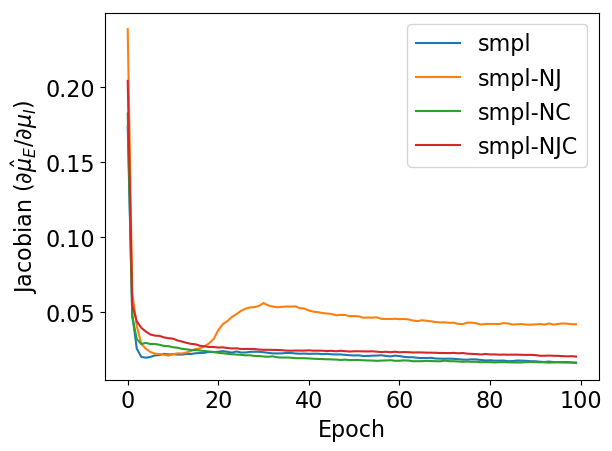}
		\caption{Empirical curves of loss terms for disentanglement ablations on SMPL. 
				NC, NJ, and NJC mean no covariance, no Jacobian, or neither penalties, respectively.
				Top row: log-likelihood reconstruction loss, intra-group TC, dimension-wise KL divergence.
				Middle row: off-diagonal intra-group covariance, on-diagonal covariance terms (i.e., variances), inter-group TC.
				Bottom row: inter-group covariance, Jacobian penalty (intrinsics with respect to extrinsics), Jacobian penalty (extrinsics with respect to intrinsics).
			}
		\label{smpl-curves}
	\end{figure*}

To further examine the effect of the disentanglement penalties, in particular the covariance and Jacobian terms, we considered two ablation experiments on the SMPL and SMAL datasets:
    (1) looking at the loss curves of the estimated entanglement measures during training 
    and
    (2) considering the pose-aware retrieval performance.
We examined three conditions, in addition to the regular hyper-parameter values (REG): no Jacobian penalty (NJ), no covariance penalty (NC), and no Jacobian or covariance penalties (NJC).

The various loss terms are shown over training epochs in Figures \ref{smal-curves} and \ref{smpl-curves}, 
    for SMAL and SMPL respectively.
On SMAL, we see that REG and NJ have worse KL-divergence and reconstruction loss, but better inter-group TC and covariance. NJC scores the worst on all four entanglement measures, except $\partial \hat{\mu}_I/\partial \mu_E$ (where surprisingly NJ holds the smallest value).
On SMPL, we see that NJ consistently has the highest Jacobian penalties, and that REG and NJ have the worst TC. 
On both datasets, we see that the covariance penalty is necessary to ensure the inter-group covariance is small, since NC and NJC always have much higher inter-group covariance than NJ and REG.

We next considered the pose-aware retrieval task under the various ablation conditions.
Results are shown in Table \ref{table:ablations}.
Recall that low $E_\theta$ using $z_E$ and low $E_\beta$ using $z_I$ are good (indicating $z_I$ and $z_E$ hold intrinsic shape and pose respectively), 
while low $E_\theta$ using $z_I$ and low $E_\beta$ using $z_E$ are not (as it means entanglement is present).
For SMAL, we see that REG has the best $E_\beta$ using $z_I$, while NJC is worse than both NJ and NC. We can also see poor entanglement in the low $E_\beta$ using $z_E$ of NJC.
The scores for $E_\theta$ are poor across all the representations, even including $X$. 
NC has the best $E_\theta$ using $z_E$, but it has a worse (i.e., lower) $E_\beta$ using $z_E$, 
suggesting increased entanglement.
For SMPL, REG has the best $E_\theta$ using $z_E$, 
    but NC and NJ have better $E_\beta$ values with $z_I$. 
However, NC and NJ also have lower $E_\theta$ with $z_I$, meaning pose is entangled with intrinsic shape. 
It's also worth noting that NJC on both SMAL and SMPL has lower error on both $E_\beta$ and $E_\theta$ when using $z$.
This suggests that the increased disentanglement penalties make holding information harder in general (i.e., allowing more entanglement can increase reconstruction and thus retrieval performance).


\begin{table}[h] 
	\centering
	\begin{tabular}{cc|cccc}
		$\,$ & $\,$  & $X$   & $z$   & $z_E$  & $z_I$ \\\hline        
		SMAL    & $E_\beta$  & 0.0006 & 0.0282 & 0.0035  & 0.0381 \\
		(M-SEM) & $E_\theta$ & 0.0003 & 0.0015 & 0.0015  & 0.0015 \\\hline    
		SMPL    & $E_\beta$  & 0.0003 & 0.0048 & 0.0015  & 0.0032 \\
		(M-SEM) & $E_\theta$ & 0.0003 & 0.0032 & 0.0058  & 0.0059 \\\hline 
		SMAL    & $E_\beta$  & 0.0028 & 0.0020 & 0.0030  & 0.0024 \\
		(S-SEM) & $E_\theta$ & 0.0008 & 0.0005 & 0.0006  & 0.0008 \\\hline    
		SMPL    & $E_\beta$  & 0.0015 & 0.0013 & 0.0020  & 0.0053 \\
		(S-SEM) & $E_\theta$ & 0.0008 & 0.0017 & 0.0022  & 0.0070
	\end{tabular}
	\caption{
		SEMs across model training runs (M-SEM) and shape samplings (S-SEM).
		Model training SEMs are computed over the mean of the shape sampling runs;
		shape sampling SEMs are computed by taking the SEM over point samplings per model, and then the maximum SEM across models.
	}
	\label{table:ret_var}
\end{table}

{
	\setlength\tabcolsep{7.49pt}
	\begin{table}[h] 
		\centering
		\begin{tabular}{cc|cccc}
			$\,$ & $\,$  & $X$   & $z$   & $z_E$  & $z_I$ \\\hline        
			SMAL            & $E_\beta$ & 0.641 & 0.724 & 0.961  & 0.666 \\
			$\sigma=0.05$  & $E_\theta$ & 0.937 & 0.969 & 0.980  & 0.981 \\\hline    
			SMAL           & $E_\beta$  & 0.641 & 0.723 & 0.900  & 0.868 \\
			$\sigma=0.1$   & $E_\theta$ & 0.938 & 0.969 & 0.979  & 0.994 \\\hline 
			SMAL           & $E_\beta$  & 0.636 & 0.810 & 0.893  & 0.838 \\
			$\sigma=0.2$   & $E_\theta$ & 0.938 & 0.979 & 0.988  & 0.996 \\\hline   
			\multirow{2}{*}{SMAL-R}
			& $E_\beta$  & 0.643 & 0.822 & 0.791  & 0.913 \\
			& $E_\theta$ & 0.938 & 0.977 & 0.985  & 0.998 \\\hline
			SMAL-P         & $E_\beta$  & 0.639 & 0.629 & 0.783  & 0.929 \\
			$n_p=1.2\text{K}$ & $E_\theta$ & 0.939 & 0.973 & 0.973  & 0.995 \\\hline
			SMAL-P     & $E_\beta$  & 0.641 & 0.698 & 0.636  & 0.895 \\
			$n_p=2\text{K}$     & $E_\theta$ & 0.938 & 0.980 & 0.984  & 0.993 \\\hline 
			SMPL           & $E_\beta$  & 0.857 & 0.910 & 1.000  & 0.913 \\
			$\sigma=0.05$  & $E_\theta$ & 0.578 & 0.735 & 0.774  & 0.966 \\\hline    
			SMPL           & $E_\beta$  & 0.856 & 0.909 & 0.979  & 0.929 \\
			$\sigma=0.1$   & $E_\theta$ & 0.578 & 0.694 & 0.720  & 0.979 \\\hline 
			SMPL           & $E_\beta$  & 0.858 & 0.919 & 0.980  & 0.925 \\
			$\sigma=0.2$   & $E_\theta$ & 0.578 & 0.710 & 0.810  & 0.946 \\\hline   
			\multirow{2}{*}{SMPL-R}     
			& $E_\beta$  & 0.857 & 0.933 & 0.924  & 0.987 \\
			& $E_\theta$ & 0.579 & 0.694 & 0.826  & 0.943 \\\hline
			SMPL-P            & $E_\beta$  & 0.856 & 0.946 & 0.972  & 0.948 \\
			$n_p=1.2\text{K}$ & $E_\theta$ & 0.577 & 0.673 & 0.731  & 0.954 \\\hline
			SMPL-P            & $E_\beta$  & 0.856 & 0.926 & 0.990  & 0.944 \\
			$n_p=2\text{K}$   & $E_\theta$ & 0.578 & 0.695 & 0.739  & 0.982 \\
		\end{tabular}
		\caption{
			Retrieval results in the presence of spectral noise.
			Each row corresponds to retrieval results in the presence of spectral noise
			either due to 
			multiplicative Gaussian noise (with strength $\sigma$),
			replacing the spectrum with independent Gaussian noise 
			(mean zero, sigma 50; denoted ``-R''),
			or
			using an LBO estimated from a point cloud 
			(denoted ``-P'', using a point cloud of size $n_p$, either 1200 or 2000).
		}
		\label{table:snoise}
	\end{table}
}

\section{Pose-Aware Retrieval Variance}

In Table~\ref{table:ret_var}, we consider the standard error of the mean (SEM) for the retrieval experiments.
Each model was trained three times with the same hyper-parameters (to account for training stochasticity), and then each model was run three times to account for randomness in the point set sampling of the input shapes.
We show the SEM across models (i.e., trainings) after averaging over point samplings per model.
Since each model has its own SEM over samplings, 
    we display the maximum SEM across models. 
Notice that all SEMs are less than $0.01$, except for two (over model trainings): 
    the $E_\beta$ when using $z$ and $z_I$ on SMAL, 
    which are the most unstable retrieval performances.
In general, training instability is a useful consideration for future work.


\section{Pose-Aware Retrieval with Spectral Noise}

We also investigated the effect of spectral noise on model performance.
We considered three forms of noise:
    (1) by directly injecting Gaussian multiplicative noise 
        into the spectra,
    (2) replacing the spectrum with Gaussian noise (mean zero, sigma 50),
    and
    (3) by extracting the LBO from sampled point clouds 
        (rather than meshes).
To add the noise in (1), we multiply each eigenvalue with independent noise
    $\xi \sim \normd(1,\sigma^2)$, and then enforced non-negativity (via clipping) and monotonicity of the spectrum (by re-sorting the spectrum).
To obtain the spectra in (3), 
    we used a simple Laplacian 
    computed from an affinity matrix, 
    via a radial basis function kernel 
    on the inter-point $L_2$ distance, 
    with bandwidth chosen as 
    $ b_\sigma = d_N^{ -1 / (2 + \xi) }  / 4 $, 
    where $d_N$ is the mean distance of each point to its nearest neighbour 
    and $\xi=0.01$ (similar to \cite{belkin2009constructing}).

We note that the shapes (input point clouds) themselves do not have any additional noise 
    (compared to the standard experimental setup), only the spectra do.
Thus, poorer performance may manifest itself as increased retrieval accuracy 
    with the \textit{wrong} latent space segment 
    (e.g., lower $E_\beta$ when retrieving with $z_E$).

Results are shown in Table~\ref{table:snoise}.
However, we find that the network can tolerate moderate spectral noise, 
    with decreasing performance as noise increases.
For instance, 
    one can see 
    $E_\beta$ with $z_I$ on SMAL 
    and 
    $E_\theta$ with $z_E$ on SMPL
    degrade as $\sigma$ increases.
Very extreme noise, as when replacing the spectrum with Gaussian random values 
    (``SMAL-R'' and ``SMPL-R'' in Table~\ref{table:snoise}), 
    destroys the disentanglement, as the network no longer has access to the isolated intrinsics.
This is similar to ablating the spectral loss, 
    except that predicting the random spectrum adds additional burden on $z_I$.
Finally, 
    for the spectra extracted from the point cloud Laplacians,
    the network degrades slightly on SMPL, compared to using the mesh LBO, but much more so on SMAL.

In practice, we note that our method does not require spectra at test time 
    (for inference or novel sample generation). 
However, it will be affected by noise in the training data 
    (whether in the meshes, spectra, or point clouds).
One approach to improve generalization to noisy point clouds at test time is to use additional noise for data augmentation while training.

\end{document}